\documentclass{article}

\usepackage{arxiv}
\usepackage{hyperref}

\usepackage{graphicx, amsfonts, amsthm, amsxtra, amssymb, verbatim, makeidx}
\usepackage{subeqnarray, relsize}
\usepackage[mathscr]{euscript}
\usepackage[english]{babel}
\usepackage[fixlanguage]{babelbib}
\usepackage[utf8]{inputenc}
\usepackage[english]{babel}

\usepackage{wrapfig}
\usepackage{amssymb, amsmath, amsthm}
\usepackage{amssymb, amsmath, amsthm}
\usepackage{graphicx}
\usepackage{color}
\usepackage{amssymb}
\usepackage{url}
\usepackage{pdfpages}
\usepackage{fancyhdr}
\usepackage{subfig}
\usepackage{titlesec}
\usepackage{enumerate}
\usepackage{comment}
\usepackage{bigints}
\usepackage{diagbox}
\usepackage{cite}
\usepackage{float}
\usepackage{algorithm}
\usepackage{algpseudocode}
\usepackage{array}
\usepackage{multirow}
\usepackage{MnSymbol} 
\usepackage{amsmath}
\usepackage{multirow}
\usepackage{float}
\usepackage{subcaption}
\usepackage{booktabs}
\usepackage{multirow}
\usepackage{enumitem}

\usepackage{tikz}
\usepackage{xcolor}

\newtheorem{thm}{Theorem}[section]

\newtheorem{lem}{Lemma}[section]

\renewenvironment{proof}{{\bfseries \noindent Proof} }{ \qed \\}
\newcommand{\blue}[1]{\begingroup\color{black}#1\endgroup}

\def\R{\mathbb{R}}                   

\title{A Data-Driven Interpolation Method on Smooth Manifolds via Diffusion Processes and Voronoi Tessellations}

\author{
 Alvaro Almeida Gomez \\
 Centro de Modelamiento Matemático\\
  Universidad de Chile\\
Beaucheff 851, Santiago, Chile \\
  \texttt{alvaroalmeidagomez182@gmail.com} \\
}

\begin{document}
\maketitle
\begin{abstract}
We propose a data-driven interpolation framework for reconstructing real-valued functions on smooth manifolds from scattered pointwise observations. The method combines a Gaussian Nadaraya--Watson kernel interpolant with a Voronoi-adaptive bandwidth determined entirely by the geometry of the sampled data, yielding an explicit closed-form construction that requires neither training, iterative optimization, preprocessing, nor parameter tuning.

The proposed interpolant satisfies several theoretical properties. It reproduces the observed data exactly, enforces a vanishing intrinsic gradient at every sample point, and, in the dense-sampling limit, attenuates high-frequency oscillatory components through the geometric regularization induced by the adaptive bandwidth. Furthermore, the construction admits an interpretation in terms of minimizing a discrete total variation--type functional, establishing a natural connection with compressed sensing and sparsity-promoting regularization.

Unlike classical kernel interpolation methods employing a fixed global bandwidth, the proposed adaptive strategy automatically adjusts to the local sampling geometry through the Voronoi tessellation while preserving an explicit analytical formulation. Because the interpolant is available in closed form, the overall computational cost is entirely determined by the inference stage: evaluating the interpolant at a query point requires only the computation of Gaussian kernel weights and their weighted combination, resulting in linear complexity with respect to the number of sample points. In contrast to many data-driven interpolation approaches, no additional offline computational stage is required before inference.

Finally, we demonstrate the practical performance of the proposed methodology in sparse-angle computed tomography reconstruction, where interpolation of the sinogram prior to filtered back-projection produces accurate reconstructions while substantially reducing the overall computational time compared with standard total variation--based reconstruction methods.

\end{abstract}

\blue{
Interpolation is a fundamental problem in applied mathematics, scientific
computing, and machine learning, where the objective is to reconstruct an
unknown function from a finite collection of observations. The subject has a
long history, originating with the classical polynomial schemes of Lagrange
and Newton and evolving, through the resolution of phenomena such as Runge's
instability for high-degree polynomial interpolants and the Whittaker--Shannon
sampling theorem for bandlimited signals, into the modern theory of splines,
radial basis functions, and finite element approximations
\cite{wendland2004scattered,schumaker2007spline}. These methods furnish
powerful and mathematically rigorous frameworks for function reconstruction
and have been applied extensively in approximation theory, numerical
analysis, and scientific computing. Nevertheless, their performance typically
depends on assumptions regarding the regularity of the target function, the
geometry of the underlying domain, or the availability of suitable basis
functions, and highly oscillatory signals often require dense sampling or
high-dimensional approximation spaces to achieve satisfactory reconstruction
accuracy.

Recent developments have extended classical Euclidean interpolation toward
geometry-aware and data-adaptive methodologies. Scattered-data interpolation,
kernel-based techniques, and Gaussian process models provide flexible
nonparametric frameworks for irregularly distributed samples
\cite{fasshauer2007meshfree,rasmussen2006gaussian}. However, these methods
typically rely on similarity measures defined in the ambient space and become
less effective when the data concentrate near a nonlinear, low-dimensional
manifold embedded in a high-dimensional space. This observation motivates
manifold-based interpolation, in which distances, kernels, and differential
operators are adapted to the intrinsic geometry of the data rather than to
the ambient Euclidean metric.

Manifold learning provides a rich and rigorous mathematical framework for
recovering such intrinsic geometric structures from high-dimensional
observations. Laplacian eigenmaps and diffusion maps approximate
differential operators on manifolds through discrete, graph-based
constructions \cite{belkin2003laplacian,coifman2006diffusion}, and have
established that suitably normalized kernel operators converge to the
Laplace--Beltrami operator of the underlying manifold. This connection
between kernel methods, spectral analysis, and differential geometry has
motivated a broader class of data-driven methodologies exploiting intrinsic
geometric information, and it is precisely this connection that the present
work extends to the setting of interpolation.

More recently, geometric deep learning has generalized neural-network
architectures to non-Euclidean domains such as graphs and manifolds
\cite{bronstein2021geometric}. Graph neural networks and message-passing
architectures exploit local connectivity to learn geometric representations
directly from data \cite{kipf2017semi,hamilton2017inductive}. Despite their
empirical success, these architectures are designed primarily for prediction
or representation learning rather than for the construction of explicit
interpolation operators satisfying prescribed pointwise constraints, and they
generally require a learned, data-dependent training procedure rather than an
explicit closed-form construction.

In parallel, neural implicit representations have emerged as powerful tools
for continuous signal reconstruction, representing functions, geometric
shapes, and physical fields through neural networks acting as continuous
approximators \cite{mescheder2019occupancy,sitzmann2020implicit}. Although
highly expressive, such approaches generally require solving large-scale
nonconvex optimization problems and often lack an explicit, analytically
tractable connection to the intrinsic geometry of the sampled data. Taken
together, these two strands of the deep learning literature highlight the
need for interpolation frameworks that combine the geometric adaptivity of
modern data-driven methods with the explicit construction and theoretical
guarantees traditionally associated with approximation theory.

A further limitation of neural-network-based approximators concerns the
reconstruction of highly oscillatory functions. While deep networks are
universal approximators \cite{hornik1989multilayer}, their practical
approximation behavior is governed by the spectral content of the target
function: empirical and theoretical studies have identified a \emph{spectral
bias}, or \emph{frequency principle} \cite{rahaman2019spectral,xu2019frequency},
whereby low-frequency components are learned substantially faster than
high-frequency ones. Consequently, highly oscillatory signals typically
require increased model complexity, larger training sets, or specialized
architectures. This limitation is particularly relevant for multiscale
phenomena, localized singularities, and oscillatory solutions of partial
differential equations
\cite{wang2021understanding},
and several remedies based on Fourier feature embeddings and other spectral
representations have been proposed \cite{tancik2020fourier}. Although
effective in practice, such techniques generally increase computational
complexity and introduce additional hyperparameters whose optimal choice may
be problem-dependent.

Highly oscillatory data are typically characterized by large local
gradients, which pose significant challenges for both classical
interpolation methods and modern learning-based approaches. Motivated by
these difficulties, we develop an interpolation framework that incorporates
a local gradient-control mechanism at the sampling points while preserving
consistency with the observed data. As a consequence, oscillatory features
are regularized directly through the geometry of the interpolation operator,
rather than through an additional optimization procedure or a separate
training stage. This intrinsic regularization provides a natural
alternative to the frequency-dependent mechanisms often employed in
contemporary neural-network architectures.

Building on these insights, we propose a geometric interpolation
framework on a smooth manifold $\mathcal{M}\subset\mathbb{R}^{n}$, based on a
modified Nadaraya--Watson kernel construction
\cite{nadaraya1964estimating,watson1964smooth} with a Gaussian kernel and a
spatially adaptive, Voronoi-type bandwidth inspired by recent developments in
nonparametric statistics \cite{vanLieshout2024}.

Given a finite collection of samples $\{x_i\}_{i=1}^{N}\subset\mathcal{M}$
together with associated values $\{g(x_i)\}_{i=1}^{N}$, we construct a
continuous interpolant $\overline{g}:\mathcal{M}\rightarrow\mathbb{R}$
satisfying the following properties, established in Section~\ref{sec:interpol}:
\begin{enumerate}[label=(\roman*)]
\item \label{prop:exact} Exact interpolation:
\[
\overline{g}(x_i) = g(x_i), \qquad i = 1,\ldots,N.
\]
\item \label{prop:grad} Vanishing intrinsic-gradient condition:
\[
\nabla_{\mathcal{M}}\overline{g}(x_i) = 0, \qquad i = 1,\ldots,N.
\]
\item \label{prop:atten} Attenuation of highly oscillatory components in
the dense-sampling limit, in the sense that high-frequency error components
are damped as $N\to\infty$ with $\{x_i\}_{i=1}^{N}$ becoming dense
in~$\mathcal{M}$; see Section~\ref{sec:smoothness_decay} for the precise
statement.
\item \label{prop:closed} Explicit, closed-form construction, requiring no
iterative optimization or training procedure.
\end{enumerate}
Property~\ref{prop:exact} guarantees exact recovery of the observed data,
while Property~\ref{prop:grad} introduces a local regularization effect by
enforcing critical points, with respect to the intrinsic Riemannian
structure of $\mathcal{M}$, at the interpolation locations; together, these
properties promote smoothness in the reconstructed signal without
sacrificing interpolation accuracy. Property~\ref{prop:atten} stands in
deliberate contrast to the spectral-bias behavior of neural-network-based
approximators discussed above: rather than struggling to resolve
high-frequency content, the proposed interpolant attenuates it by
construction, through the geometric regularization induced by the adaptive
bandwidth. Property~\ref{prop:closed} further distinguishes the construction
from learning-based approaches, since no training procedure, loss function,
or optimization solver is required. Furthermore, since the proposed interpolant is given by an explicit closed-form expression rather than by a trained parametric model, it can be updated by simply recomputing the interpolation with the enlarged dataset when new sample points become available. In contrast, standard neural network approaches generally require retraining or fine-tuning their parameters after incorporating additional data.

Classical kernel interpolation methods commonly employ a fixed global bandwidth,
which determines the scale over which neighboring samples influence the
reconstruction. Although conceptually simple, the choice of this parameter is
well known to be critical. A bandwidth that is too small often leads to highly
oscillatory interpolants and poor generalization between neighboring samples,
whereas an excessively large bandwidth oversmooths the reconstructed function
and may obscure important local features. Moreover, the optimal bandwidth
typically depends on the sampling density and the local behavior of the target
function, making a single global parameter difficult to select in practice.
These limitations motivate the development of adaptive bandwidth strategies
that automatically adjust to the local geometry of the data while preserving
the simplicity of kernel-based interpolation.

Spatially adaptive bandwidth selection has been extensively
studied in the nonparametric statistics literature, beginning
with Abramson's square-root rule~\cite{abramson1982bandwidth},
which sets the local bandwidth proportional to the reciprocal
of the square root of a pilot density estimate, and extended
through balloon and sample-point estimators~\cite{terrell1992variable},
local polynomial regression with data-driven
bandwidths~\cite{fan1996local}, and comprehensive kernel density
frameworks~\cite{silverman1986density}.
These approaches share a common structure: the bandwidth at each
point is chosen---analytically or via cross-validation---to
minimise the mean squared error of a \emph{statistical estimator},
and therefore require either a pilot density estimate or an
auxiliary tuning procedure before the estimator can be applied.

The Voronoi-based bandwidth proposed here differs fundamentally
from these classical strategies in three respects:
\begin{enumerate}
    \item \textbf{Deterministic interpolation.}
    Its purpose is deterministic interpolation rather than statistical
    estimation: the resulting operator satisfies
    $\overline{g}(x_i)=g(x_i)$ exactly at every sample point
    (Property~\ref{prop:exact}), whereas classical adaptive estimators retain a
    nonzero bias at sample sites by design.

    \item \textbf{Closed-form bandwidth selection.}
    The bandwidth is determined analytically from the Voronoi
    tessellation induced by the sample. More precisely, the bandwidth
    at a point $x$ is chosen to be proportional to
    $\min_{j}\lVert x-x_j\rVert$, the distance from $x$ to its
    nearest sample point. This construction requires neither a pilot
    density estimate nor cross-validation, and introduces no auxiliary
    tuning parameters. Consequently, the bandwidth selection procedure
    is entirely explicit and available in closed form
    (Property~\ref{prop:closed}).

    \item \textbf{Regularization of high-frequency components.}
    In contrast to classical adaptive bandwidth methods, whose
    theoretical analysis is based on the bias--variance trade-off,
    our analysis establishes a regularization property of the
    proposed interpolant. As proved in
    Section, the constructed interpolant
    is smoother than the original signal and naturally attenuates
    high-frequency oscillations. Moreover, this smoothing effect is
    shown to be closely related to the minimization of a discrete
    total variation functional, as established in
    Section. Thus, the adaptive bandwidth
    serves not only to preserve exact interpolation but also to
    provide an intrinsic regularization mechanism for highly
    oscillatory data.
\end{enumerate}
Taken together, these distinctions yield concrete advantages in
\emph{accuracy} (exact interpolation at sample sites versus nonzero
estimator bias), \emph{robustness} (gradient control at sample locations
without free parameters, Property~\ref{prop:grad}), and \emph{computational
efficiency} (no iterative, optimisation-based, or pilot-estimation procedure
is required). Unlike classical adaptive bandwidth strategies, whose primary
objective is statistical estimation, our construction is designed
specifically for deterministic interpolation and geometric data analysis.

Properties~\ref{prop:exact} and~\ref{prop:grad} also admit a natural,
if informal, interpretation in the language of discrete total-variation
regularization. The interpolant is not obtained as the minimizer of an
explicit variational problem; nevertheless, the simultaneous enforcement of
exact interpolation and a vanishing intrinsic gradient at the sample points
is exactly the condition under which the sample-restricted functional
\begin{equation}\label{eq:tv_discrete}
  F_N(f) \;:=\; \sum_{i=1}^{N}
    \Bigl(\,\lvert f(x_i)-g(x_i)\rvert
         + \lambda\,\lVert\nabla_{\mathcal{M}} f(x_i)\rVert\,\Bigr)
\end{equation}
attains its global minimum value zero, since both summands are non-negative
and Properties~\ref{prop:exact}--\ref{prop:grad} guarantee that each
vanishes at~$\overline{g}$. We emphasize that~\eqref{eq:tv_discrete}
constrains $f$ only at the sample points and is therefore not, by itself, a
meaningful selection criterion among interpolants; the observation above is
included only to situate Properties~\ref{prop:exact}--\ref{prop:grad} within
the broader spirit of sparsity-promoting, $\ell^{1}$-type regularization
that underlies compressed sensing and related inverse-problem methodology
\cite{candes2006stable,taocomp},
rather than to claim a variational characterization of the proposed
interpolant. Such formulations are particularly effective for
piecewise-smooth functions and are standard tools in inverse problems and
image reconstruction.

As an application, we consider sparse-angle computed tomography
reconstruction \cite{zhang2018sparse,bao2019sparse}. Classical reconstruction
methods typically require solving large-scale regularized optimization
problems, often involving total variation penalties and iterative numerical
algorithms. The proposed methodology instead approximates the sinogram
directly from sparse measurements and reconstructs the image via filtered
back-projection; this introduces a regularization effect already at the
interpolation stage, reducing computational cost while remaining compatible
with standard variational reconstruction methods.

In summary, the contributions of this work are twofold:
(a)~a geometric Nadaraya--Watson interpolant with a Voronoi-adaptive
bandwidth, satisfying the exact interpolation, intrinsic gradient control,
high-frequency attenuation, and closed-form construction properties
established in Section~\ref{sec:interpol}, together with a heuristic
connection to discrete total-variation regularization that situates the
construction within the broader framework of sparsity-promoting inverse-problem
methodology; and
(b)~an application to sparse-angle computed-tomography reconstruction,
illustrating the practical benefits of the proposed geometric regularization
mechanism.

The remainder of this paper is organized as follows.
Section~\ref{sec:interpol} introduces the proposed interpolation methodology
and develops its theoretical properties. Section~\ref{sec:applications}
presents applications to numerical approximation and sparse computed
tomography reconstruction. Section~\ref{sec:numeexp} reports numerical
experiments illustrating the performance of the proposed approach. Finally,
Section~\ref{sec:conclusions} concludes the paper and discusses directions
for future research.
}


\section{Proposed Methodology}
\label{sec:interpol}

In this section, we describe in detail the proposed data-driven interpolation methodology, along with its main mathematical properties. The section is organized as follows. In Section~\ref{sec:gauss}, we introduce the notation and provide the motivation for the proposed approach. In Section~\ref{sec:stabil}, we present the interpolation methodology and the associated algorithm. In Section~\ref{sectcompcomplexity}, we analyze the computational complexity and compare the proposed method with other data-driven interpolation techniques. In Section~\ref{sec:smoothness_decay}, we study the local first-order regularity of the method. Finally, in Section~\ref{sec:highfrequency}, we discuss its spectral properties.

\subsection{Gaussian approximations}
\label{sec:gauss}

In this paper, we consider the Gaussian kernel $G$ defined by
\begin{equation}
    G(z) = \exp(-\|z\|^2).
\end{equation}
For a fixed $\varepsilon > 0$, we introduce the rescaled Gaussian kernel $G_{\varepsilon}$ as
\begin{equation*}
    G_{\varepsilon}(z) = G\left(\frac{z}{\varepsilon}\right) = \exp\left(-\frac{\|z\|^2}{\varepsilon^2}\right).
\end{equation*}
We assume the \emph{manifold hypothesis}, which states that the dataset lies on a  d-dimensionalsmooth submanifold $\mathcal{M} \subset \mathbb{R}^n$ and is distributed according to a smooth probability density function $\rho : \mathcal{M} \to \mathbb{R}$. Let $g : \mathcal{M} \to \mathbb{R}$ be a smooth function defined on the manifold $\mathcal{M}$. We define the convolution operator
\begin{equation}
\label{eq1}
    \hat{g}_{\varepsilon}(x) = \int_{\mathcal{M}} G_{\varepsilon}(x - y)\, g(y)\, \rho(y)\, d\mathrm{Vol}(y),
\end{equation}
where $d\mathrm{Vol}$ denotes the volume form induced by the Riemannian metric on $\mathcal{M}$. We also define the normalization function
\begin{equation}
\label{eq2}
    d_{\varepsilon}(x) = \int_{\mathcal{M}} G_{\varepsilon}(x - y)\, \rho(y)\, d\mathrm{Vol}(y),
\end{equation}
and the corresponding normalized approximation
\begin{equation}
\label{eq3}
    g_{\varepsilon}(x) = \frac{\hat{g}_{\varepsilon}(x)}{d_{\varepsilon}(x)}.
\end{equation}
The function $g_{\varepsilon}$ provides a smooth approximation of $g$, obtained via convolution with the Gaussian kernel. This approximation is closely connected to the infinitesimal generator of the heat equation on the manifold, as established in the Diffusion Maps framework~\cite{coifman2006diffusion} and in Eigenmaps \cite{belkin2003laplacian}. The following result, whose proof can be found in~\cite{coifman2006diffusion}, formalizes this convergence and provides an asymptotic error estimate in terms of the diffusion scaling parameter $\varepsilon$.

It is important to emphasize that this connection is purely analytical. In the present work, the asymptotic expansion is used only to motivate the Gaussian kernel estimator, whereas the proposed interpolation methodology does not rely on the construction of diffusion operators, spectral embeddings, or Laplace--Beltrami approximations.

\begin{thm}
\label{teoremaprincipal}
Let $g \in C^{\infty}(\mathcal{M})$. Then its Gaussian approximation $g_{\varepsilon}$ satisfies the infinitesimal expansion
\begin{equation*}
    g_{\varepsilon}(x) = g(x) + \frac{\varepsilon^2}{2} \Delta g(x) + O(\varepsilon^3),
\end{equation*}
where $\Delta$ denotes the Laplace--Beltrami operator. In particular, $g_{\varepsilon}$ converges pointwise to $g$ as $\varepsilon \to 0$:
\begin{equation*}
    \lim_{\varepsilon \to 0} g_{\varepsilon}(x) = g(x), \quad \text{for all } x \in \mathcal{M}.
\end{equation*}
\end{thm}

Therefore, for sufficiently small \( \varepsilon \),  the approximation\( g_{\varepsilon} \) serves as an effective approximation of \( g \).  Although $g_{\varepsilon}$ is originally defined only on the manifold $\mathcal{M}$, the formulation in \eqref{eq1}--\eqref{eq3} naturally extends its definition to the ambient space $\mathbb{R}^n$. This extension enables the evaluation of $g_{\varepsilon}(x)$ for any point $x \in \mathbb{R}^n$ using only information from $\mathcal{M}$. In other words, based on the values of $g$ restricted to the manifold, we can infer a smooth extension of the function to points outside $\mathcal{M}$.

\medskip

From a computational perspective, the Law of Large Numbers provides a theoretical foundation for approximating the function $g_{\varepsilon}$ via Monte Carlo integration. Given a collection of $N$ i.i.d.\ samples
\[
X = \{x_1, \dots, x_N\}
\]
drawn from a probability density $\rho$, the function $g_{\varepsilon}(x)$ can be approximated by
\begin{equation}
\label{approximacion1}
g_{\varepsilon}(x) \approx \frac{1}{ \mathbf{Nm}_{\varepsilon}} \sum_{i=1}^{N} G_{\varepsilon}(x - x_i)\, g(x_i),
\end{equation}
where the empirical normalization factor is defined as
\begin{equation}
\label{factordiscnorm}
 \mathbf{Nm}_{\varepsilon} = \sum_{i=1}^{N} G_{\varepsilon}(x - x_i).
\end{equation}

The objective of this paper is to exploit the approximation in \eqref{approximacion1} to estimate the value of the function \( f(x) \) at points \( x \notin X \), that is, points that do not belong to the training dataset \( X \). We assume that these points are realizations of a random variable distributed according to the same smooth probability density \( \rho \).

We remark that the proposed methodology extends naturally to vector-valued functions.
If, instead of a real-valued function \( g : \mathcal{M} \to \mathbb{R} \), we consider a vector-valued function
\( g : \mathcal{M} \to \mathbb{R}^n \),
the procedure can be applied componentwise.
That is, the proposed approximation is applied independently to each coordinate function, each of which is real-valued.

\blue{A particularly relevant application of this vector-valued formulation arises in tomographic reconstruction. In this setting, a sinogram can be modeled as a vector-valued function
\[
g:S^1\rightarrow\mathbb{R}^m,
\]
where \(S^1\) parameterizes the acquisition angle and \(m\) denotes the number of detector pixels in each tomographic projection. The proposed interpolation method can then be applied componentwise to interpolate the sinogram prior to image reconstruction. A detailed treatment of this application, including the corresponding angular regularization framework, is presented in Section~\ref{subsec:angular_regularization}.}

\medskip

\subsection{Infinitesimal scale \texorpdfstring{$\varepsilon$}{epsilon}, Voronoi cells, and numerical stability}
\label{sec:stabil}

A crucial component of the proposed methodology is the selection of the \blue{bandwidth} parameter $\varepsilon$, which plays the role of an infinitesimal scale in accordance with Theorem~\ref{teoremaprincipal}. As shown therein, $\varepsilon$ must be sufficiently small for the kernel-based approximation $g_\varepsilon$ to accurately recover the underlying smooth function $g$.

However, choosing $\varepsilon$ too small may lead to severe numerical instabilities. In particular, the Gaussian weights
\begin{equation}
    G_\varepsilon(z) = \exp\!\left(-\frac{\|z \|^2}{\varepsilon^2}\right),
\end{equation}
which appear in the discrete approximation \eqref{approximacion1}, decay rapidly as $\varepsilon \to 0$. As a consequence, the kernel becomes overly localized, effectively neglecting the contribution of neighboring data points and deteriorating the quality of the approximation. 

Moreover, the normalization factor $\mathbf{Nm}_\varepsilon$, defined in \eqref{factordiscnorm}, may become arbitrarily small. This leads to numerical instability due to the presence of the reciprocal term
\begin{equation}
    \frac{1}{\mathbf{Nm}_\varepsilon}.
\end{equation}

To address these issues, we introduce a spatially adaptive scale parameter $\varepsilon(x) > 0$, defined for all $x \in \blue{ \mathcal{M}\setminus \{ x_i \}_{i=1}^N}$, such that the normalization factor remains uniformly bounded from below. More precisely, we require
\begin{equation}
    \mathbf{Nm}_{\varepsilon(x)}(x) \geq C > 0,
    \qquad \text{for all } x \in \blue{ \mathcal{M}\setminus \{ x_i \}_{i=1}^N},
\end{equation}
where $C$ is a constant independent of $x$. This condition prevents numerical instabilities arising from the term
\begin{equation}
    \frac{1}{\mathbf{Nm}_{\varepsilon(x)}(x)}.
\end{equation}

We define the adaptive parameter $\varepsilon(x)$ for \blue{$x \in \mathcal{M}\setminus \{ x_i \}_{i=1}^N$} in terms of the local geometry of the data, namely as the distance from $x$ to the nearest sample point:
\begin{equation}\label{parametroepsilon}
    \varepsilon(x) := \min_{1 \le i \le N} \|x - x_i \| > 0.
\end{equation}

\blue{
An important feature of the adaptive bandwidth function $\varepsilon(x)$ is that it encodes the Voronoi decomposition (Figure \ref{fig:voronoi_4points}) generated by the sampling points $\{x_i\}_{i=1}^N$. More precisely,  the Voronoi cell associated with $x_i$ is given by
\[
V_i=\left\{x\in\mathcal{M}:\varepsilon(x)=\|x-x_i\|\right\}.
\]
Hence, the Voronoi tessellation admits a natural analytic representation through the function $\varepsilon(x)$.
}


\begin{figure}[htbp]
    \centering
    \begin{tikzpicture}[scale=0.5]
        \definecolor{cellblue}{RGB}{210, 225, 242}
        \definecolor{cellred}{RGB}{245, 215, 215}
        \definecolor{cellgreen}{RGB}{215, 240, 215}
        \definecolor{cellyellow}{RGB}{252, 243, 207}

        \path[use as bounding box] (0,0) rectangle (8,8);

        \fill[cellblue] (0,0) -- (4,0) -- (4.5, 3.8) -- (3.5, 4.2) -- (0, 4.5) -- cycle;
        
        \fill[cellred] (4,0) -- (8,0) -- (8, 4.2) -- (4.5, 3.8) -- cycle;
        
        \fill[cellgreen] (4.5, 3.8) -- (8, 4.2) -- (8,8) -- (3,8) -- (3.5, 4.2) -- cycle;
        
        \fill[cellyellow] (0, 4.5) -- (3.5, 4.2) -- (3,8) -- (0,8) -- cycle;

        \draw[line width=1.2pt, color=darkgray] (0, 4.5) -- (3.5, 4.2);
        \draw[line width=1.2pt, color=darkgray] (3,8) -- (3.5, 4.2);
        \draw[line width=1.2pt, color=darkgray] (3.5, 4.2) -- (4.5, 3.8);
        \draw[line width=1.2pt, color=darkgray] (4.5, 3.8) -- (4,0);
        \draw[line width=1.2pt, color=darkgray] (4.5, 3.8) -- (8, 4.2);

        \draw[line width=1.5pt, black] (0,0) rectangle (8,8);

        \filldraw[black] (1.8, 2.2) circle (2.5pt) node[below right] {$P_1$};
        \filldraw[black] (6.2, 1.8) circle (2.5pt) node[below left] {$P_2$};
        \filldraw[black] (5.8, 6.0) circle (2.5pt) node[above left] {$P_3$};
        \filldraw[black] (1.5, 6.2) circle (2.5pt) node[above right] {$P_4$};

    \end{tikzpicture}
\caption{Example of a Voronoi tessellation generated by four sample points, with each Voronoi cell displayed in a different color.}
    \label{fig:voronoi_4points}
\end{figure}


With this definition, we immediately obtain
\begin{equation}
    \mathbf{Nm}_{\varepsilon(x)}(x)
    \;\geq\;
    \exp\!\left(
        -\left(
            \frac{\min_{1 \le i \le N} \|x - x_i \|}{\varepsilon(x)}
        \right)^{\!2}
    \right)
    \;=\;
    \exp(-1),
\end{equation}
which guarantees that $\mathbf{Nm}_{\varepsilon(x)}(x)$ is uniformly bounded away from zero over  \blue{ the set $\mathcal{M}\setminus \{ x_i \}_{i=1}^N$ }.

Observe that the choice of $\varepsilon(x)$ is intrinsically geometric: it is determined by the Voronoi cells associated with the dataset $X = \{x_1,\dots,x_N\}$. Consequently, the method adapts to the local density and structure of the data, while the Gaussian kernel induces a diffusion process that reflects this geometry.

Summarizing, given a collection of $N$ i.i.d.\ samples
\[
X = \{x_1, \dots, x_N\},
\]
together with the function evaluations $g(x_1), g(x_2), \dots, g(x_N)$, we define the following interpolant \blue{for all $x \in \mathcal{M}\setminus \{ x_i \}_{i=1}^N$ as:}
\begin{equation}
\label{appropfinalsample}
    \bar{g}_{x_1,\dots,x_N}(x)
    :=
    \frac{1}{\mathbf{Nm}_{\varepsilon(x)}(x)}
    \sum_{i=1}^{N}
    G_{\varepsilon(x)}(x - x_i)\, g(x_i),
\end{equation}
where the adaptive scale $\varepsilon(x)$ is given by \eqref{parametroepsilon}. This interpolant will be the main object of study throughout the remainder of the paper. 

Structurally, the interpolant defined in \eqref{appropfinalsample} can be interpreted as a modified Nadaraya--Watson kernel estimator. While the classical estimator typically employs a fixed bandwidth $\epsilon$, our formulation introduces a spatially adaptive bandwidth $\epsilon(x)$ determined by the local Voronoi tessellation of the manifold. This spatial adaptation ensures the geometric consistency and numerical stability discussed previously.

The complete methodology, including the adaptive selection of $\varepsilon(x)$, is summarized in Algorithm~\ref{alg:datadriven_interpolant}.

\begin{algorithm}[H]
\caption{Data-Driven Interpolant}
\label{alg:datadriven_interpolant}

\textbf{Input:}
\begin{itemize}
    \item Data set $X = \{x_1, x_2, \dots, x_N\} \subset \mathcal{M}$,
    \item Function values $\{g(x_i)\}_{i=1}^N$,
    \item Query point $x \in \mathcal{M} \setminus X$.
\end{itemize}

\textbf{Procedure:}
\begin{enumerate}
    \item Compute the parameter $\varepsilon(x)$ according to \eqref{parametroepsilon}.
    \item Evaluate the interpolant $g_{\varepsilon}(x)$ using \eqref{appropfinalsample}.
\end{enumerate}

\textbf{Output:} Approximation $\bar{g}_{x_1,\dots,x_N}(x)$.
\end{algorithm}

\blue{
Finally, we note that the interpolant produced by
Algorithm~\ref{alg:datadriven_interpolant} is initially defined only on
the set $\mathcal{M}\setminus X$, where
$X=\{x_1,\ldots,x_N\}$ denotes the collection of sampling points.
To obtain a function defined on the entire manifold $\mathcal{M}$,
we extend the interpolant by assigning to each sampling point its
prescribed data value. Specifically, we define
}

\begin{equation}
\widetilde{g}_{x_1,\ldots,x_N}(x):=
\begin{cases}
\bar{g}_{x_1,\ldots,x_N}(x),
& \text{if } x\in\mathcal{M}\setminus X,\\[4pt]
g(x_i),
& \text{if } x=x_i \text{ for some } i\in\{1,\ldots,N\}.
\end{cases}
\end{equation}

\blue{
Here, $\bar{g}_{x_1,\ldots,x_N}$ denotes the interpolant generated by
Algorithm~\ref{alg:datadriven_interpolant}. By construction, the extension
$\widetilde{g}_{x_1,\ldots,x_N}$ is defined on all of $\mathcal{M}$ and
satisfies the interpolation conditions
\[
\widetilde{g}_{x_1,\ldots,x_N}(x_i)=g(x_i),
\qquad i=1,\ldots,N.
\]
The continuity and regularity properties of the extended interpolant will
be analyzed in Section~\ref{sec:smoothness_decay}. 
}

\subsection{Computational Complexity}
\label{sectcompcomplexity}
The computational cost of evaluating the interpolant $\bar{g}_{x_1, \dots, x_N}(x)$ at a query point $x \in \mathcal{M}$ is primarily determined by two operations: the calculation of the adaptive scale $\varepsilon(x)$ and the weighted summation of the Gaussian kernels. 

\begin{enumerate}
    \item \textbf{Adaptive Scale Computation:} According to \eqref{parametroepsilon}, $\varepsilon(x)$ is determined by the minimum distance to the set $X$. A brute-force search involves $N$ distance computations in $\mathbb{R}^n$, yielding a complexity of $O(Nn)$. 
    \item \textbf{Interpolant Evaluation:} The evaluation of the sum in \eqref{appropfinalsample} requires computing $N$ Gaussian weights $G_{\varepsilon(x)}(x-x_i)$, each involving a $n$-dimensional vector subtraction and norm. This results in a complexity of $O(Nn)$.
\end{enumerate}

Consequently, the overall computational complexity per query point is $O(Nn)$. This choice is justified by the fact that, although the evaluation scales linearly with the number of samples $N$, the method does not require any prior training or iterative optimization procedures (such as backpropagation in neural networks). As a result, the computational cost is entirely concentrated in the inference stage, making the approach particularly suitable for applications where the model must be constructed on demand.

Moreover, for large-scale datasets, the summation in \eqref{appropfinalsample} can be restricted to a local neighborhood of $x$. This truncation is justified by the exponential decay of the Gaussian kernel $G_{\varepsilon(x)}$, which makes the contribution of distant points negligible. As a result, the computational cost of the proposed algorithm is significantly reduced.

\blue{We conclude with a remark on the computational complexity of the proposed scheme.
Evaluating the interpolant at a single query point requires computing all $N$
Gaussian weights and forming their linear combination, at a cost of
$\mathcal{O}(N)$. We emphasize that this asymptotic complexity is shared by
many kernel-based interpolation methods and is not, by itself, the source of
the computational advantage of our approach. The relevant comparison is
instead the \emph{total} cost of producing a prediction, which for classical
data-driven methods must also include an upfront training or preprocessing
stage. Radial basis function (RBF) interpolation, for instance, requires
solving a dense $N \times N$ linear system for the interpolation coefficients
before any query can be answered, at a cost of $\mathcal{O}(N^3)$; Gaussian
process regression requires factorizing an $N \times N$ covariance matrix,
typically via Cholesky decomposition, at a cost of $\mathcal{O}(N^3)$, with
each subsequent posterior-mean evaluation costing an additional
$\mathcal{O}(N)$. The proposed method, by contrast, requires no training or
preprocessing step whatsoever: each query is answered directly and locally
from the sample data, so that its $\mathcal{O}(N)$ per-query cost constitutes
the \emph{entire} computational expense of the method. Consequently, although
the asymptotic inference complexity of our approach is comparable to that of
several classical interpolants, its total cost is substantially lower
whenever the training overhead of these methods cannot be amortized over
many queries---for instance, when only a few evaluations are needed, or when
the sample set is updated frequently and classical methods must be refit
from scratch. It is this absence of a training phase, rather than the linear
scaling of inference per se, that is the principal source of the
computational efficiency of the proposed methodology.}

\subsection{Smoothness and Decay Properties}
\label{sec:smoothness_decay}

In this section, we study several analytical properties of the interpolant 
$\bar{g}_{x_1,\dots,x_N}(x)$ associated with the function $g$, defined in 
\eqref{appropfinalsample} and \eqref{factordiscnorm}. 
We emphasize that this interpolant is implemented in practice via 
Algorithm~\ref{alg:datadriven_interpolant}.

We recall that the interpolant $\bar{g}_{x_1,\dots,x_N}(x)$ is defined by
\begin{equation*}
    \bar{g}_{x_1,\dots,x_N}(x)
    :=
    \frac{1}{\mathbf{Nm}_{\varepsilon(x)}(x)}
    \sum_{i=1}^{N}
    G_{\varepsilon(x)}(x - x_i)\, g(x_i),
\end{equation*}
where the normalization factor is given by
\begin{equation*}
   \mathbf{Nm}_{\varepsilon}(x) 
   := 
   \sum_{i=1}^{N} G_{\varepsilon(x)}(x - x_i).
\end{equation*}
Moreover, according to the results in Section~\ref{sec:stabil}, for each fixed $x \in \mathcal{M}$ the parameter 
$\varepsilon(x)$ is chosen as
\begin{equation*}
    \varepsilon(x) 
    := 
    \min_{1 \le i \le N} \|x - x_i\|.
\end{equation*}

Under this choice of parameter, and following the construction introduced in Section~\ref{sec:interpol}, 
the interpolant $\bar{g}_{x_1,\dots,x_N}(x)$ is initially defined only on 
$\mathcal{M} \setminus X$, where $X = \{x_1, \dots, x_N\}$. 
However, it admits a natural extension to the entire manifold $\mathcal{M}$ given by
\begin{equation}
\label{eq:extension}
\bar{g}_{x_1,\dots,x_N}(x)
=
\begin{cases}
\bar{g}_{x_1,\dots,x_N}(x), & \text{if } x \in \mathcal{M} \setminus X, \\[4pt]
g(x_i), & \text{if } x = x_i \text{ for some } i \in \{1,\ldots,N\}.
\end{cases}
\end{equation}

Without loss of generality, and with a slight abuse of notation, we denote by 
$\bar{g}_{x_1,\dots,x_N}(x)$ both the original interpolant and its extension to $\mathcal{M}$.

\blue{}
Let $1 \le p < \infty$ and let $v : \mathcal{M} \to \mathbb{R}^l$ be a smooth vector-valued function 
defined on the manifold $\mathcal{M}$. Assume that $x_1, x_2, \dots, x_N$ are independent and identically 
distributed random samples drawn from a smooth and strictly positive probability density $\rho$ on $\mathcal{M}$. 

We define the discrete $\ell^p$ norm by
\begin{equation*}
    \|v\|_{\ell^p} 
    := 
    \left( \sum_{j=1}^N \|v(x_j)\|^p \right)^{1/p}.
\end{equation*}
Similarly, for an open subset $B \subset \mathcal{M}$, we define the $L^p(B)$ norm by
\begin{equation*}
    \|v\|_{L^p(B)} 
    := 
    \left( \int_B \|v(x)\|^p \, dx \right)^{1/p}.
\end{equation*}

For $r>0$ and $x \in \mathbb{R}^n$, we denote by $B(x,r)$ the open ball centered at $x$ with radius $r$, i.e.,
\begin{equation*}
    B(x,r) := \{\, y \in \mathbb{R}^n \mid \|x - y\| < r \,\}.
\end{equation*}

We now study the continuity and differentiability properties of the function 
$\bar{g}_{x_1,\dots,x_N}$. \blue{%
To prove continuity, observe that for any point
$x \in \mathcal{M} \setminus \{x_i\}_{i=1}^N$
the function $\varepsilon(x)>0$, where $\varepsilon(x)$ is defined
in Equation~\eqref{parametroepsilon}.
Since $\varepsilon$ is continuous, the reciprocal $1/\varepsilon(x)$
is continuous on $\mathcal{M}\setminus\{x_i\}_{i=1}^N$,
and hence $\bar{g}_{x_1,\dots,x_N}$ is continuous on this set.

It remains to verify continuity at the sample points.
On each Voronoi cell
\begin{equation*}
  R_k = \bigl\{ y \in \mathcal{M} :
        \|y - x_k\| \le \|y - x_i\|
        \text{ for all } 1 \le i \le N \bigr\},
\end{equation*}
the interpolant admits the representation
\begin{equation*}
  \bar{g}_{x_1,\dots,x_N}(x)
  =
  \frac{\displaystyle\sum_{i=1}^{N}
        \exp\!\Bigl(-\|x - x_i\|^2 / \|x - x_k\|^2\Bigr)\,g(x_i)}
       {\displaystyle\sum_{i=1}^{N}
        \exp\!\Bigl(-\|x - x_i\|^2 / \|x - x_k\|^2\Bigr)}.
\end{equation*}
For $i \neq k$, as $x \to x_k$ the ratio
$\|x - x_i\|^2 / \|x - x_k\|^2 \to +\infty$,
since the numerator converges to $\|x_k - x_i\|^2 > 0$
while the denominator tends to zero.
Consequently,
\[
  \lim_{x \to x_k}
  \exp\!\Bigl(-\|x - x_i\|^2 / \|x - x_k\|^2\Bigr) = 0,
  \qquad i \neq k.
\]
For $i = k$, the exponent equals
$-\|x - x_k\|^2/\|x - x_k\|^2 = -1$ for all $x \neq x_k$,
so that term contributes $e^{-1}g(x_k)$ to the numerator
and $e^{-1}$ to the denominator.
Taking the limit in the representation above therefore gives
\begin{equation*}
  \lim_{x \to x_k} \bar{g}_{x_1,\dots,x_N}(x) = g(x_k),
\end{equation*}
which establishes continuity of the interpolant over the entire manifold.
Moreover, the representation above shows that
$\bar{g}_{x_1,\dots,x_N}$ is $C^{\infty}$ on the interior of
each cell $R_k \setminus \{x_k\}$.
The next result shows that differentiability extends to the sample
points $x_k$ and, moreover, that the gradient of
$\bar{g}_{x_1,\dots,x_N}$ vanishes there.%
}

We now turn to the behavior of $\bar{g}_{x_1,\dots,x_N}$ at the points $x_k$. 
The following results establish additional analytical properties of the interpolant. 
In particular, they show that $\bar{g}_{x_1,\dots,x_N}$ acts as a regularization of $g$, 
in the sense that, within sufficiently small neighborhoods of the sample points $x_k$, 
the gradient of $\bar{g}_{x_1,\dots,x_N}$ decays rapidly.

\begin{thm}
\label{thm:local_decay}
Let $\{x_i\}_{i=1}^N \subset \mathcal{M}$ be a set of distinct i.i.d.\ sample points drawn from a smooth and strictly positive density $\rho$. Let $\bar{g}_{x_1,\dots,x_N}$ denote the interpolant defined on $\mathcal{M}$. Then, $\bar{g}_{x_1,\dots,x_N}$ is smooth in a neighborhood of each sample point $x_i$ and satisfies the following properties.

For each $1 \le i \le N$, define the local separation radius
\[
r_i := \min_{j \neq i} \|x_i - x_j\|.
\]
Then, for every $x \in B(x_i, r_i/2) \cap (\mathcal{M} \setminus \{x_i\})$, there exists a constant $C>0$ such that
\[
\|\nabla \bar{g}_{x_1,\dots,x_N}(x)\| 
\le 
C \, \frac{\|g\|_{\ell^1}}{\|x - x_i\|^4} 
\exp\!\left( -\frac{r_i^2}{4\|x - x_i\|^2} \right).
\]
Moreover,
\[
\nabla \bar{g}_{x_1,\dots,x_N}(x_i) = 0, 
\quad \text{for all } i = 1, \dots, N.
\]
\end{thm}

The proof of this result is given in Appendix~\ref{pruebateoprin}.

As a consequence, we obtain decay estimates in $L^p$ norms over neighborhoods of the sample points.

\begin{thm}
\label{thm:lp_decay}
Under the assumptions of Theorem~\ref{thm:local_decay}, let $1 \le p < \infty$. 
Then, for $\beta>0$ sufficiently small, the following estimate holds:
\begin{equation*}
\|\nabla \bar{g}_{x_1,\dots,x_N}\|_{L^p\!\left( \mathcal{M} \cap \bigcup_{i=1}^N B(x_i,\beta) \right)}
\le 
C \|g\|_{\ell^1}
\, \beta^{-4}
\exp\!\left(-\frac{r^2}{4\beta^2}\right)
\left(
\mathrm{vol}\!\left(\mathcal{M}\right)
\right)^{1/p}.
\end{equation*}
where 
\[
r := \min_{1 \le i \le N} r_i.
\]
\end{thm}

The proof is given in Appendix~\ref{pruebateocota}.

\blue{We remark that, in general, smoothness may fail across the boundaries of the Voronoi cells, since the bandwidth function $\varepsilon(x)$ is defined as the minimum of $k$ continuous functions (the distances from $x$ to each data point). At points where this minimum distance from $x$ to the data set $X$ is attained simultaneously by two or more distinct data points, that is, at the intersection of the boundaries of the corresponding Voronoi cells, the function $\varepsilon(x)$ fails to be differentiable, since the minimum of smooth functions is differentiable only where the minimizer is unique.}

\blue{
Theorem~\ref{thm:local_decay} establishes that the gradient norm of the interpolant is bounded by an exponentially decaying factor, whose decay rate depends jointly on the separation radius $r_i$ and the distance $\|x - x_i\|$ between a point $x$ and the center $x_i$ of the Voronoi cell containing it.

The dominant term in this bound is the ratio $\|x-x_i\|/r_i$: as $x \to x_i$, the exponential factor tends to zero, forcing the gradient norm to vanish at the same rate. Consequently, the interpolant is approximately constant in a neighborhood of $x_i$ whose radius is controlled by $r_i$.

Since the separation radius $r_i$ typically decreases as the number of sample points $N$ increases, the region around $x_i$ in which the gradient remains strongly attenuated also shrinks. Thus, refining the sampling set does not eliminate the locally constant behavior of the interpolant; rather, it localizes such behavior within increasingly small neighborhoods of each Voronoi center.

This provides a geometric interpretation of Theorem~\ref{thm:local_decay}: the Voronoi decomposition induced by the sample points determines the local structure of the interpolant, while $r_i$ sets the characteristic scale over which the interpolant deviates from its value at $x_i$. In this sense, the estimate precisely quantifies how the geometry and density of the sampling set control the local regularity of the interpolant, so that finer sampling produces sharper, more localized transitions rather than a globally smoother reconstruction.
}

\begin{figure}[ht]
\centering
\includegraphics[width=\textwidth]{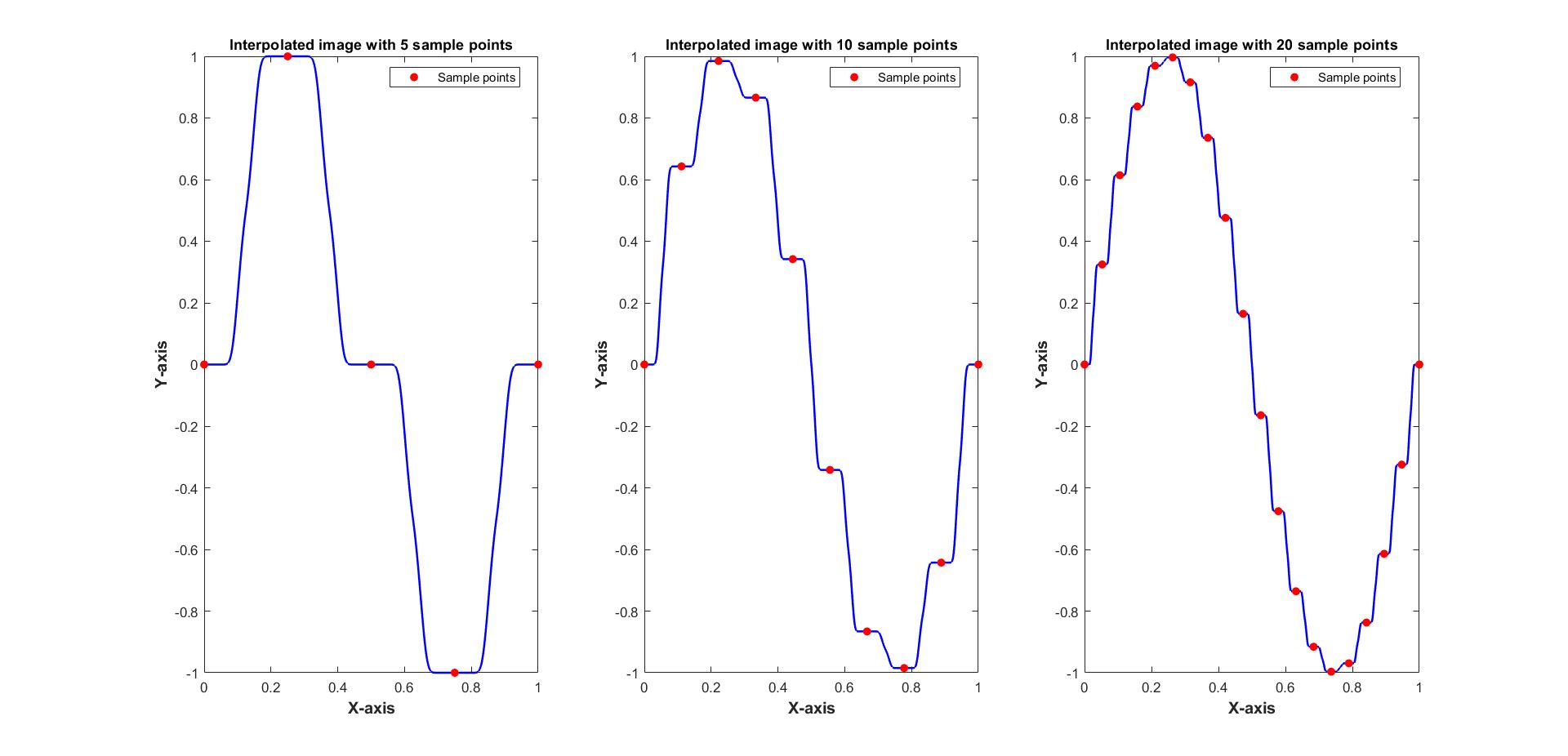}
\caption{Interpolation of the sine function on the interval $[0,1]$ using the proposed method with uniformly spaced sample points. From left to right, the interpolants are constructed from $5$, $10$, and $20$ sample points, respectively. The red markers indicate the sampling points used in the interpolation process.}
\label{fig:sine_interpolation}
\end{figure}

\blue{To illustrate this phenomenon, consider the function $f:[0,1] \to \R$ given by $f(x)=\sin(x)$. We apply the proposed methodology to interpolate $f$ using $5$, $10$, and $20$ uniformly spaced sample points. The resulting interpolants are displayed in Figure \ref{fig:sine_interpolation}, where the red markers indicate the sample points. Observe that as the number of sample points increases from $5$ to $20$, the size of the neighborhood around each point where the interpolant is flat—or nearly constant—decreases. This behavior is consistent with the theory developed in Theorem~\ref{thm:local_decay}.}

As a consequence of the previous results, given a set of sample points $\{x_i\}_{i=1}^N$, 
the interpolant $\bar{g}_{x_1,\dots,x_N}$ can be interpreted as a regularized version of the function $g$.

More precisely, for any $\lambda > 0$ and $\beta$ sufficiently small, the function 
$\bar{g}_{x_1,\dots,x_N}$ is a minimizer of the total variation regularization problem
\begin{equation}
\label{regindugeral}
  \min_{f} \left\{
    \| f - g \|_{\ell^1}
    + \lambda \, \| \nabla f \|_{\ell^1}
  \right\}.
\end{equation}

In fact, a stronger property holds. The interpolation error vanishes at the sample points, namely,
\[
\bar{g}_{x_1,\dots,x_N}(x_i) = g(x_i), \quad i=1,\dots,N,
\]
and the gradient satisfies
\[
\nabla \bar{g}_{x_1,\dots,x_N}(x_i) = 0, \quad i=1,\dots,N.
\]
Furthermore, Theorem~\ref{thm:lp_decay} shows that the gradient not only vanishes at the sample points, 
but also decays rapidly in neighborhoods of each $x_i$.

Therefore, as a consequence of the proposed methodology, we conclude that it solves the total variation 
regularization problem in the particular case where the measurement matrix is the identity. 
Moreover, according to compressive sensing theory \cite{dohcomp, taocomp, candes2006stable}, 
if the original signal $g$ is sparse, it can be accurately reconstructed from a set of subsampled observations.

A key advantage of the proposed algorithm, compared to standard compressive sensing approaches, 
is that it does not require solving an optimization problem. Instead, it provides an explicit interpolation 
formula whose construction is based on the geometry of the input data set.

\subsection{High-Frequency Attenuation via Gradient Smoothing}
\label{sec:highfrequency}

We now establish the spectral properties of the proposed methodology, specifically concerning the attenuation of high-frequency components for a discrete signal defined on a compact, connected, and without boundary manifold $\mathcal{M}$. This regularization is achieved by penalizing the Dirichlet energy of the signal through the smoothing procedure detailed in Algorithm~\ref{alg:datadriven_interpolant}.

The underlying principle rests on the observation that high-frequency components correspond to large eigenvalues of the Laplace–Beltrami operator $\Delta_{\mathcal{M}}$, and consequently, to large gradient norms. By minimizing the gradient energy subject to the interpolation constraints, the proposed method suppresses high-frequency oscillations while ensuring the interpolant $f$ exactly reproduces the sampled data $y_i = f(x_i)$.

By filtering the upper tail of the manifold's spectrum, the reconstructed signal concentrates its energy within the low-frequency eigenfunctions. The resulting representation is not only regularized but also yields a more compact spectral decomposition, effectively denoising the signal while maintaining fidelity to the observed points.

\medskip

We now formalize this intuition. Let $\mathcal{M}$ be a compact smooth manifold without boundary, and let $\Delta$ denote the Laplace--Beltrami operator on $\mathcal{M}$. For $g \in L^p(\mathcal{M})$, we define
\begin{equation*}
\|g\|_{L^p(\mathcal{M})}
=
\left(
\int_{\mathcal{M}} |g(x)|^p \, d\mathrm{vol}(x)
\right)^{1/p}.
\end{equation*}

Consider the spectral decomposition
\[
-\Delta \varphi_k = \lambda_k \varphi_k, \quad k \geq 1,
\]
where
\[
0 = \lambda_1 \leq \lambda_2 \leq \lambda_3 \leq \cdots,
\]
and $\{\varphi_k\}_{k \geq 1}$ forms an orthonormal basis of $L^2(\mathcal{M})$.

Let $g : \mathcal{M} \to \mathbb{R}$ be a smooth function. Then
\[
g = \sum_{k=1}^{\infty} c_k \varphi_k,
\quad
c_k = \langle g, \varphi_k \rangle_{L^2(\mathcal{M})}.
\]

It follows that
\[
\|\nabla g\|_{L^2(\mathcal{M})}^2
=
\sum_{k=1}^{\infty} \lambda_k |c_k|^2.
\]

This identity shows that high-frequency components (corresponding to large eigenvalues $\lambda_k$) contribute more significantly to the gradient energy. In particular, for each $k \geq 2$, we have
\begin{equation}
\label{eq:grad_energy_bound}
|c_k|^2
\leq
\frac{1}{\lambda_k}
\|\nabla g\|_{L^2(\mathcal{M})}^2.
\end{equation}

Thus, high-frequency coefficients are controlled by the gradient energy. In particular, minimizing $\|\nabla g\|_{L^2}$ suppresses high-frequency modes.

\medskip

We now show how the proposed methodology enforces this behavior. Let
\[
X = \{x_1, x_2, \dots, x_N\} \subset \mathcal{M}
\]
be a collection of i.i.d.\ samples drawn from a smooth, strictly positive probability density $\rho$ on $\mathcal{M}$. Using Algorithm~\ref{alg:datadriven_interpolant}, we construct an interpolant
\[
\overline{g}_{x_1,\dots,x_N}
\]
from the values $g(x_1), \dots, g(x_N)$.

According to Theorem~\ref{thm:local_decay}, this interpolant satisfies:

\begin{itemize}
    \item[(i)] \textbf{Interpolation property:}
    \[
    \overline{g}(x_i) = g(x_i), \quad i = 1, \dots, N.
    \]
    
    \item[(ii)] \textbf{Vanishing gradient at sample points:}
    \[
    \nabla \overline{g}(x_i) = 0, \quad i = 1, \dots, N.
    \]
\end{itemize}

Hence, $\overline{g}_{x_1,\dots,x_N}$ interpolates $g$ exactly at the sampled points while locally flattening the function.

However, the fact that the gradient vanishes on $X$ does not immediately imply global control of the gradient over $\mathcal{M}$. The next result shows that, as the number of samples increases, the global $L^p$-norm of the gradient becomes small with high probability.

\begin{thm}
\label{teoremafiltrofreq}
Let $\mathcal{M} \subset \mathbb{R}^n$ be a $d$-dimensional compact manifold without boundary. Let $\rho:\mathcal{M} \to \mathbb{R}$ be a smooth, strictly positive probability density, and denote by $\mathbb{P}$ the induced probability measure.

Let $g:\mathcal{M} \to \mathbb{R}$ be a smooth function, and let $\overline{g}_{x_1,\dots,x_N}$ be the interpolant constructed via Algorithm~\ref{alg:datadriven_interpolant}.

Then there exist constants $M>0$ and $L_1>0$ such that for every sufficiently small $\varepsilon > 0$, there exists $s(\varepsilon)>0$ satisfying
\begin{equation*}
\mathbb{P}\!\left(
\|\nabla \overline{g}_{x_1,\dots,x_N}\|_{L^p(\mathcal{M})}^p
\leq
M \, \mathrm{Vol}(\mathcal{M}) \, \varepsilon
\right)
\geq
1 - s(\varepsilon)\bigl(1 - L_1 \varepsilon^d\bigr)^N.
\end{equation*}
\end{thm}

The proof is provided in Appendix~\ref{pruebacotareduc}.

\medskip

For fixed $\varepsilon > 0$, the term $s(\varepsilon)(1 - L_1 \varepsilon^d)^N$ converges to zero as $N \to \infty$. Therefore, with high probability, the gradient energy $\|\nabla \overline{g}_{x_1,\dots,x_N}\|_{L^p(\mathcal{M})}^p$ becomes arbitrarily small as the number of samples increases.

Combining this result with inequality~\eqref{eq:grad_energy_bound}, we conclude that the high-frequency coefficients of $\overline{g}_{x_1,\dots,x_N}$ are suppressed with high probability as $N$ increases. In other words, the proposed method acts as a data-driven low-pass filter, attenuating high-frequency oscillations while preserving interpolation constraints.

\section{Applications}
\label{sec:applications}

In this section, we present applications of the methodology introduced in Section~\ref{sec:interpol}. 
The first application concerns numerical interpolation, while the second addresses computational tomography reconstruction from sparse projection data.

Before presenting these applications, we make an important remark regarding the numerical implementation. 
To accurately capture the exponential decay near the sample points $x_i$ and to ensure sufficient smoothness, it is not enough to evaluate Algorithm~\ref{alg:datadriven_interpolant} only at the interpolation nodes. 
Instead, the algorithm must also be evaluated at points lying in sufficiently small neighborhoods of each $x_i$. 
In practice, this is achieved by generating additional evaluation points around the data. 
This procedure improves the resolution of the local structure of the interpolant and prevents spurious oscillations near the interpolation points. 
This strategy is systematically adopted in all numerical experiments presented below.

\subsection{Data-Driven Interpolation of High-Frequency Signals}

The first application concerns the data-driven interpolation of a \blue{high-frequency smooth function} defined on a smooth manifold.
More precisely, let $x_1, x_2, \ldots, x_N \in \mathcal{M}$ be a set of sample points on a smooth manifold $\mathcal{M}$, and let $g(x_1), g(x_2), \ldots, g(x_N)$ denote the corresponding evaluations of a \blue{high-frequency smooth function} $g : \mathcal{M} \to \mathbb{R}$.
Given a point $x \in \mathcal{M}$ sufficiently close to a sample point $x_i$, for some $1 \leq i \leq N$, Algorithm~\ref{alg:datadriven_interpolant} provides an approximation of $g(x)$.
An important feature of this interpolation scheme is that it satisfies several desirable properties established in Section~\ref{sec:interpol}.
In particular, the approximation interpolates the data exactly, i.e., it reproduces the values $g(x_i)$ at the sample points.
Moreover, its gradient vanishes at each point $x_i$ and exhibits exponential decay in a neighborhood of these points.
\blue{More precisely, there exists a positive constant $M$ such that, for every
$x \in B(x_i,r_i/2)$,

\[
\|\nabla \bar{g}_{x_1,\dots,x_N}(x)\|
\le
\frac{M}{\|x-x_i\|^{4}}
\exp\!\left(
-\frac{r_i^{2}}{4\|x-x_i\|^{2}}
\right),
\]

where

\[
r_i:=\min_{j\neq i}\|x_i-x_j\|.
\]

Since the exponential term dominates every algebraic power as
$\|x-x_i\|\to 0$, it follows that, for every positive integer $m$,

\[
\|\nabla \bar{g}_{x_1,\dots,x_N}(x)\|
=
O\!\left(\|x-x_i\|^{m}\right).
\]

In other words, the norm of the gradient vanishes faster than any polynomial
of the form $\|x-x_i\|^{m}$ as $x\to x_i$.}

These properties ensure that the interpolant is smooth and locally stable, in the sense that it does not exhibit large variations near the sampling locations.
\blue{We also remark that these properties allow our algorithm to convert high frequencies into low frequencies within a neighborhood of each sample point, and it is precisely this local frequency conversion that makes the interpolation procedure efficient. This is desirable because most data-driven interpolation methods suffer from an inherent bias toward low frequencies, typically capturing only the low-frequency content of a signal while failing to resolve its high-frequency components. In the context of neural networks, this phenomenon is known as \emph{spectral bias} \cite{rahaman2019spectral}, and an analogous effect has been observed for Gaussian processes, which likewise tend to capture low frequencies while neglecting high ones \cite{wilson2013gaussian}. By converting high frequencies into low frequencies within a neighborhood of each sample point, our algorithm avoids this limitation and allows for a more accurate interpolation.}

\subsection{Angular Regularization for Sparse-Angle Tomographic Projections}
\label{subsec:angular_regularization}

As a second application of the methodology introduced in Section~\ref{sec:interpol}, we consider the sparse-angle tomography problem. In this context, a small quantity of sparse-angle setups is used to reduce data acquisition time, radiation exposure, or cost. However, this sparsity makes the reconstruction problem highly ill-posed and may induce spurious oscillations in the angular variable, thereby degrading the quality of the reconstructed image.

To address this issue, we impose angular stationarity at a selected set of sampling angles. This provides an effective regularization mechanism that suppresses oscillations while remaining consistent with the measured data. 

The objective of this section is to construct a stabilized reconstruction whose angular gradient is small at the prescribed sampling angles. Importantly, this procedure is intended as a preprocessing step. The main contribution of this work is a strategy to smooth the tomographic projection data with respect to the angular variable prior to reconstruction.

After this preprocessing, the final image can be reconstructed using standard methods, such as the classical filtered back-projection (FBP), or used as an initial estimate in an iterative reconstruction scheme, for instance in total variation regularization methods.

We now explain the methodology in details. Let $p(\theta_k,s)$ denote the measured tomography projection data, where $\theta_k \in \Theta := [0,2\pi)$ and $s \in S \subset \mathbb{R}$. To capture the local smoothness of the approximation, we introduce an auxiliary set of angular points $\{\hat{\theta}_\ell\}_{\ell=1}^{\hat N}$ distinct from the sampling angles $\{\theta_k\}_{k=1}^N$, and compute an approximation $\tilde p(\hat{\theta}_\ell,s)$ by applying Algorithm~\ref{alg:datadriven_interpolant}.  Finally, using the new data-set $\tilde p(\hat{\theta}_\ell,s)$, we reconstruct the tomography image using the filtered back-projection method.

Thus, the proposed procedure consists of two main steps:
\begin{enumerate}
    \item Generate additional tomographic projections at new angular positions using the proposed method.
    \item Apply filtered back projection (FBP) to the augmented dataset to obtain an initial estimate, which is then used as the starting point for an iterative scheme that solves a variational regularization problem with a total variation penalty.
\end{enumerate}

We emphasize that the proposed procedure is closely related to the variational formulation introduced in Equation~\eqref{regindugeral} of Section~\ref{sec:interpol}, and therefore induces an implicit regularization on the data. 
More precisely, the proposed methodology can be interpreted as the numerical solution of the following variational problem:
\begin{equation}
\label{eq:variaciontomo}
\min_{q} \; 
\| q - p \|_{L^2(\Theta \times S)}^2 
\;+\; 
\lambda \sum_{k=1}^N \| \partial_\theta q(\theta_k,\cdot) \|_{L^1(S)},
\end{equation}
where $\lambda > 0$ is a regularization parameter. 
The first term enforces fidelity to the measured data, while the second promotes angular regularity by penalizing variations in the angular direction at the sampled angles. 
In particular, this term encourages sparsity of the angular derivative, thereby reducing spurious oscillations caused by noise and discrete angular sampling.

The regularized projection $\tilde p$ is then used as input for classical reconstruction methods, yielding a stabilized tomographic reconstruction. The overall procedure is summarized in Algorithm~\ref{algoreconstructiontomografia}.

\paragraph{Relation to compressed sensing.}
We emphasize that the proposed methodology is closely related to ideas from compressed sensing and total variation regularization. In particular, as discussed at the end of Section~\ref{sec:interpol}, the method can be interpreted within the compressed sensing framework.

From this perspective, the variational problem~\eqref{eq:variaciontomo} aims to recover analytically the function $p(\cdot,s)$, which represents the tomographic projection data across all rotation angles, from a sparse set of measured projections. The key assumption is that the angular derivative of the projection data is small at the sampling angles, which can be interpreted as a sparsity prior in the angular domain. Therefore, problem~\eqref{eq:variaciontomo} promotes solutions that are sparse with respect to the representation induced by the angular derivative operator.

This viewpoint is analogous to classical compressed sensing approaches, where sparsity in a suitable transform domain (e.g., wavelets or gradient-based representations) is exploited to recover signals from incomplete or noisy measurements. In the present setting, the sparsity structure is enforced locally at the acquisition angles, leading to a regularized reconstruction that is robust under sparse-view sampling and noise contamination.

The main advantage of the proposed methodology compared to standard compressed sensing and total variation approaches is its computational efficiency. In contrast to these methods, which typically require solving large-scale optimization problems, our approach yields a closed-form analytical reconstruction. This is achieved by exploiting the geometric structure induced by the set of tomographic projection data, thereby significantly reducing the computational cost.

\begin{algorithm}[H]
\caption{Angular Regularization of Tomographic Projection Data}
\label{algoreconstructiontomografia}
\begin{flushleft}

\textbf{Input:}
\begin{itemize}
    \item Measured projection data $p(\theta,s)$.
    \item Sampling angles $\{\theta_k\}_{k=1}^N \subset \Theta$.
\end{itemize}

\textbf{Procedure:}
\begin{enumerate}
    \item Generate an auxiliary set of angular points 
    $\{\hat{\theta}_m\}_{m=1}^M \subset \Theta$.

    \item For each fixed $s \in S$, compute the regularized approximation 
    $\tilde p(\cdot,s)$ by applying Algorithm~\ref{alg:datadriven_interpolant} to the function 
    $p(\cdot,s)$ with respect to the angular variable, and evaluate the resulting approximation at the points 
    $\{\hat{\theta}_m\}_{m=1}^M$.

    \item Compute the tomographic reconstruction by applying a classical reconstruction method 
    (e.g., filtered back-projection or total variation regularization) to the regularized projection data 
    $\tilde p(\hat{\theta}_m,s)$.
\end{enumerate}

\textbf{Output:} Reconstructed tomographic image obtained from the regularized projection data 
$\{\tilde p(\hat{\theta}_m,s)\}_{m=1}^M$.

\end{flushleft}
\end{algorithm}

\section{Numerical Experiments}
\label{sec:numeexp}

In this section, we present two numerical experiments to validate and illustrate the proposed methodology.

\blue{The first experiment considers a synthetic example on a two-dimensional torus parameterized by the domain $[0,1]\times[0,1]\subset\mathbb{R}^2$. The objective is to assess and visualize the performance of the proposed data-driven interpolation method in reconstructing a high-frequency signal, while also evaluating its ability to capture the underlying geometric structure of the data. To benchmark its performance, we compare the proposed approach against several established interpolation and learning methods, including a fully connected feedforward neural network, Gaussian process regression, and radial basis function networks.}

The second experiment concerns the sparse-view CT tomography problem, following the methodology developed in Section~\ref{sec:applications}. In this setting, only a limited number of angular projections are available, and the goal is to reconstruct the image using the angular regularization introduced in Section~\ref{subsec:angular_regularization}. This problem is a classical ill-posed inverse problem.

All experiments employ Algorithm~\ref{alg:datadriven_interpolant}. The numerical implementation was carried out in \textsc{Matlab R2017b} on a laptop equipped with an Intel Core i5-1235U 1.30~GHz processor and 8~GB of RAM.

+

\subsection{Synthetic Experiment on the Square \texorpdfstring{$[0,1] \times [0,1]$}{[0,1] x [0,1]}}

To illustrate the behavior of the proposed method in a low-dimensional setting, we consider the two-dimensional domain $[0,1] \times [0,1]$, which we denote by $\mathcal{M}$.

We define a real-valued function $f: \mathcal{M} \to \mathbb{R}$ by
\begin{equation}
    f(x,y) = \sin(10\pi x)\cos(10\pi y).
    \label{eq:test_function}
\end{equation}

\blue{which is smooth ($f \in C^\infty(\mathcal{M})$) and completes five full oscillation periods along each coordinate direction over $\mathcal{M}$, corresponding to a wavelength of $1/5$ in both $x$ and $y$.}

\begin{figure}[htp]
\centering
\includegraphics[width=0.7\textwidth]{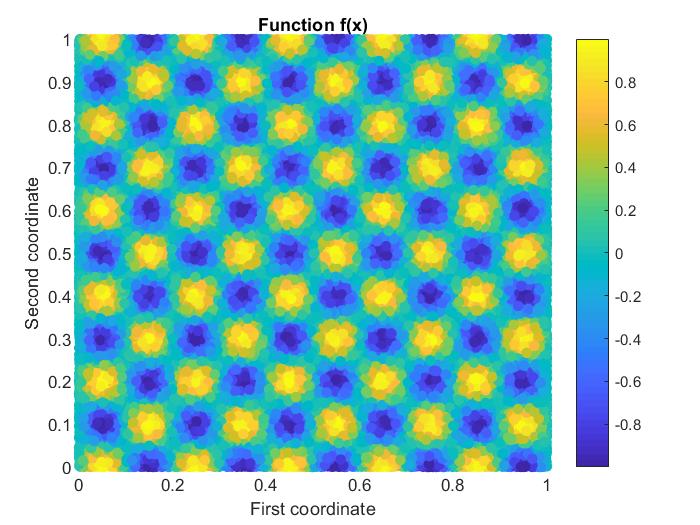}
\caption{The domain $\mathcal{M} = [0,1]^2$ and the function $f(x,y) = \sin(10\pi x)\cos(10\pi y)$.}
\label{fig:cube}
\end{figure}

Figure~\ref{fig:cube} displays both the domain $\mathcal{M}$ and the function $f$. The objective of this experiment is to interpolate $f$ using the proposed methodology and to compare its performance with that of alternative methods. The function is characterized by relatively high-frequency oscillations, which may pose challenges for interpolation methods, particularly when the sampling is limited. we compare the proposed method against several data-driven interpolation methods, including a feedforward neural network with $10$ hidden layers, Gaussian process regression, and a radial basis function network. The computational implementation of the experiment can be found in the GitHub repository~\cite{codigoTV}.

To assess the performance of the various interpolation schemes, we evaluate their accuracy in approximating the function $f$ over the domain $\mathcal{M} \subset [0,1]^2$. Training sets of varying sizes, $n \in \{100, 500, 1000, 2000\}$, are constructed by sampling points i.i.d.\ from a uniform distribution over $\mathcal{M}$. Additionally, a fixed out-of-sample test set consisting of $10^4$ points is generated uniformly over the unit cube.

\begin{figure}[htbp]
\centering

\begin{minipage}[t]{1\textwidth}
    \centering
    \includegraphics[width=\textwidth]{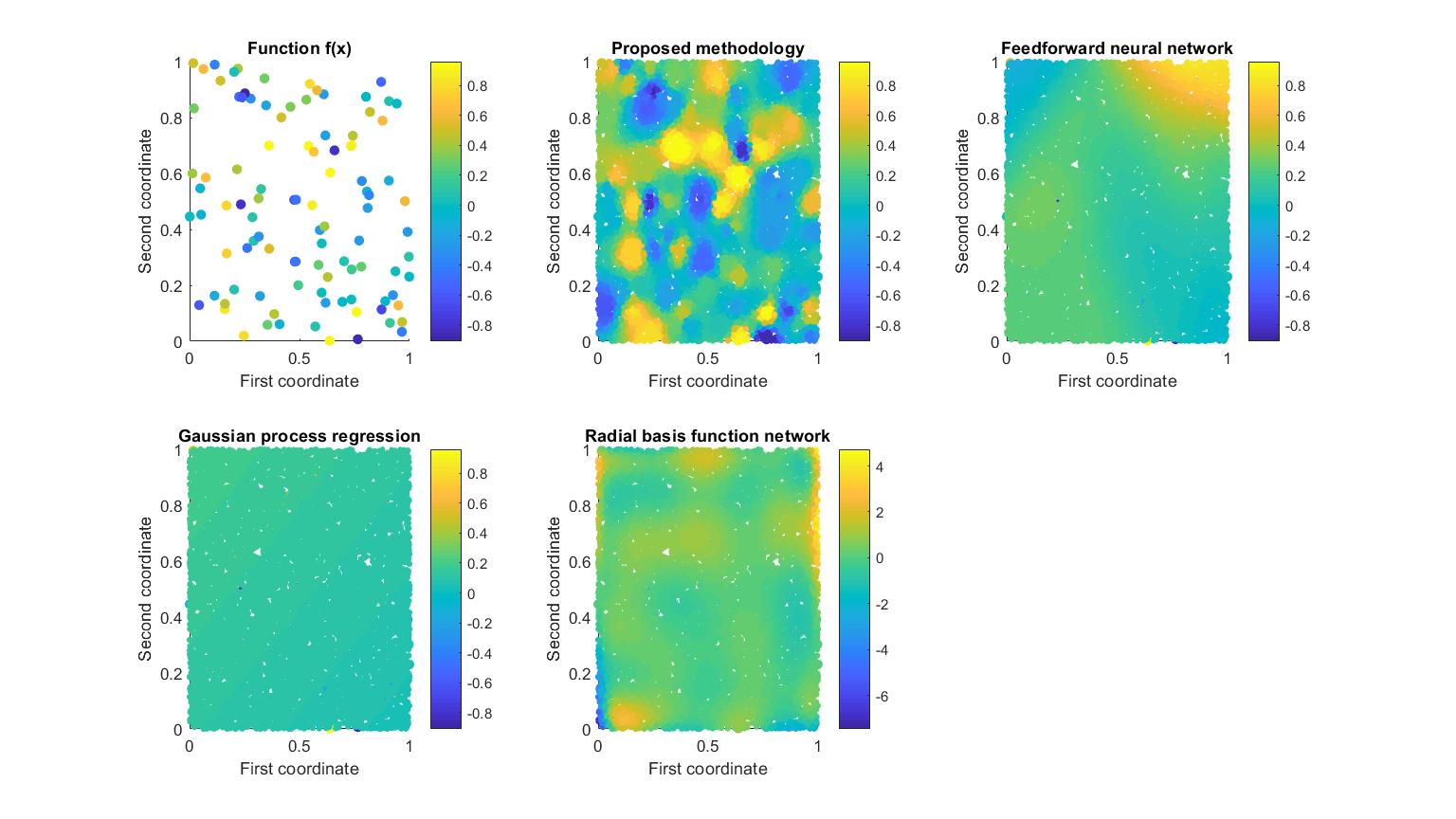}
    
    \smallskip
    \small (a) Interpolation with 100 training samples.
\end{minipage}

\vspace{0.5cm}

\begin{minipage}[t]{1\textwidth}
    \centering
    \includegraphics[width=\textwidth]{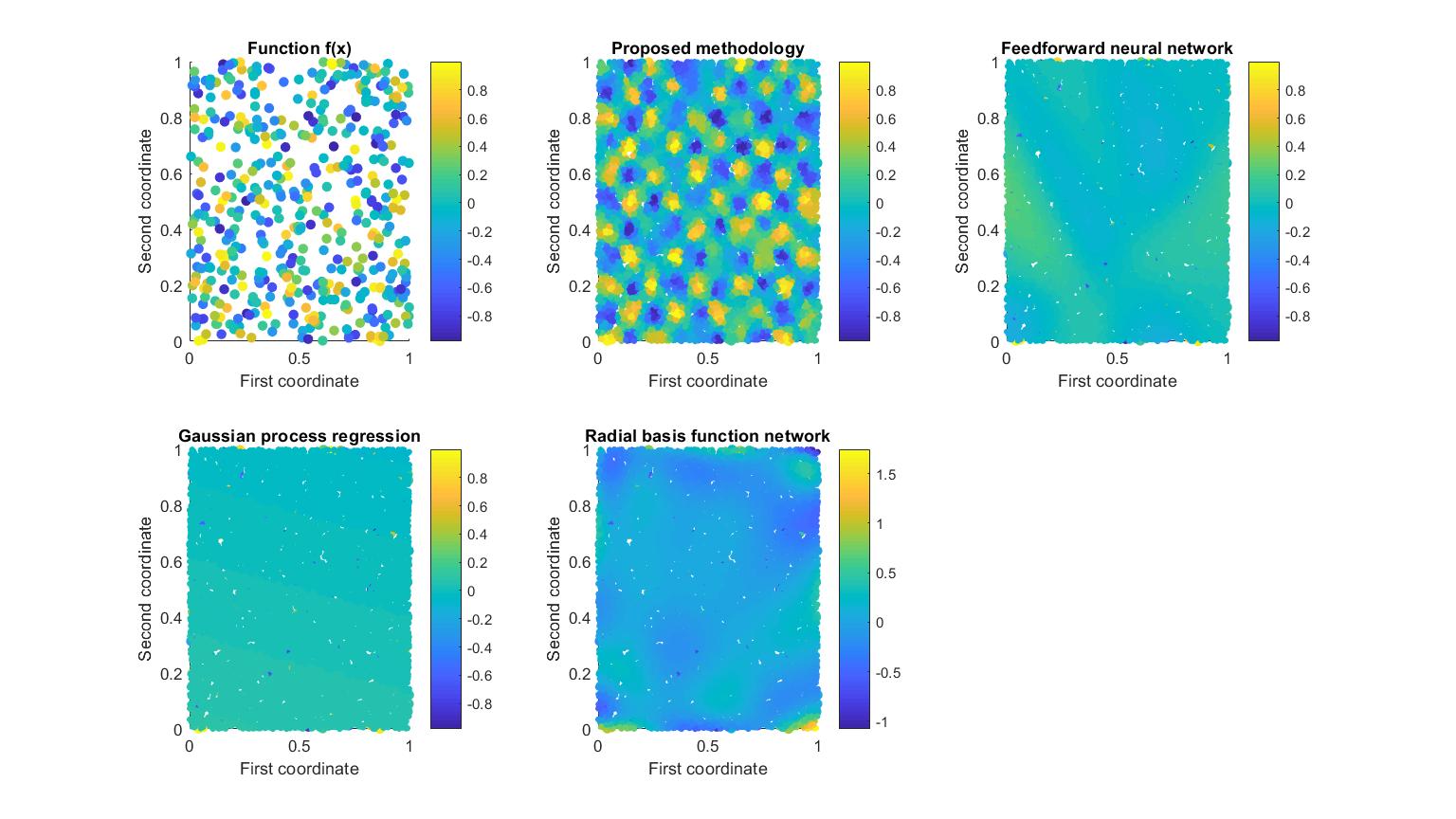}
    
    \smallskip
    \small (b) Interpolation with 500 training samples.
\end{minipage}

\caption{
Interpolation results over the domain $[0,1]^2$ using different data-driven methods.
Each block contains five subplots. The top block corresponds to training with 100 samples,
and the bottom block to 500 samples. In both cases, $10^4$ out-of-sample points are used
for evaluation. The panel labeled $f(x)$ shows the training data, while the others display
the predictions of the proposed method, a feedforward neural network, Gaussian process
regression, and a radial basis function network.
}
\label{SE15}

\end{figure}

\begin{figure}[htbp]
\centering

\begin{minipage}[t]{1\textwidth}
    \centering
    \includegraphics[width=\textwidth]{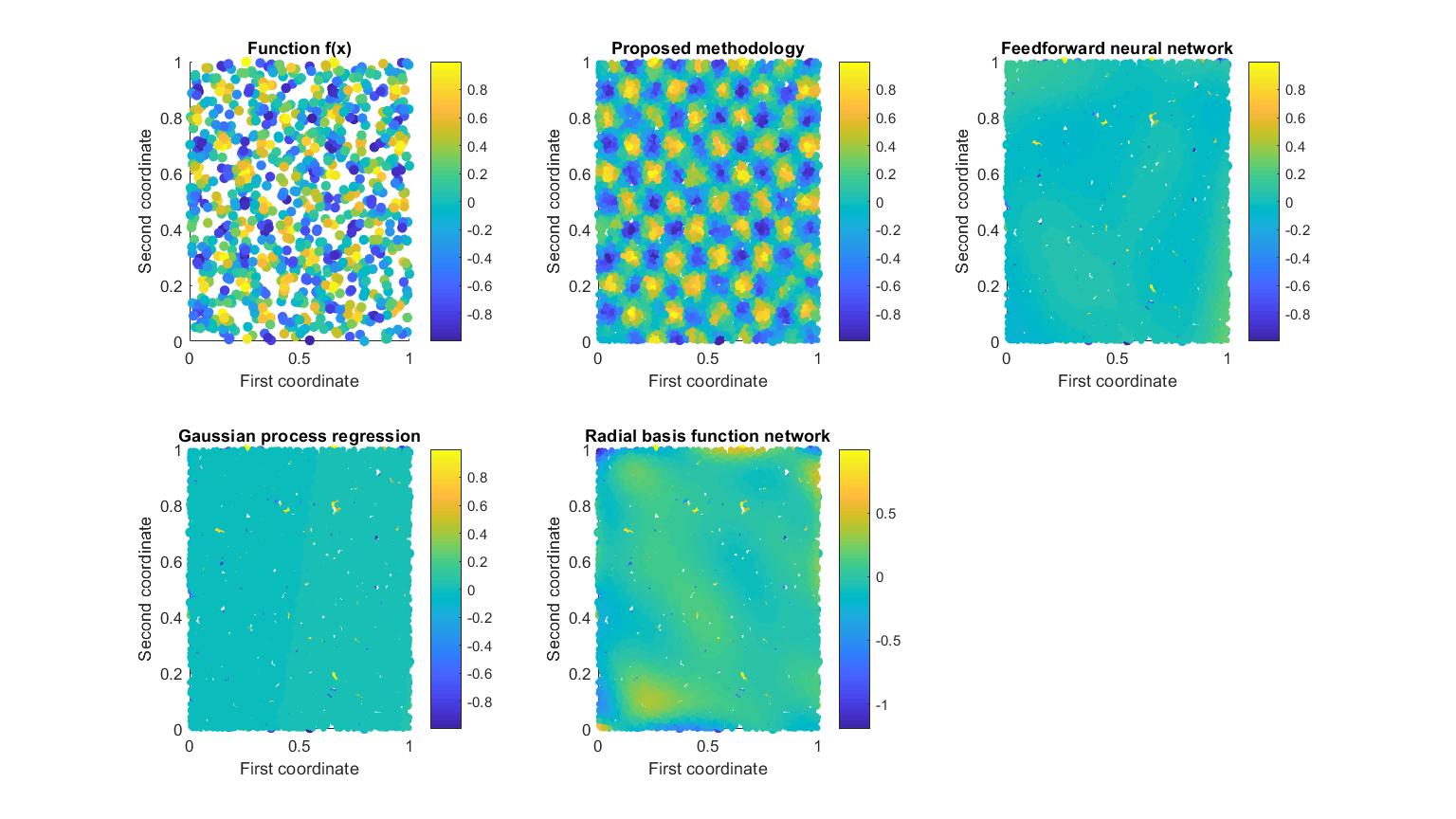}
    
    \smallskip
    \small (a) Interpolation with 1000 training samples.
\end{minipage}

\vspace{0.5cm}

\begin{minipage}[t]{1\textwidth}
    \centering
    \includegraphics[width=\textwidth]{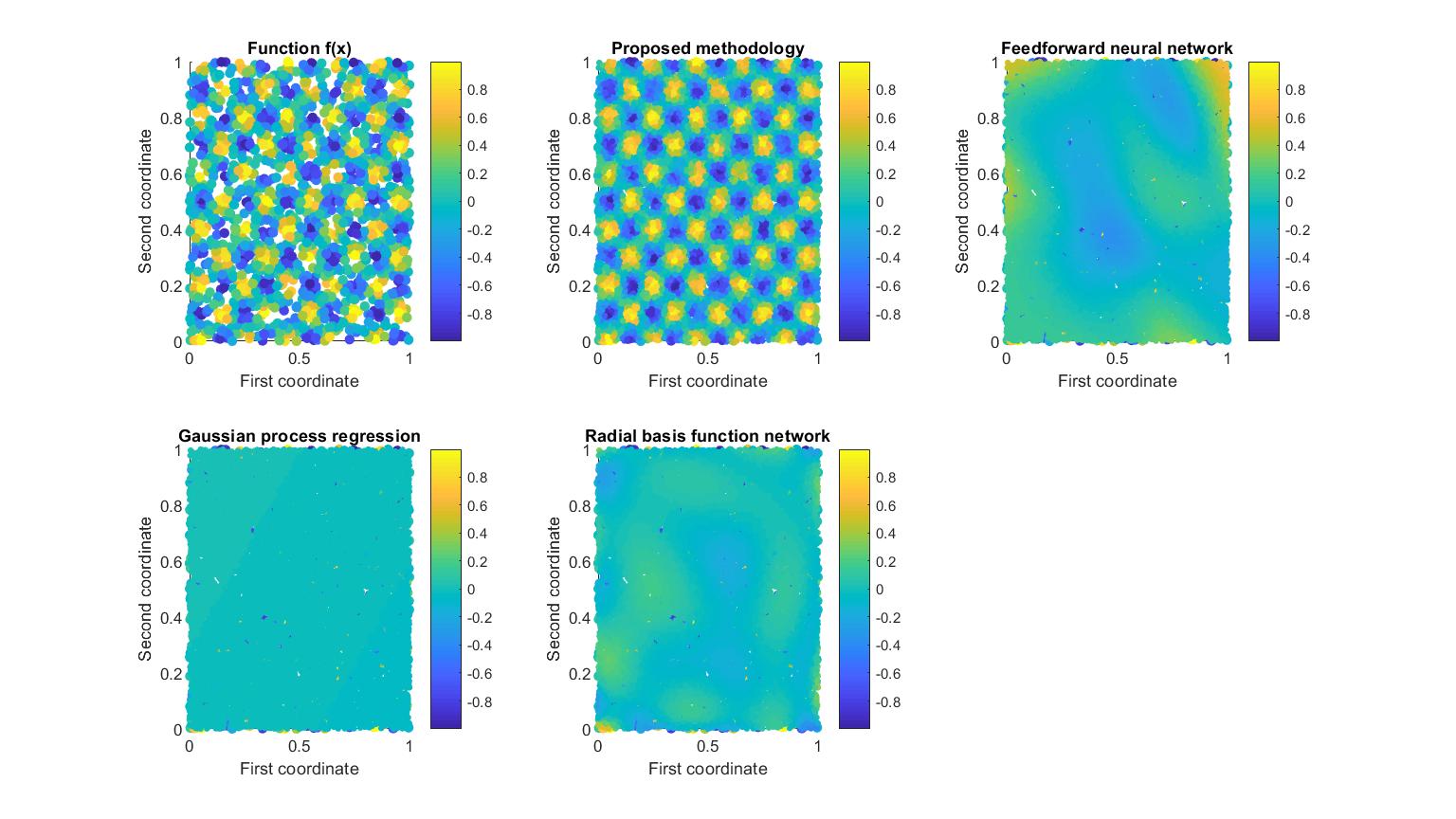}
    
    \smallskip
    \small (b) Interpolation with 2000 training samples.
\end{minipage}

\caption{
Interpolation results over the domain $[0,1]^2$ using different data-driven methods.
Each block contains five subplots. The top block corresponds to training with 1000 samples,
and the bottom block to 2000 samples. In both cases, $10^4$ out-of-sample points are used
for evaluation. The panel labeled $f(x)$ shows the training data, while the others display
the predictions of the proposed method, a feedforward neural network, Gaussian process
regression, and a radial basis function network.
}
\label{SE120}

\end{figure}

Qualitative results are illustrated in Figures~\ref{SE15} and \ref{SE120}. Figure~\ref{SE15} displays reconstructions for $n=100$ and $n=500$, while Figure~\ref{SE120} covers the $n=1000$ and $n=2000$ cases. Visually, the proposed method provides the most faithful reconstruction of $f$. In contrast, competing benchmarks fail to capture the underlying functional structure, resulting in interpolations that deviate significantly from the ground truth.

For a quantitative comparison, we report the Mean Squared Error (MSE) and the total computational time (comprising both training and inference phases). Table~\ref{tab:time_combined_SE} summarizes these metrics, where the MSE is reported as the Frobenius norm of the difference between true and predicted values over the test set. 

The results show that the proposed method consistently attains the lowest error across all sample sizes, outperforming feedforward neural networks (FNN), Gaussian process regression (GPR), and radial basis function networks (RBFN). In addition, it exhibits superior computational efficiency. 

\blue{The comparatively lower performance of the competing methods can be explained by well-documented limitations in approximating highly oscillatory functions. Neural network-based approaches, including FNN and RBFN, are affected by the \emph{spectral bias} phenomenon, whereby low-frequency components are learned significantly faster than high-frequency ones, making the accurate interpolation of rapidly oscillating signals particularly challenging~\cite{rahaman2019spectral}. Likewise, conventional Gaussian process regression (GPR) models typically employ smooth covariance kernels that favor low-frequency functions and therefore struggle to accurately represent highly oscillatory behavior unless specifically designed spectral kernels are incorporated~\cite{wilson2013gaussian}. Since the target function considered in this experiment contains substantial high-frequency content, these limitations provide a plausible explanation for the observed performance gap between the proposed method and the competing approaches.}

These results indicate that the proposed approach achieves an effective trade-off between accuracy and scalability, particularly in settings where the preservation of latent geometric structure is crucial. This empirical behavior is consistent with the theoretical results established in Section~\ref{sec:interpol}.

In summary, the proposed method attains high interpolation accuracy while maintaining low computational cost, making it well suited for large-scale applications that require both efficiency and the faithful representation of underlying geometric features.

\begin{table}[H]
\centering
\begin{tabular}{llcccc} 
\toprule
\textbf{Metric} & \textbf{Batch Size} & \textbf{Proposed} & \textbf{FNN} & \textbf{GPR} & \textbf{RBFN} \\ 
\midrule
\multirow{4}{*}{Mean squared error} 
 & 100  & 56.74 & 56.15 & 51.01 & 87.40 \\ 
 & 500  & 31.11 & 50.51 & 50.03 & 52.45 \\ 
 & 1000 & 23.03 & 50.22 & 49.98 & 50.96 \\ 
 & 2000 & 14.66 & 52.95 & 50.25 & 50.45 \\ 
\midrule
\multirow{4}{*}{Computational time} 
 & 100  & 0.02 & 0.49 & 0.10 & 0.83 \\ 
 & 500  & 0.18 & 0.44 & 0.60 & 10.68 \\
 & 1000 & 0.30 & 0.60 & 2.53 & 69.17 \\
 & 2000 & 0.53 & 0.57 & 13.49 & 1436.16 \\
\bottomrule
\end{tabular}
\caption{Mean squared interpolation error and computational time (in seconds) required for interpolation, including both training and testing phases, across all methods. Results are reported for different batch sizes used in training the proposed method. We compare the proposed approach against a feed-forward neural network (FNN), Gaussian process regression (GPR), and a radial basis function network (RBFN). Performance is evaluated out-of-sample over the domain $[0,1]^2$.}
\label{tab:time_combined_SE}
\end{table}

\subsection{Sparse-View Tomographic Reconstruction}

\begin{figure}[htp]
\centering
\includegraphics[width=0.3\textwidth]{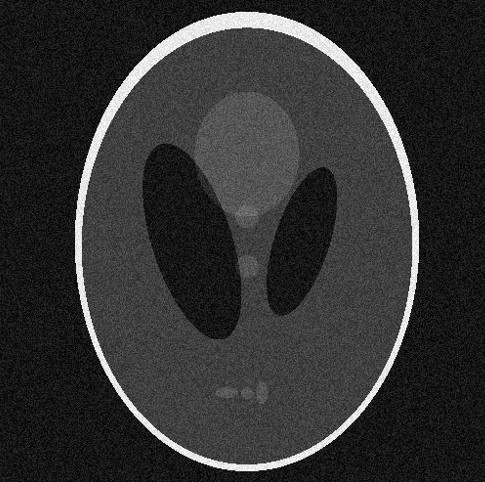}
\includegraphics[width=0.3\textwidth]{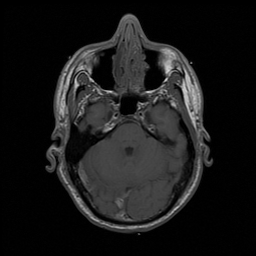}
\includegraphics[width=0.3\textwidth]{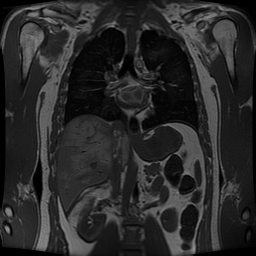}
\caption{From left to right: Shepp-Logan phantom, human head, and human abdomen. The human images were taken from~\cite{imagesrad}. }
\label{ctfigures}
\end{figure}

In the following numerical experiments, we evaluate the methodology proposed in Section~\ref{subsec:angular_regularization} by applying Algorithm~\ref{algoreconstructiontomografia} to three images: the Shepp--Logan phantom, a human head, and a human abdomen. The human images were obtained from a public dataset provided by the National Library of Medicine, as reported in Ref.~\cite{imagesrad}. These images are shown in Fig.~\ref{ctfigures}.

In the following experiments, we use the parameter settings described below.
The methodology is tested with batch sizes of $100$ and $300$, corresponding 
to sparse angular positions uniformly distributed in $[0,180]$, at which the 
tomographic projections are acquired. We refer to this set of projections as the 
\emph{training dataset}.

The choice of these sample sizes is motivated by previous numerical studies, 
which indicate that approximately $720$ projections are required to achieve 
high-accuracy tomographic reconstruction \cite{xuanhao2022effect}. 
Our goal, however, is to obtain high-resolution reconstructions using a 
significantly smaller number of projections (i.e., fewer than $720$).

An additional batch of $10^4$ angles is randomly generated from the uniform 
distribution on $[0,180]$. We refer to this set of projections as the 
\emph{learned dataset}. 

In the proposed model, the reconstruction problem is analyzed under additive noise with magnitude $20\%$. Let $X \in \mathbb{R}^{d \times d}$ denote the image to be reconstructed. The corresponding noisy observation is defined as
\begin{equation*}
    X_{\mathrm{noisy}} = X + 0.2\, Z,
\end{equation*}
where $Z \in \mathbb{R}^{d \times d}$ is a random matrix whose entries are independent and identically distributed according to the uniform distribution on $[0,1]$.

We compare the proposed methodology with several baseline approaches, including 
a feed-forward neural network trained via gradient descent, cubic spline 
interpolation, Gaussian process regression, and a radial basis function network. 
These methods are selected to provide a comprehensive comparison with our approach.

The neural network consists of $10$ hidden layers. In all comparative algorithms, 
the projection angle is taken as input and the corresponding projection is 
produced as output. Training is performed on the \emph{training dataset}, while 
inference on the \emph{learned dataset} is used to generate sinograms for all 
methods under comparison.

The resulting sinograms are then used to reconstruct the tomographic image 
using the Filtered Back Projection (FBP) algorithm. All numerical implementations 
were carried out in \textsc{Matlab}, and the code is publicly available in the 
GitHub repository~\cite{codigoCT}.

\begin{figure}[H]
\centering
\includegraphics[width=1\textwidth]{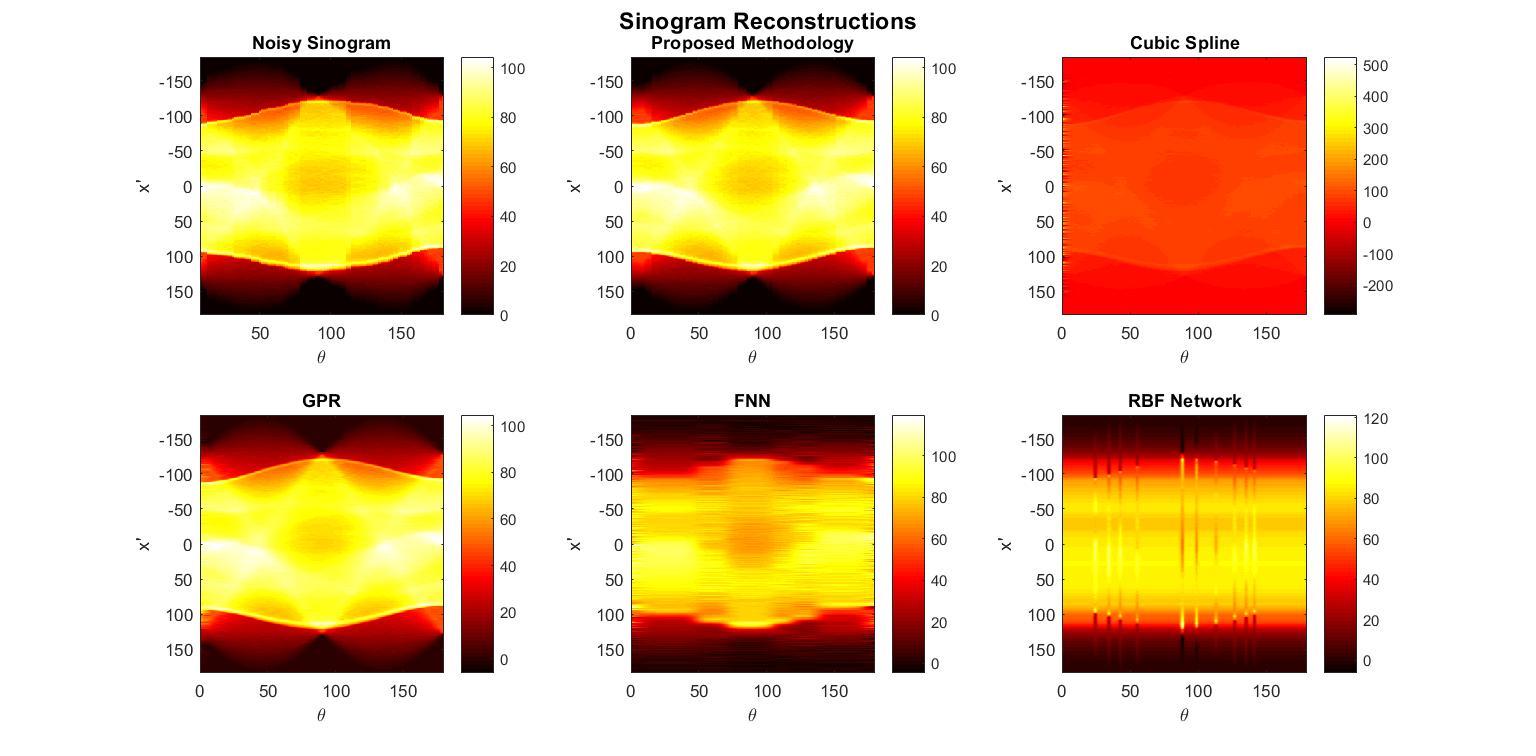}
\includegraphics[width=1\textwidth]{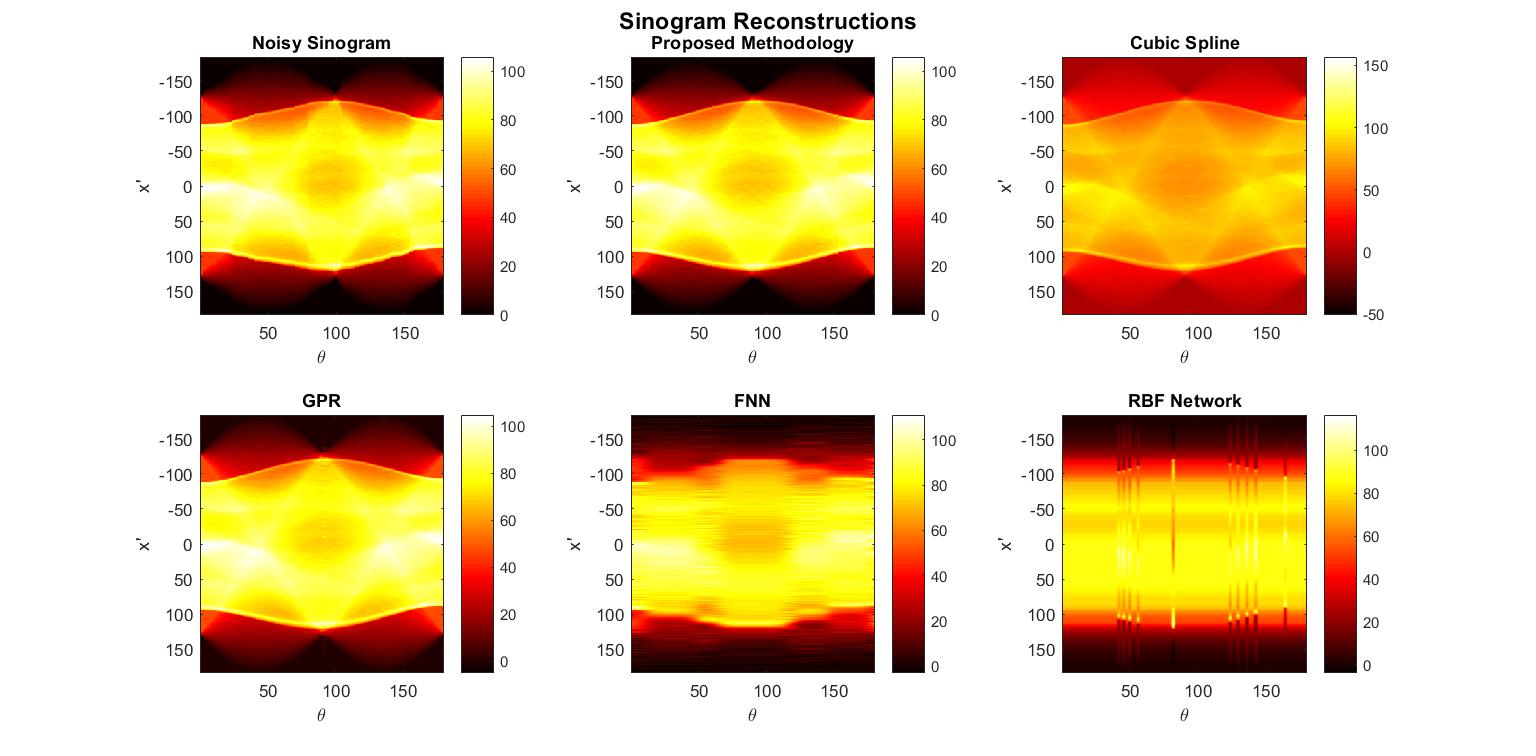}
\caption{Sinograms of tomographic projections of the Shepp--Logan phantom reconstructed using different interpolation methods. The first two rows correspond to $100$ angular samples, while the last two rows correspond to $300$ angular samples.
From left to right, the panels represent: \emph{Noisy sinogram} (noisy sparse projections from the training set), \emph{Proposed method}, \emph{Cubic spline} interpolation, \emph{Gaussian process regression (GPR)}, \emph{Feed-forward neural network (FNN)}, and \emph{Radial basis function (RBF)} network.}
\label{fig:sp}
\end{figure}

\begin{figure}[H]
\centering
\includegraphics[width=\textwidth]{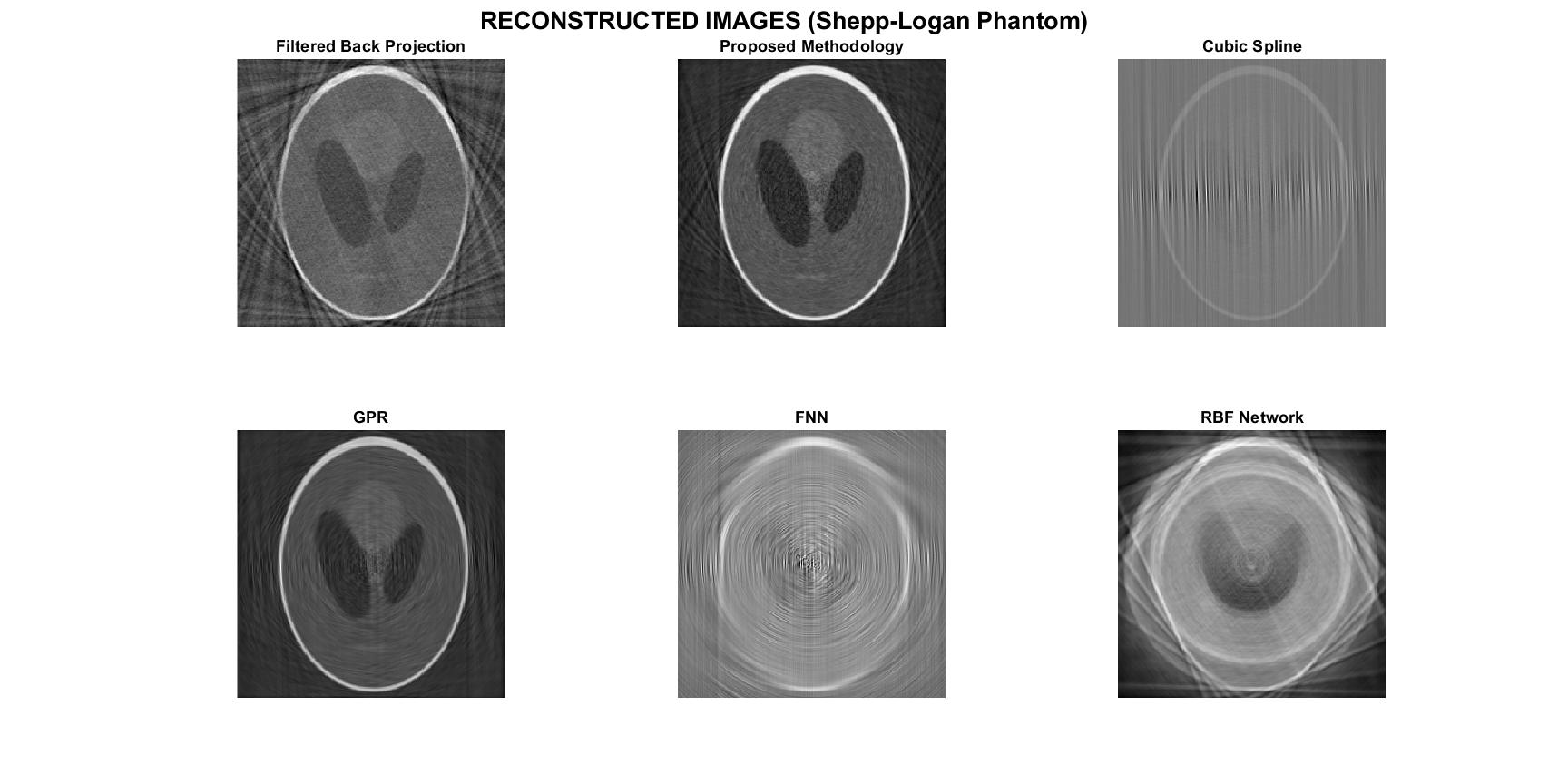}
\includegraphics[width=\textwidth]{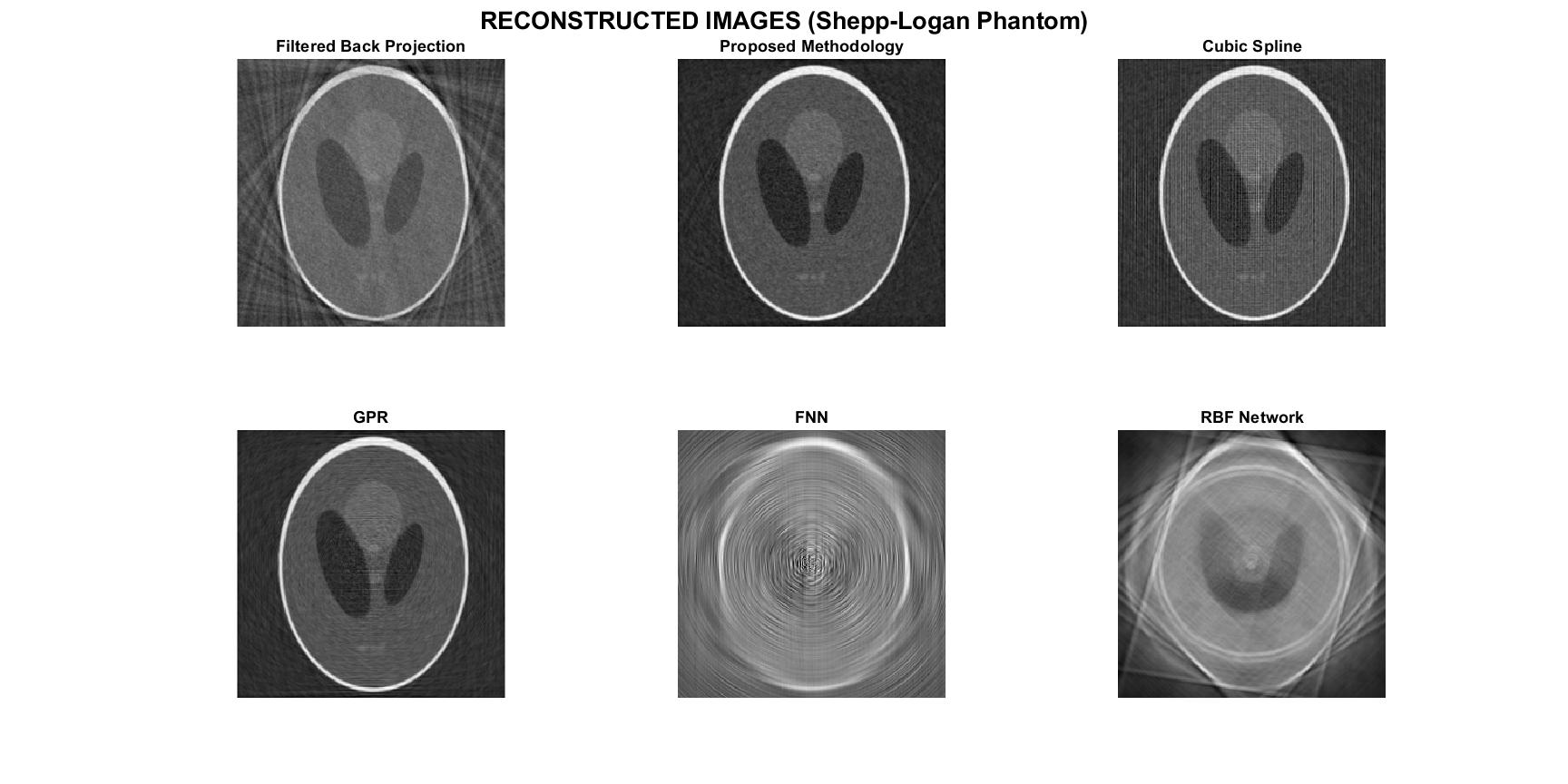}
\caption{Tomographic reconstructions of the Shepp--Logan phantom obtained using different interpolation methods. The first two rows correspond to $100$ angular samples, while the last two rows correspond to $300$ angular samples.
From left to right, the panels show: \emph{Filtered back projection} (tomographic reconstruction obtained by directly applying filtered back projection to the noisy sparse projections from the training set), \emph{Proposed method}, \emph{Cubic spline} interpolation, \emph{Gaussian process regression (GPR)}, \emph{Feed-forward neural network (FNN)}, and \emph{Radial basis function (RBF)} network.}
\label{fig:rp}
\end{figure}

\begin{figure}[H]
\centering
\includegraphics[width=1\textwidth]{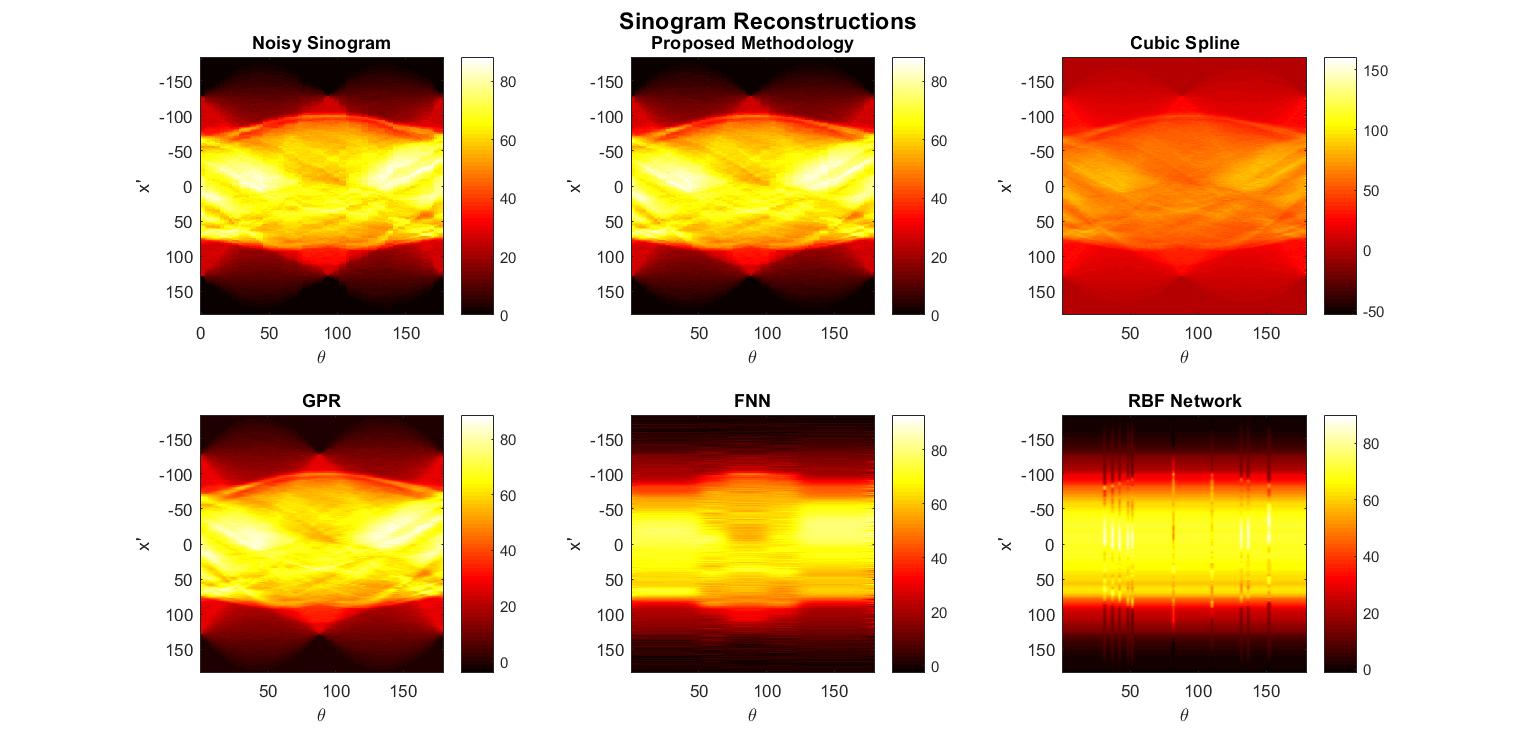}
\includegraphics[width=1\textwidth]{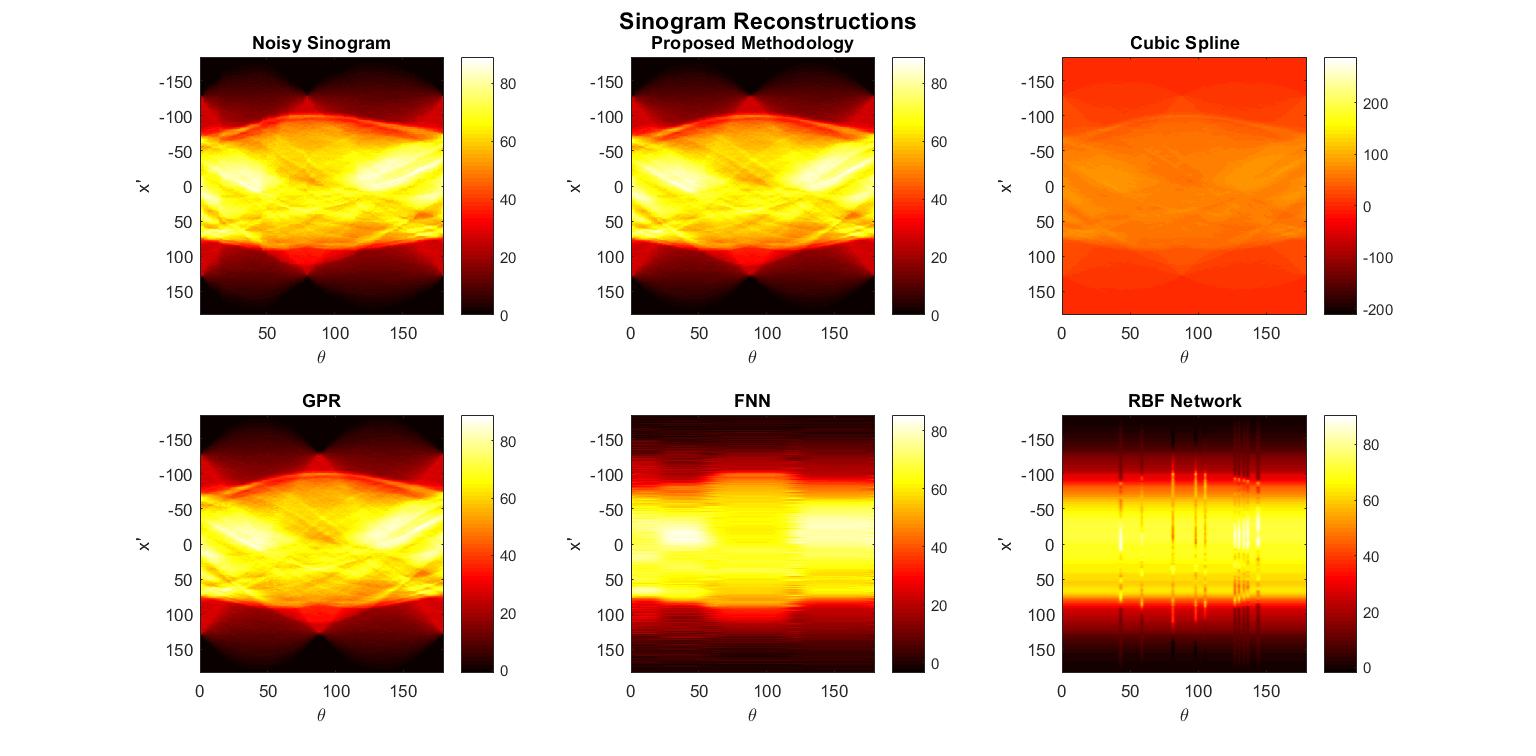}
\caption{Sinograms of tomographic projections  of a human head reconstructed using different interpolation methods. The first two rows correspond to $100$ angular samples, while the last two rows correspond to $300$ angular samples.
From left to right, the panels represent: \emph{Noisy sinogram} (noisy sparse projections from the training set), \emph{Proposed method}, \emph{Cubic spline} interpolation, \emph{Gaussian process regression (GPR)}, \emph{Feed-forward neural network (FNN)}, and \emph{Radial basis function (RBF)} network.}
\label{fig:sh}
\end{figure}

\begin{figure}[H]
\centering
\includegraphics[width=\textwidth]{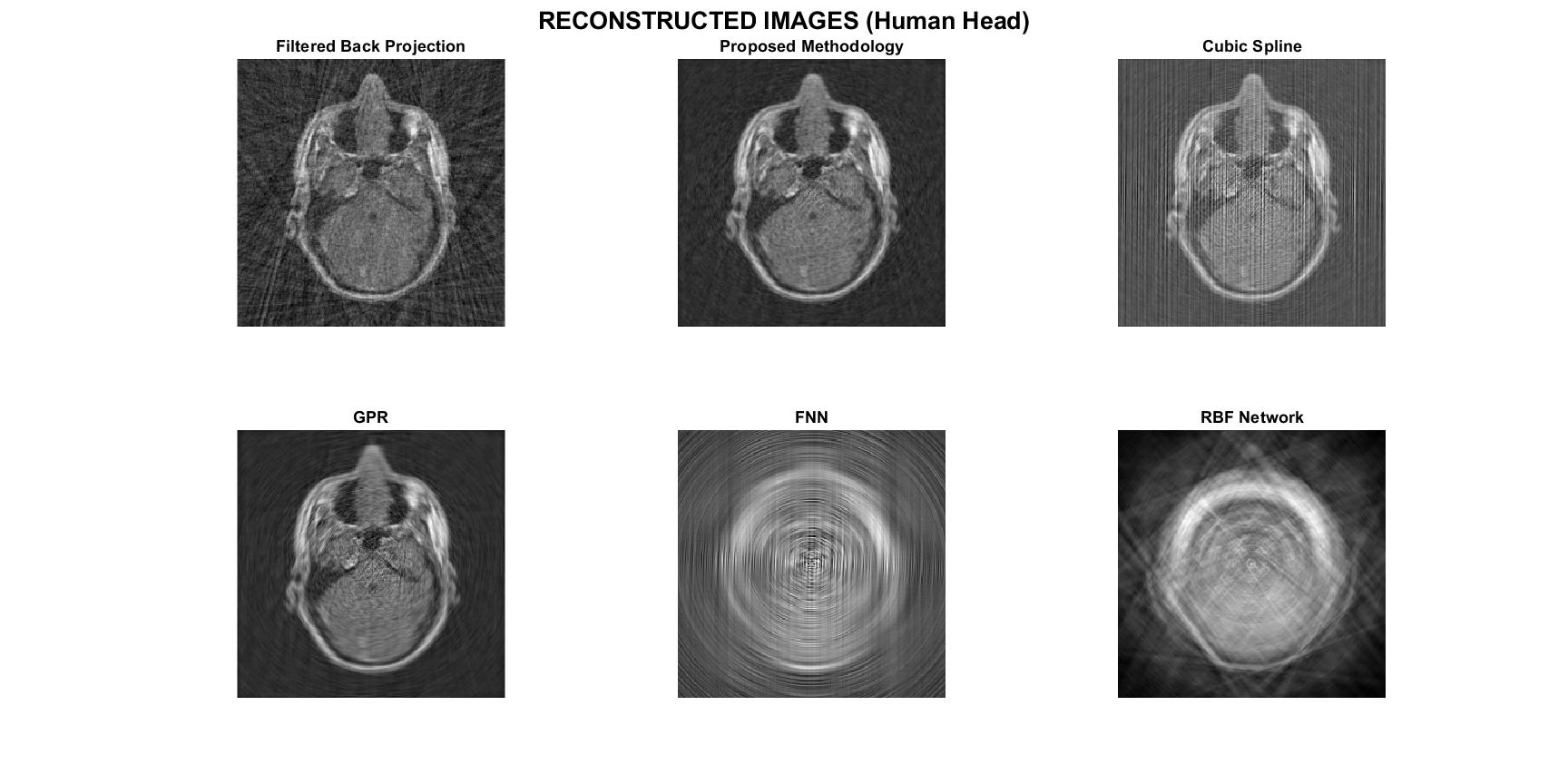}
\includegraphics[width=\textwidth]{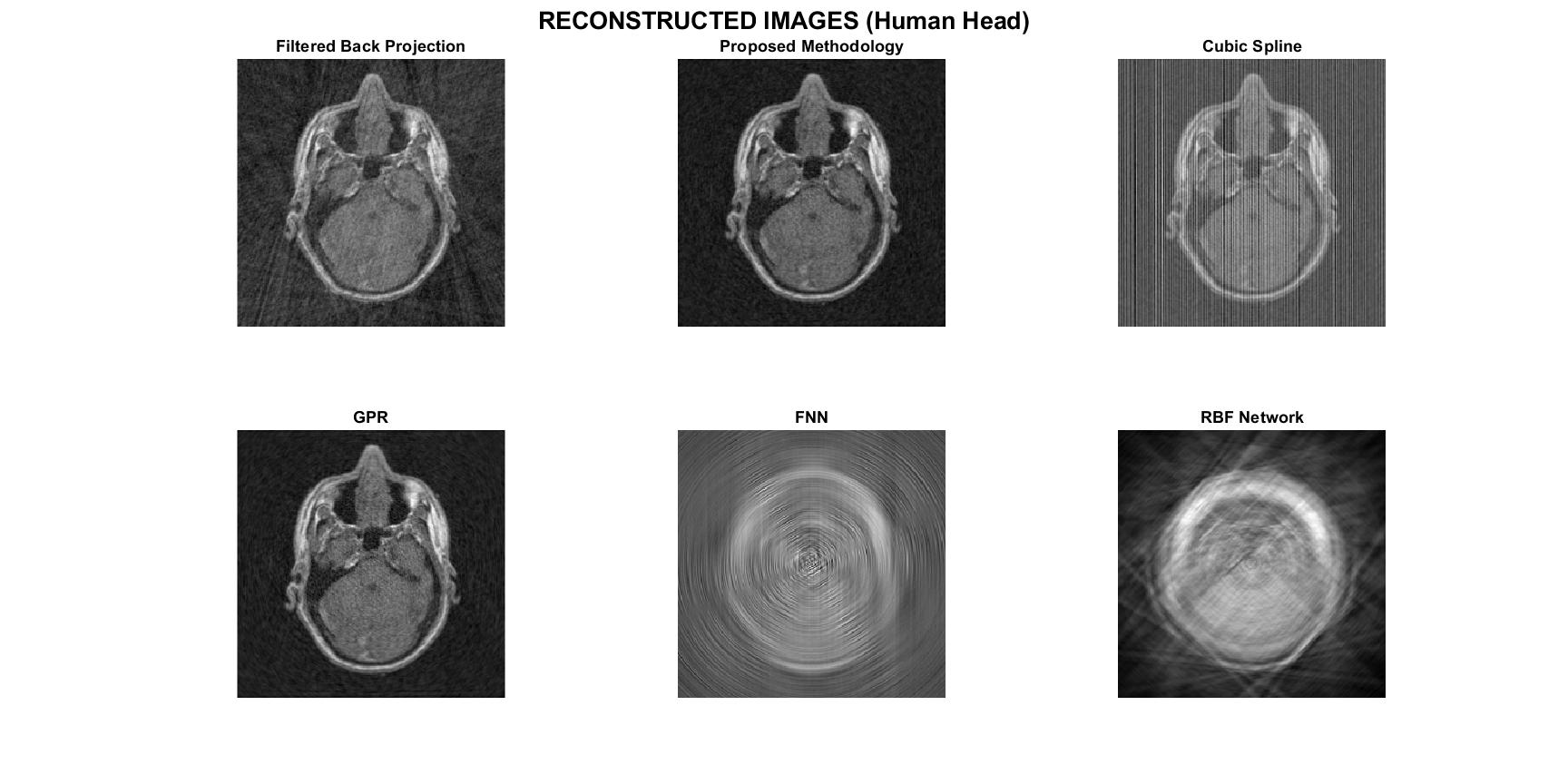}
\caption{Tomographic reconstructions of a human head obtained using different interpolation methods. The first two rows correspond to $100$ angular samples, while the last two rows correspond to $300$ angular samples.
From left to right, the panels show: \emph{Filtered back projection} (tomographic reconstruction obtained by directly applying filtered back projection to the noisy sparse projections from the training set), \emph{Proposed method}, \emph{Cubic spline} interpolation, \emph{Gaussian process regression (GPR)}, \emph{Feed-forward neural network (FNN)}, and \emph{Radial basis function (RBF)} network.}
\label{fig:rh}
\end{figure}

\begin{figure}[H]
\centering
\includegraphics[width=1\textwidth]{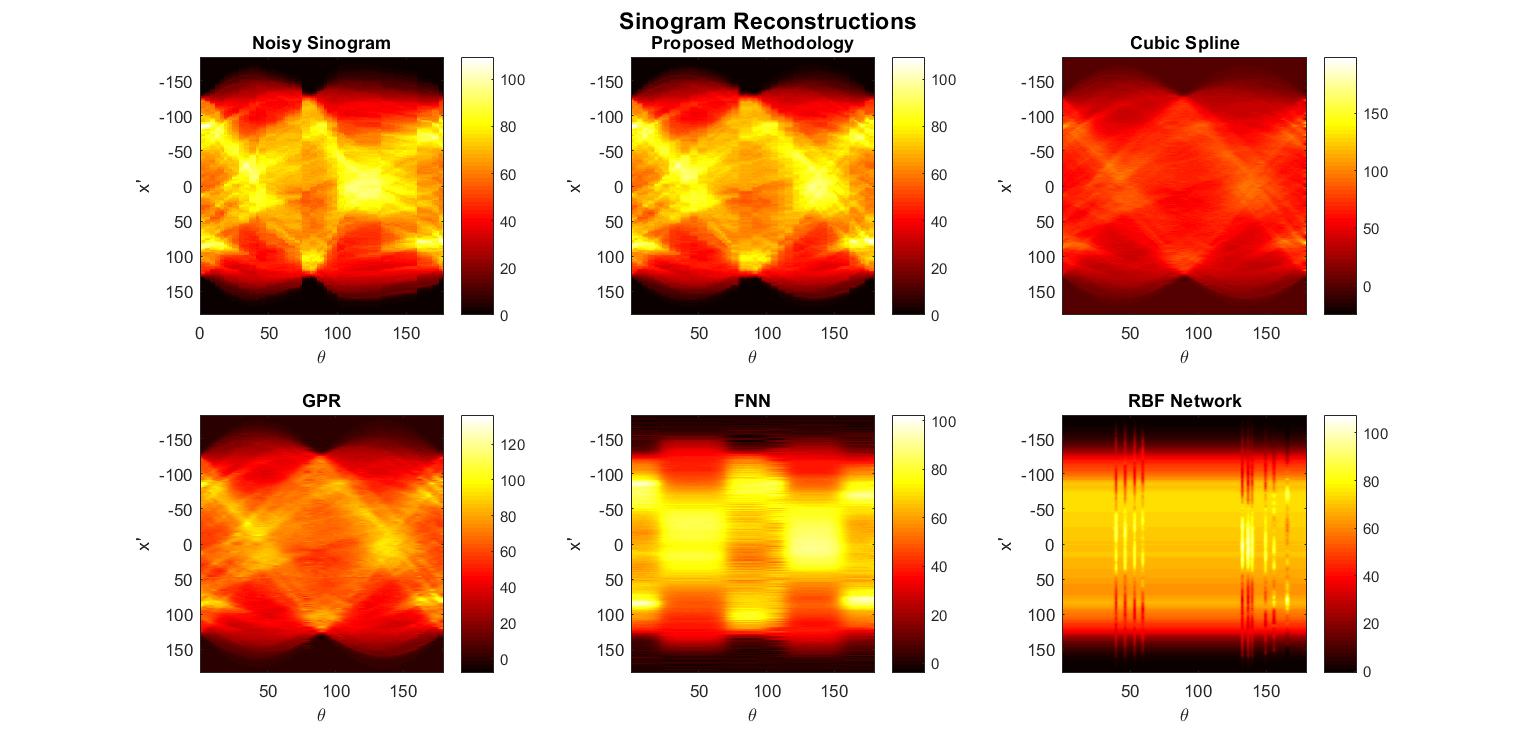}
\includegraphics[width=1\textwidth]{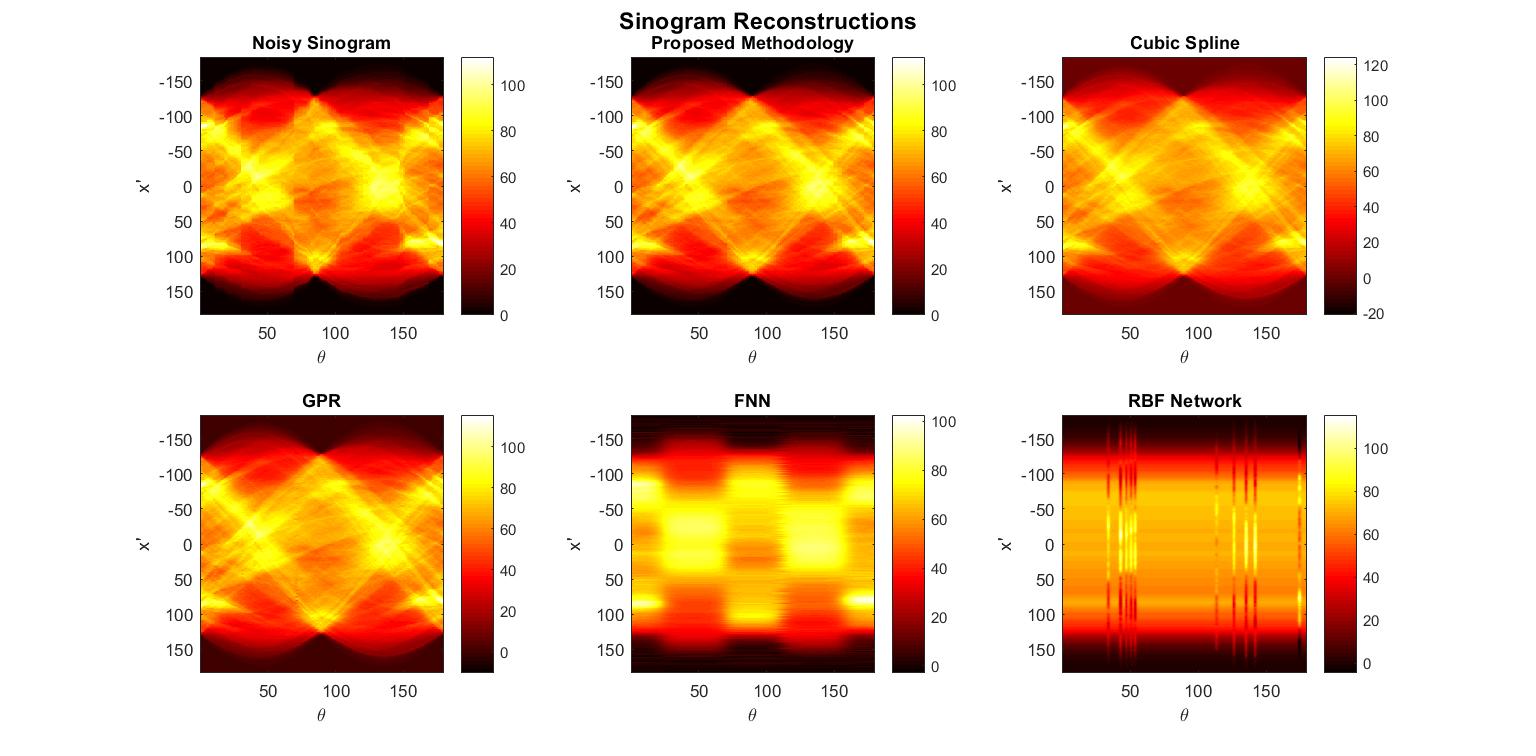}
\caption{Sinograms of tomographic projections  of a human abdomen  reconstructed using different interpolation methods. The first two rows correspond to $100$ angular samples, while the last two rows correspond to $300$ angular samples.
From left to right, the panels represent: \emph{Noisy sinogram} (noisy sparse projections from the training set), \emph{Proposed method}, \emph{Cubic spline} interpolation, \emph{Gaussian process regression (GPR)}, \emph{Feed-forward neural network (FNN)}, and \emph{Radial basis function (RBF)} network.}
\label{fig:sa}
\end{figure}

\begin{figure}[H]
\centering
\includegraphics[width=\textwidth]{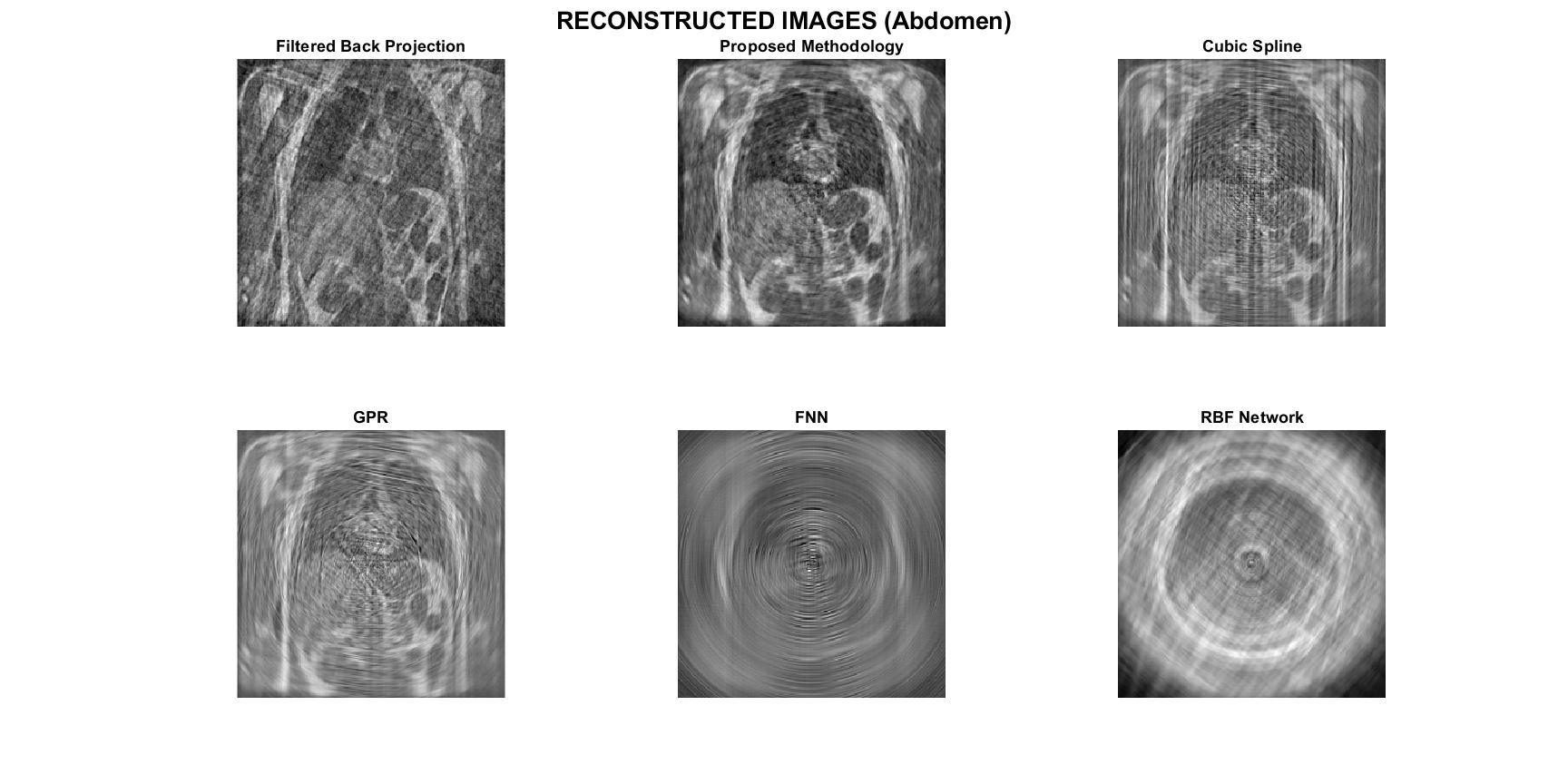}
\includegraphics[width=\textwidth]{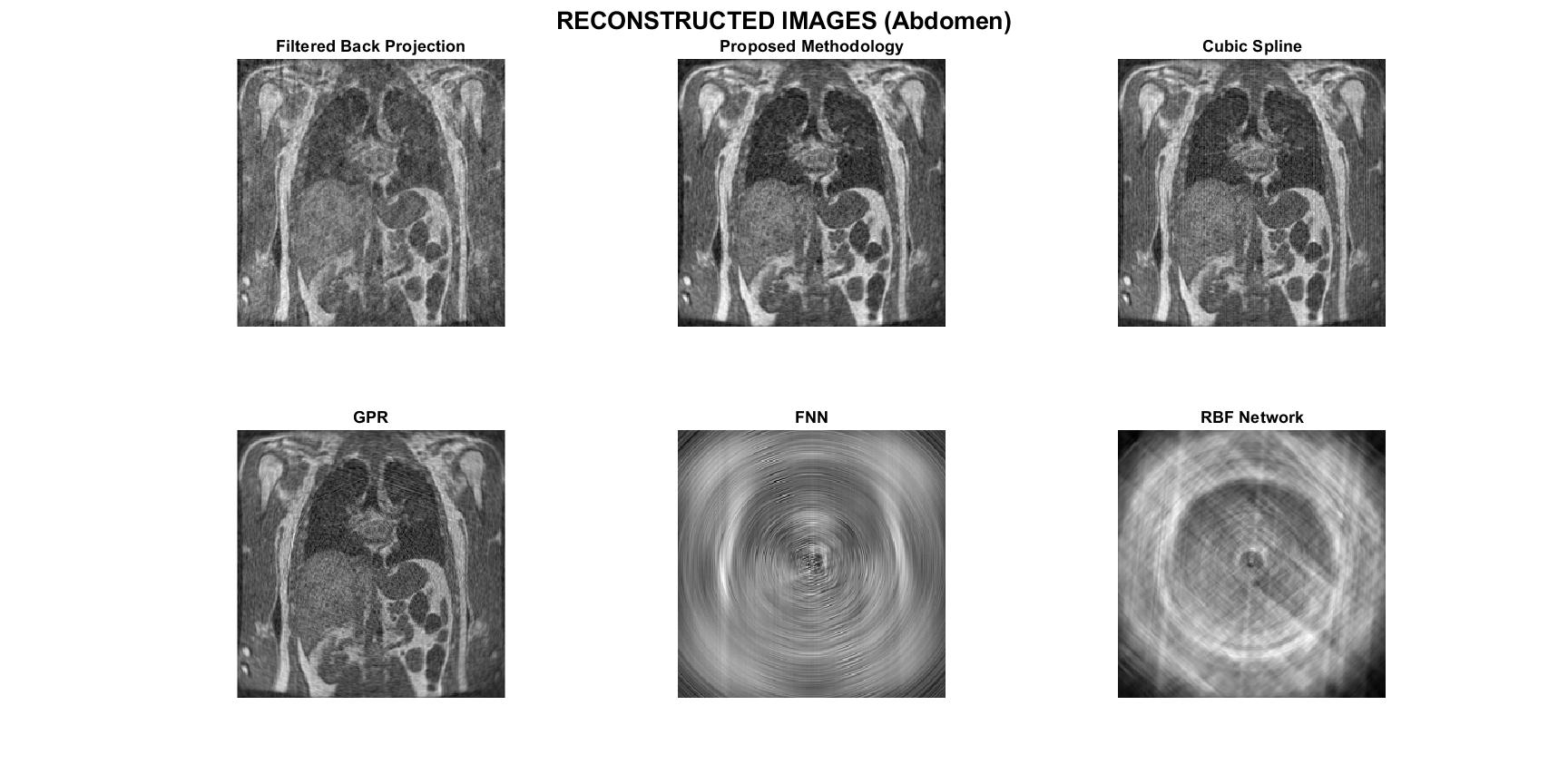}
\caption{Tomographic reconstructions of a human abdomen obtained using different interpolation methods. The first two rows correspond to $100$ angular samples, while the last two rows correspond to $300$ angular samples.
From left to right, the panels show: \emph{Filtered back projection} (tomographic reconstruction obtained by directly applying filtered back projection to the noisy sparse projections from the training set), \emph{Proposed method}, \emph{Cubic spline} interpolation, \emph{Gaussian process regression (GPR)}, \emph{Feed-forward neural network (FNN)}, and \emph{Radial basis function (RBF)} network.}
\label{fig:ra}
\end{figure}

In Figures~\ref{fig:sp}, \ref{fig:sh}, and \ref{fig:sa}, we present the corresponding sinograms, while Figures~\ref{fig:rp}, \ref{fig:rh}, and \ref{fig:ra} display the tomographic reconstructions of the Shepp--Logan phantom, a human head, and a human abdomen, respectively.

Each figure includes results obtained from: (i) sparse projections using only the training samples, (ii) interpolation via the proposed methodology, (iii) cubic spline interpolation, (iv) Gaussian process regression (GPR) applied to the sinogram, (v) a feedforward neural network (FNN), and (vi) a radial basis function neural network (RBFN).

In all figures, the first and second rows correspond to reconstructions computed using $100$ sample points, whereas the third and fourth rows correspond to $300$ sample points.

From a qualitative perspective, the proposed methodology provides stable and accurate tomographic reconstructions across all test cases, including the Shepp--Logan phantom, the human head, and the human abdomen. In comparison with the alternative approaches, both the proposed method and GPR yield the most visually consistent and reliable sinogram interpolations across all scenarios. In contrast, the remaining methods fail in specific cases. For instance, cubic spline interpolation performs poorly for the human head with $100$ sample points, while both FNN and RBFN exhibit significant reconstruction errors for the human head and human abdomen.

To quantitatively evaluate the interpolation of the sinogram and the resulting tomographic reconstructions, we report both the computational time and the reconstruction error. Table~\ref{tab:combined_results} presents the squared error, computed as the Frobenius norm of the difference between the original image and its reconstruction. This metric provides a quantitative assessment of the reconstruction quality for the Shepp--Logan phantom, the human head, and the human abdomen.

The results show that the proposed methodology consistently achieves the lowest mean squared error (MSE) among all the methods considered. Notably, Gaussian Process Regression (GPR) attains a comparable level of accuracy; however, as reported in Table~\ref{tab:combined_results}, its computational cost is significantly higher than that of the proposed approach. This highlights a trade-off between accuracy and computational efficiency.

As a second objective, we investigate the regularity properties of the tomographic reconstructions obtained with the proposed methodology, and compare them with those induced by Total Variation (TV) regularization. To this end, we use the reconstruction produced by our method as an initialization for an iterative scheme aimed at solving a TV-regularized tomographic inverse problem.

More precisely, we consider the variational problem $ \min_{u \in \mathcal{X}} \; \frac{1}{2}\|Au - f\|_2^2 + \lambda \, \mathrm{TV}(u),$
where $A$ denotes the Radon transform, $f$ represents the observed data, and $\mathrm{TV}(u)$ is the total variation seminorm of $u$. The minimization problem is solved using the algorithm proposed by Chambolle~\cite{chambolle2004algorithm}. The computational implementation of this experiment is available in the GitHub repository~\cite{codigoTV}.

Our primary objective is to quantify the decay of the reconstruction error along the iterations under two different initialization strategies: (i) initialization with the reconstruction obtained from the proposed methodology, and (ii) initialization with the filtered back-projection (FBP) reconstruction computed directly from the sampled data. In particular, we evaluate the reconstruction error as a function of the computational time elapsed during the iterative process. 

In all experiments, we fix the regularization parameter $\lambda = 0.5$ and use a step size of $10^{-3}$ and maximum number of iterations $500$.

\begin{figure}[H]
\centering

\begin{minipage}[t]{0.49\textwidth}
\centering
\includegraphics[width=\textwidth]{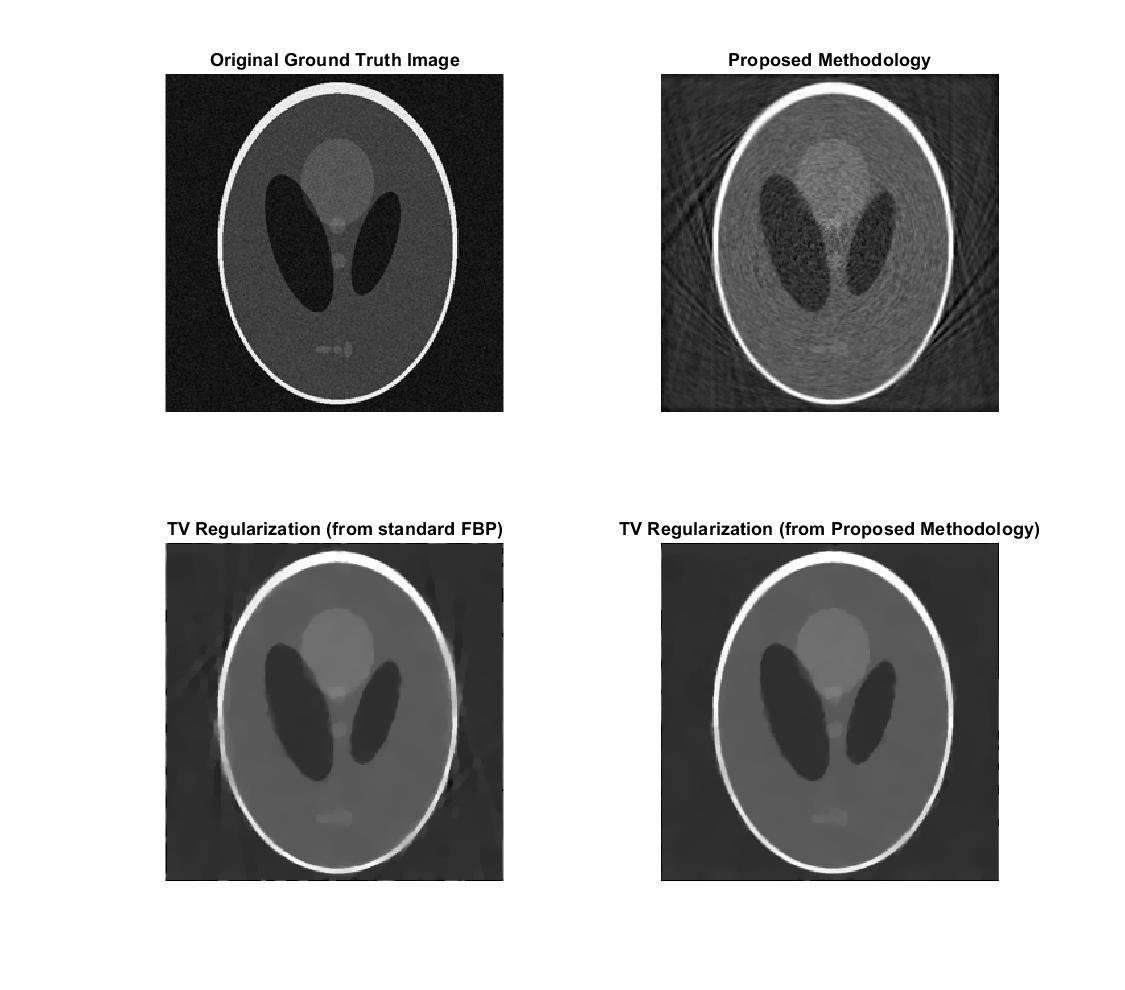}
\\[2pt]
\small (a) Reconstruction results with 100 projection samples.
\end{minipage}
\hfill
\begin{minipage}[t]{0.49\textwidth}
\centering
\includegraphics[width=\textwidth]{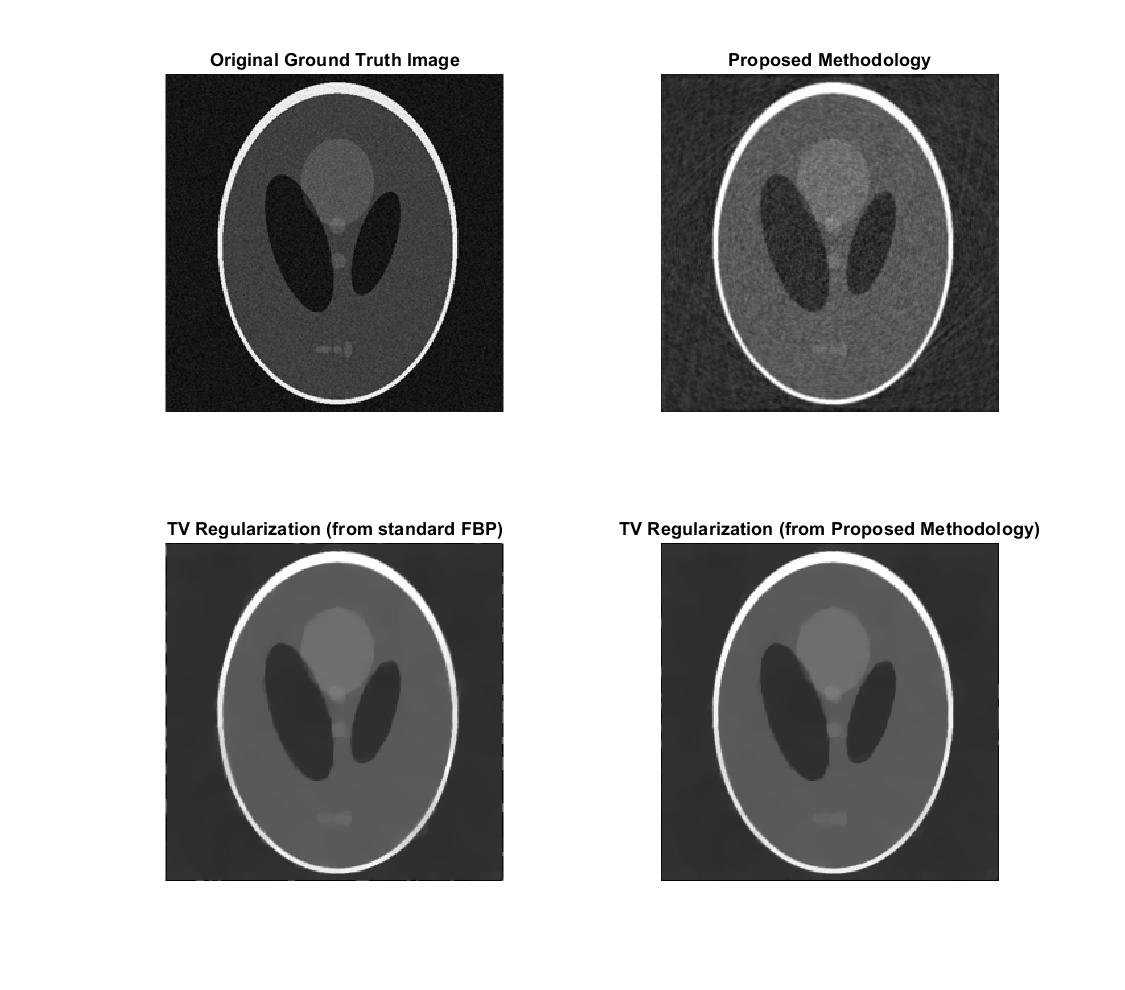}
\\[2pt]
\small (b) Reconstruction results with 300 projection samples.
\end{minipage}

\caption{Tomographic reconstruction of the Shepp--Logan phantom using total variation (TV) regularization under different sampling regimes and initializations. Each panel is arranged as a $2\times 2$ grid. 
\textbf{Top row:} ground truth (left) and reconstruction produced by the proposed method (right). 
\textbf{Bottom row:} TV reconstruction initialized with filtered back projection (FBP) (left) and TV reconstruction initialized with the proposed method (right). Panel (a) corresponds to 100 projection samples, while panel (b) corresponds to 300 projection samples.
}
\label{fig:tv_p}
\end{figure}

\begin{figure}[H]
\centering

\begin{minipage}[t]{0.49\textwidth}
\centering
\includegraphics[width=\textwidth]{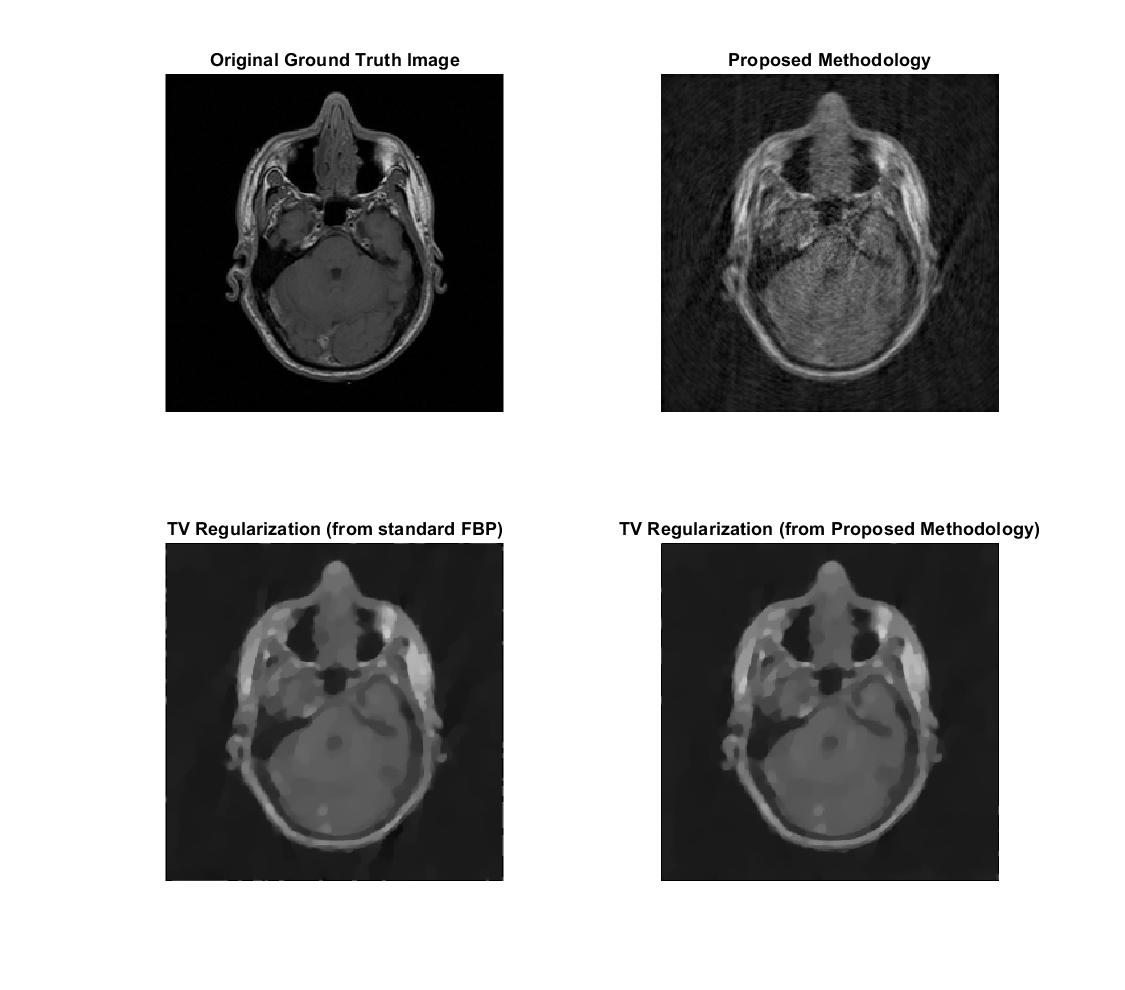}
\\[2pt]
\small (a) Reconstruction results with 100 projection samples.
\end{minipage}
\hfill
\begin{minipage}[t]{0.49\textwidth}
\centering
\includegraphics[width=\textwidth]{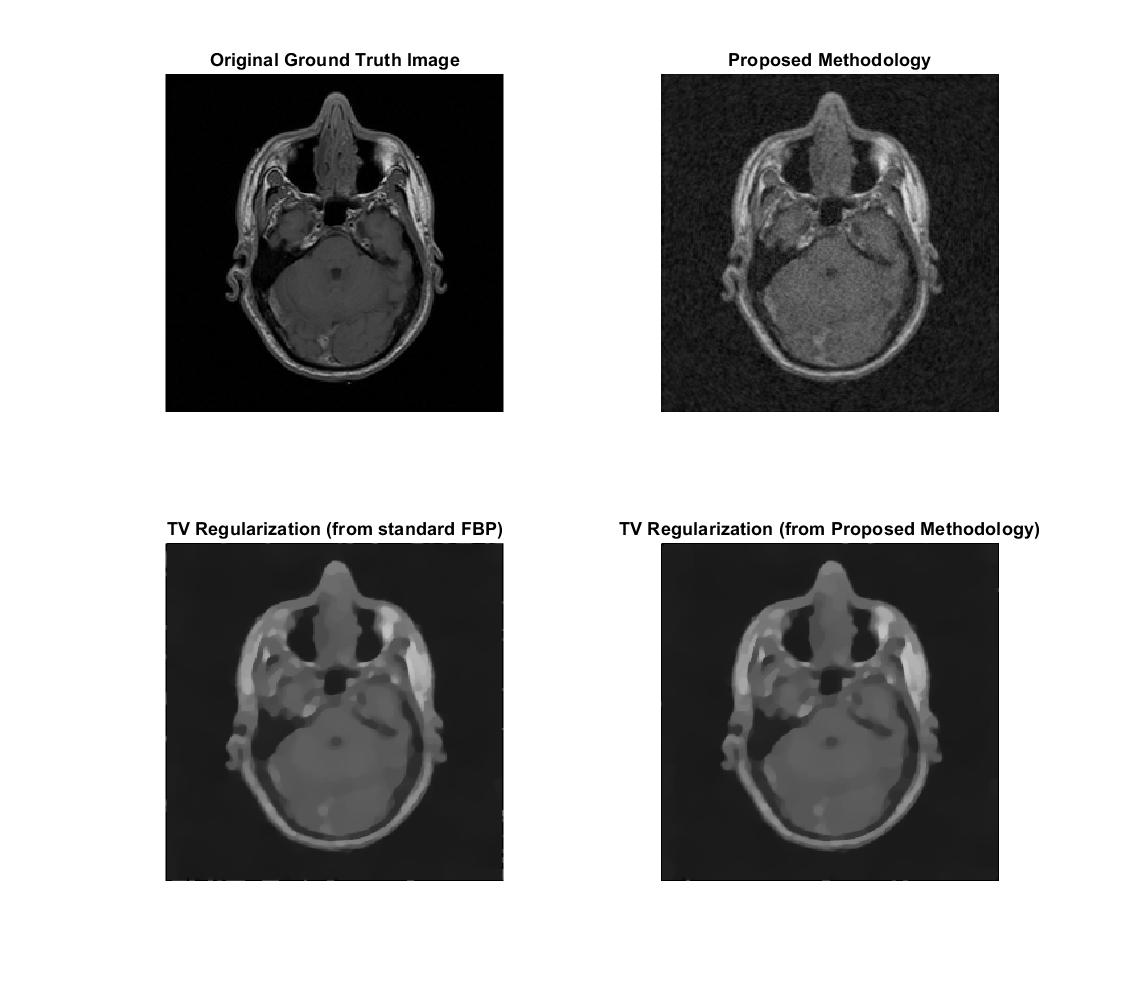}
\\[2pt]
\small (b) Reconstruction results with 300 projection samples.
\end{minipage}

\caption{Tomographic reconstruction of the human head  using total variation (TV) regularization under different sampling regimes and initializations. Each panel is arranged as a $2\times 2$ grid. 
\textbf{Top row:} ground truth (left) and reconstruction produced by the proposed method (right). 
\textbf{Bottom row:} TV reconstruction initialized with filtered back projection (FBP) (left) and TV reconstruction initialized with the proposed method (right). Panel (a) corresponds to 100 projection samples, while panel (b) corresponds to 300 projection samples.
}
\label{fig:tv_h}
\end{figure}

\begin{figure}[H]
\centering

\begin{minipage}[t]{0.49\textwidth}
\centering
\includegraphics[width=\textwidth]{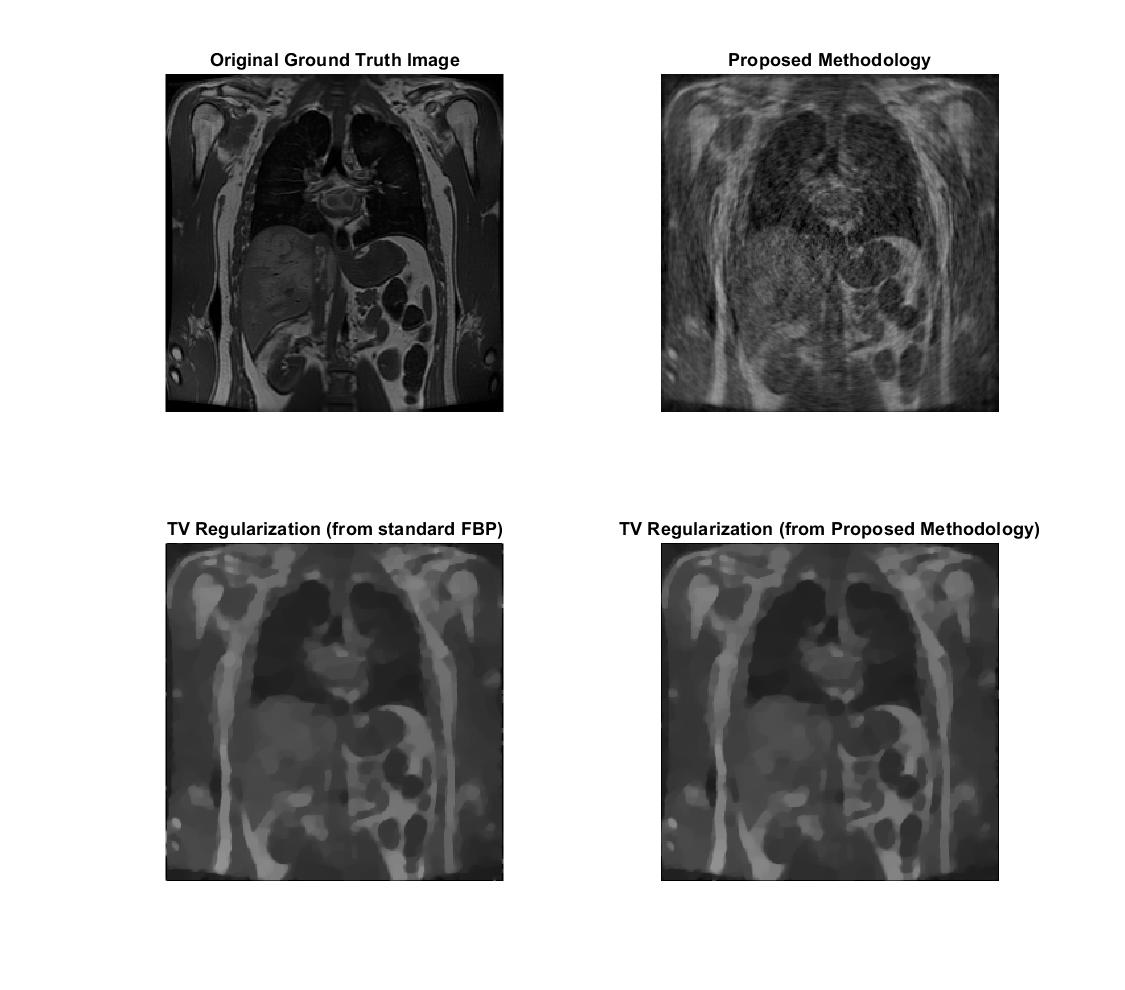}
\\[2pt]
\small (a) Reconstruction results with 100 projection samples.
\end{minipage}
\hfill
\begin{minipage}[t]{0.49\textwidth}
\centering
\includegraphics[width=\textwidth]{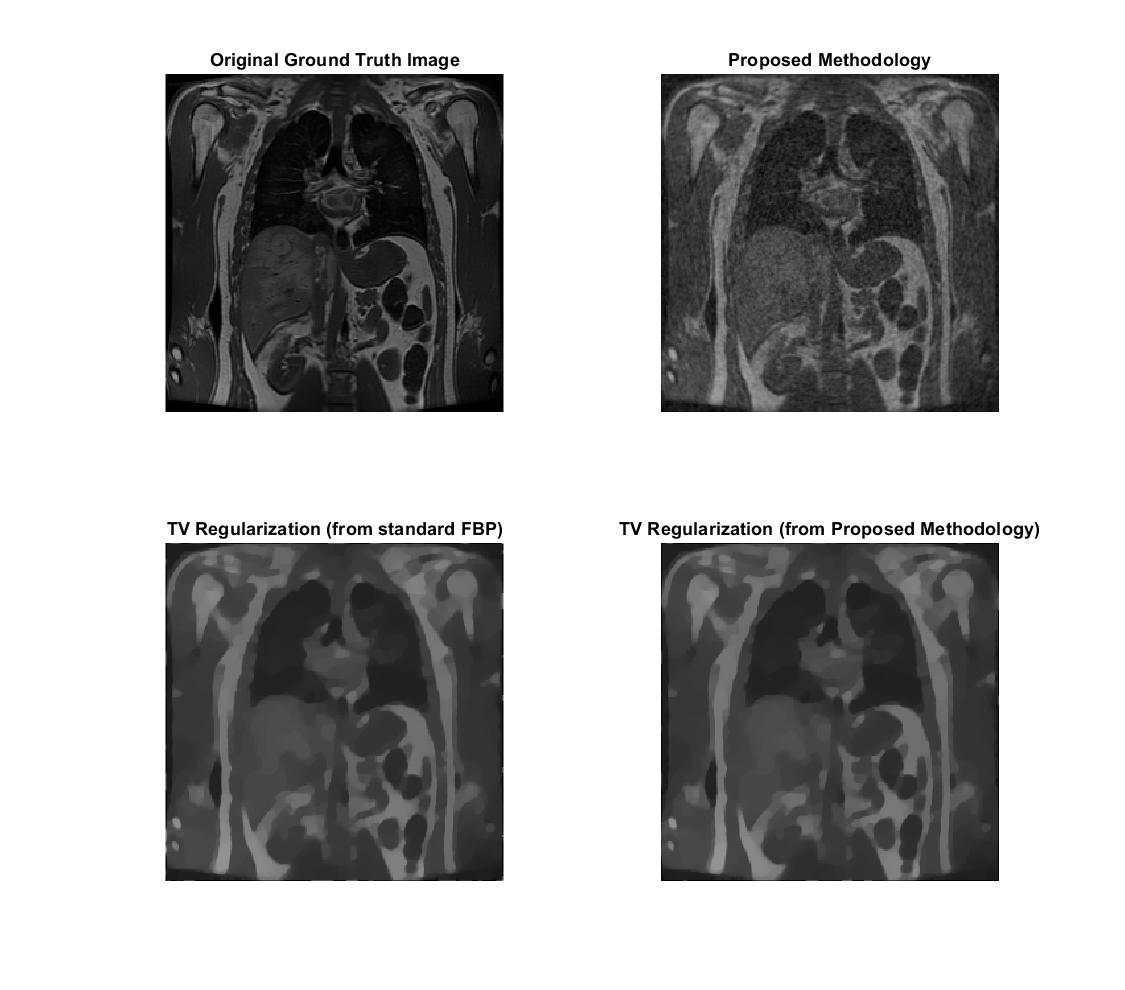}
\\[2pt]
\small (b) Reconstruction results with 300 projection samples.
\end{minipage}

\caption{Tomographic reconstruction of the human abdomen  using total variation (TV) regularization under different sampling regimes and initializations. Each panel is arranged as a $2\times 2$ grid. 
\textbf{Top row:} ground truth (left) and reconstruction produced by the proposed method (right). 
\textbf{Bottom row:} TV reconstruction initialized with filtered back projection (FBP) (left) and TV reconstruction initialized with the proposed method (right). Panel (a) corresponds to 100 projection samples, while panel (b) corresponds to 300 projection samples.
}
\label{fig:tv_a}
\end{figure}

In Figures~\ref{fig:tv_p}, \ref{fig:tv_h}, and \ref{fig:tv_a}, we present reconstructed images obtained via total variation (TV) regularization for the Shepp--Logan phantom, a human head, and a human abdomen, respectively. Reconstructions are shown for both $100$ and $300$ projection samples. Each figure consists of two $2\times 2$ blocks, arranged as follows: the top-left panel displays the ground truth, the top-right panel shows the reconstruction obtained using the proposed method, the bottom-left panel depicts the TV reconstruction initialized with filtered back projection (FBP), and the bottom-right panel displays the TV reconstruction initialized with the proposed method. Panels on the left, labeled (a), correspond to $100$ projection samples, whereas panels on the right, labeled (b), correspond to $300$ projection samples.

\begin{figure}[H]
    \centering
    \begin{minipage}[t]{0.48\textwidth}
        \centering
        \includegraphics[width=\textwidth]{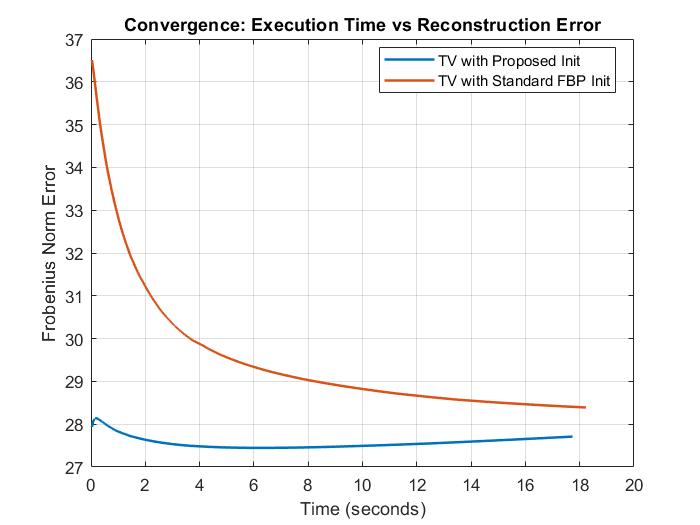}
        \small (a) $100$ projections.
    \end{minipage}
    \hfill
    \begin{minipage}[t]{0.48\textwidth}
        \centering
        \includegraphics[width=\textwidth]{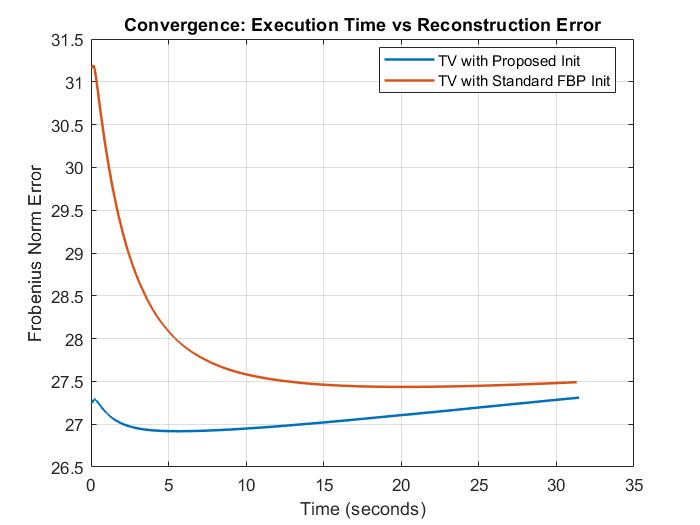}
        \small (b) $300$ projections.
    \end{minipage}

    \caption{Evolution of the Frobenius norm of the reconstruction error as a function of computational time during the Total Variation (TV) regularization process of the Shepp--Logan phantom. The left panel corresponds to $100$ projections, while the right panel corresponds to $300$ projections. We compare the convergence behavior of Filtered Back Projection (FBP, red curve) and the proposed method (blue curve).}
    
    \label{fig:tv_p_time}
\end{figure}

\begin{figure}[H]
    \centering
    \begin{minipage}[t]{0.48\textwidth}
        \centering
        \includegraphics[width=\textwidth]{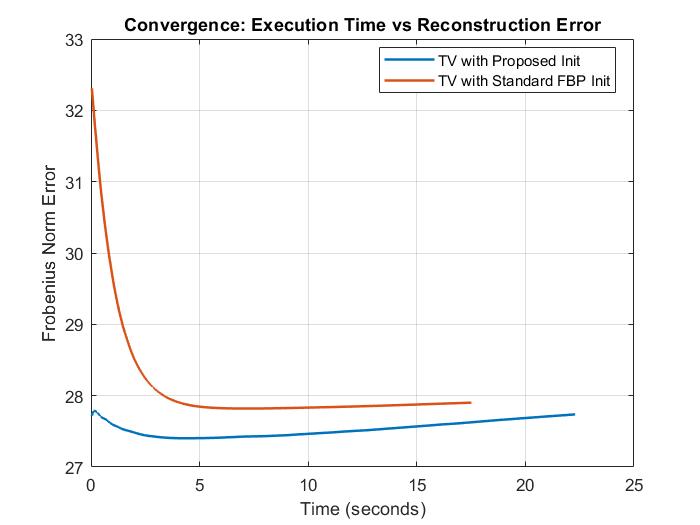}
        \small (a) $100$ projections.
    \end{minipage}
    \hfill
    \begin{minipage}[t]{0.48\textwidth}
        \centering
        \includegraphics[width=\textwidth]{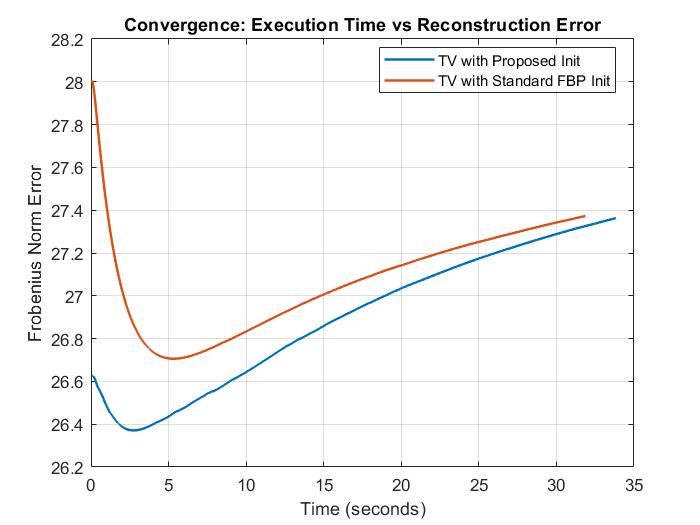}
        \small (b) $300$ projections.
    \end{minipage}

    \caption{Evolution of the Frobenius norm of the reconstruction error as a function of computational time during the Total Variation (TV) regularization process of the human head. The left panel corresponds to $100$ projections, while the right panel corresponds to $300$ projections. We compare the convergence behavior of Filtered Back Projection (FBP, red curve) and the proposed method (blue curve).}
    
    \label{fig:tv_h_time}
\end{figure}

\begin{figure}[H]
    \centering
    \begin{minipage}[t]{0.48\textwidth}
        \centering
        \includegraphics[width=\textwidth]{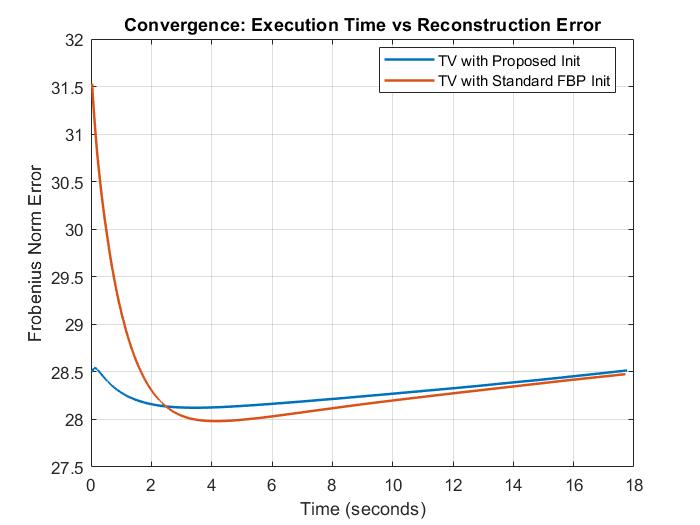}
        \small (a) $100$ projections.
    \end{minipage}
    \hfill
    \begin{minipage}[t]{0.48\textwidth}
        \centering
        \includegraphics[width=\textwidth]{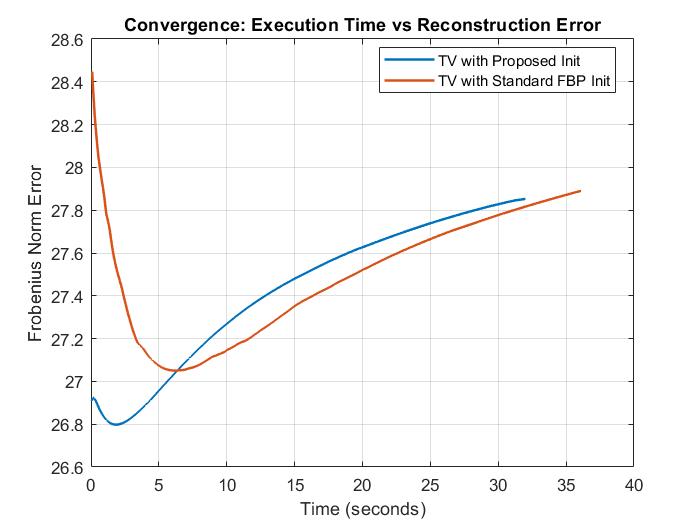}
        \small (b) $300$ projections.
    \end{minipage}

    \caption{Evolution of the Frobenius norm of the reconstruction error as a function of computational time during the Total Variation (TV) regularization process of the human abdomen. The left panel corresponds to $100$ projections, while the right panel corresponds to $300$ projections. We compare the convergence behavior of Filtered Back Projection (FBP, red curve) and the proposed method (blue curve).}
    
    \label{fig:tv_a_time}
\end{figure}

To provide a quantitative assessment of these reconstructions, Figures~\ref{fig:tv_p_time}, \ref{fig:tv_h_time}, and \ref{fig:tv_a_time} illustrate the evolution of the reconstruction error as a function of execution time for both the proposed methodology and the standard FBP-initialized TV regularization. Two primary insights can be drawn from these numerical results. 

First, for the Shepp--Logan phantom and the human head, the error reduction during the TV optimization phase is marginal when the process is initialized with the proposed method. In contrast, the FBP initialization exhibits a steep initial descent, as it starts far from the regularized solution. Most notably, in the human abdomen case with $300$ projections, the TV regularization process yields a monotonic increase in error when starting from our proposed reconstruction. This behavior suggests that the proposed methodology already yields a solution that resides within a high-fidelity region of the objective landscape; consequently, further TV iterations may lead to over-regularization, where the structural preservation properties of the model are sacrificed for an excessive reduction in total variation, thereby increasing the overall error relative to the ground truth.

\blue{
The different evolution pattern observed in Figure~\ref{fig:tv_a_time} (b), compared with Figures~\ref{fig:tv_p_time}(b) and~\ref{fig:tv_h_time}(b), is mainly due to an over-regularization effect. The proposed reconstruction already incorporates an implicit regularization mechanism, producing an estimate that lies close to a high-fidelity region of the optimization landscape. As a result, additional TV regularization tends to oversmooth subtle anatomical structures rather than improve the reconstruction, causing a slight increase in the reconstruction error. It is also worth emphasizing that the magnitude of this increase is very small in absolute terms. The apparent monotonic growth is visually accentuated by the narrow scale of the vertical axis, whereas the actual degradation in reconstruction accuracy remains marginal.
}

This underscores a key advantage of the proposed framework: it provides a reconstruction that inherently possesses strong regularizing properties. Unlike FBP, which necessitates a significant number of post-processing iterations to achieve acceptable quality, our method reaches a near-optimal state almost immediately, rendering further TV iterations largely redundant or even counterproductive in high-sampling regimes.

A second critical aspect to highlight is the computational time. In all evaluated cases---with the minor exception of the human abdomen at $100$ projections, where the difference in minimum error between the two methods is negligible---the proposed methodology reaches its minimum error faster than the FBP-initialized approach. This indicates that, in practical applications, achieving a regularized minimum error incurs a substantially lower computational cost when utilizing the proposed initialization. This empirical finding is consistent with the theoretical discussion presented in previous sections: the proposed reconstruction inherently satisfies total variation regularization guarantees, a theoretical property that standard FBP lacks.

In summary, the proposed methodology achieves the best trade-off between reconstruction accuracy and computational efficiency. It consistently yields the lowest mean squared error (MSE) while maintaining a significantly lower computational cost compared to competing methods, such as FBP, without requiring any \emph{a priori} information. As demonstrated by the quantitative results, the proposed reconstruction implicitly provides a robust degree of TV regularization. Consequently, we can compute TV-regularized images in a fraction of the computational time required when initializing the regularization scheme with standard FBP.

\begin{table}[H]
\centering
\begin{tabular}{|c|c|cccccc|} 
 \hline
\textbf{Image} & \textbf{Batch size} & \textbf{FBP} & \textbf{Proposed} & \textbf{FNN} & \textbf{Spline} & \textbf{GPR} & \textbf{RBFN} \\ 
 \hline

\multicolumn{8}{|c|}{\textbf{Mean Squared Error (MSE)}} \\
\hline

Shepp--Logan & 100 & 35.87 & 26.79 & 45.08 & 112.00& 27.03 & 40.97 \\
Shepp--Logan & 300 & 30.40 & 26.31 & 40.77 & 28.69  & 26.42 & 40.57 \\

Human head   & 100  & 30.51 & 26.68 & 34.63 & 32.411  & 26.90 & 32.47 \\ 
Human head   & 300 & 28.37 & 26.28  & 33.99 & 43.97 & 26.33 & 32.06 \\

Human abdomen & 100  & 31.04 & 27.62 & 34.09 & 32.25 & 30.44 & 32.95 \\ 
Human abdomen & 300 & 28.20  & 26.39 & 31.95 & 26.71 & 26.87 & 32.54 \\

\hline
\multicolumn{8}{|c|}{\textbf{Computational Time (seconds)}} \\
\hline

Shepp--Logan & 100 & --- & 0.24 & 2.98  & 0.20 & 20.17 & 1.33\\
Shepp--Logan & 300 & --- & 0.71 & 5.34  & 0.21 &  298.25 & 0.98 \\

Human head   & 100  & --- & 0.48 & 4.13 & 0.29  & 26.19 & 2.15 \\ 
Human head   & 300  & --- & 0.77 & 6.44 & 0.22  & 357.12 & 1.09 \\

Human abdomen & 100  & --- & 0.24 & 2.96 & 0.19 & 20.98 & 0.78 \\ 
Human abdomen & 300 & --- & 0.59 & 4.98 & 0.27 & 325.25 & 1.22\\

 \hline
\end{tabular}

\caption{Comparison of reconstruction performance in terms of mean squared error (MSE) and computational time (in seconds) for the Shepp--Logan phantom, human head, and human abdomen datasets. The \emph{FBP} column corresponds to filtered back-projection applied directly to the noisy data, while \emph{Proposed} denotes the reconstruction method introduced in this work. The \emph{FNN}, \emph{Spline}, \emph{GPR}, and \emph{RBFN} columns report results obtained by first interpolating the sinogram using, respectively, a feed-forward neural network, cubic spline interpolation, Gaussian process regression, and a radial basis function network, followed by reconstruction via filtered back-projection.}
\label{tab:combined_results}
\end{table}



\section{Conclusion and Future Work}
\label{sec:conclusions}

\blue{In this paper, we proposed a novel data-driven approach for approximating smooth functions defined on manifolds. The methodology is based on the Nadaraya--Watson kernel regression estimator, with a central contribution given by the construction of a spatially adaptive diffusion bandwidth $\varepsilon(x)$ via Voronoi tessellations. The proposed selection strategy ensures computational stability while remaining consistent with the underlying geometry of the data.

The method does not require a training phase, and the computational cost of inference scales linearly with the number of sample points. This results in improved computational efficiency compared to standard data-driven interpolation techniques, including Gaussian process regression, feedforward neural networks, and radial basis function networks. We also demonstrated the applicability of the approach to sparse computed tomography reconstruction problems.

The numerical results indicate that the proposed method achieves a favorable trade-off between accuracy and computational cost while remaining competitive with existing approaches.

The theoretical framework developed in this work assumes arbitrary interpolation points, while the regularity assumptions are imposed on the interpolating function. In particular, the interpolant is assumed to satisfy a vanishing-gradient condition at the interpolation points, and the proposed formulation inherently attenuates high-frequency components, thereby introducing an intrinsic regularization effect. Consequently, the method is particularly well suited for smooth or slowly varying functions and images. By contrast, problems involving sharp edges, fine textures, discontinuities, or highly oscillatory or stochastic patterns may exhibit oversmoothing, which limits the accurate recovery of high-frequency features. Moreover, the present deterministic framework does not provide uncertainty quantification, which remains outside the scope of this work.

Future work will focus on extending the methodology to applications in image processing, engineering, and quantitative finance. In particular, we plan to investigate its integration with reinforcement learning techniques for large-scale dynamic optimization problems in finance \cite{consigli2024optimal,coache2024reinforcement,cartea2023reinforcement}, as well as its application to interpolation schemes arising in the numerical solution of partial differential equations. Another promising direction is the development of probabilistic extensions of the proposed framework capable of incorporating uncertainty quantification while preserving the computational efficiency of the present approach.}

\vskip 0.2in

\noindent\textbf{Acknowledgements.} The author acknowledges support from the Centro de Modelamiento Matemático (CMM) through the BASAL funding program FB210005, funded by ANID-Chile. The author also gratefully acknowledges Prof.~Mingyuan Zhou (The University of Texas at Austin) for his careful review and valuable comments, as well as the participants of the Postdoctoral Seminar in the CMM for their insightful suggestions, which contributed to improving this manuscript.


\vskip 0.2in
\bibliographystyle{unsrt} 
\bibliography{bibli}

@book{wendland2004scattered,
  title={Scattered Data Approximation},
  author={Wendland, Holger},
  year={2004},
  publisher={Cambridge University Press}
}

@book{schumaker2007spline,
  title={Spline Functions: Basic Theory},
  author={Schumaker, Larry L.},
  year={2007},
  publisher={Cambridge University Press}
}

@book{rasmussen2006gaussian,
  title={Gaussian Processes for Machine Learning},
  author={Rasmussen, Carl Edward and Williams, Christopher K. I.},
  year={2006},
  publisher={MIT Press}
}

@article{belkin2003laplacian,
  title={Laplacian Eigenmaps for dimensionality reduction and data representation},
  author={Belkin, Mikhail and Niyogi, Partha},
  journal={Neural Computation},
  volume={15},
  number={6},
  pages={1373--1396},
  year={2003}
}

@article{coifman2006diffusion,
  title={Diffusion maps},
  author={Coifman, Ronald R. and Lafon, St{\'e}phane},
  journal={Applied and Computational Harmonic Analysis},
  volume={21},
  number={1},
  pages={5--30},
  year={2006}
}

@article{bronstein2021geometric,
  title={Geometric deep learning: grids, groups, graphs, geodesics, and gauges},
  author={Bronstein, Michael M. and Bruna, Joan and Cohen, Taco and Veličkovi{\'c}, Petar},
  journal={arXiv preprint arXiv:2104.13478},
  year={2021}
}

@inproceedings{mescheder2019occupancy,
  title={Occupancy Networks: Learning 3D Reconstruction in Function Space},
  author={Mescheder, Lars and Oechsle, Michael and Niemeyer, Michael and Nowozin, Sebastian and Geiger, Andreas},
  booktitle={Proceedings of CVPR},
  pages={4460--4470},
  year={2019}
}

@inproceedings{sitzmann2020implicit,
  title={Implicit Neural Representations with Periodic Activation Functions},
  author={Sitzmann, Vincent and Martel, Julien N. P. and Bergman, Alexander W. and Lindell, David B. and Wetzstein, Gordon},
  booktitle={Advances in Neural Information Processing Systems},
  year={2020}
}

@article{watson1964smooth,
  title={Smooth regression analysis},
  author={Watson, Geoffrey S},
  journal={Sankhy{\=a}: The Indian Journal of Statistics, Series A},
  pages={359--372},
  year={1964},
  publisher={JSTOR}
}

@article{nadaraya1964estimating,
  title={On estimating regression},
  author={Nadaraya, Elizbar A},
  journal={Theory of Probability \& Its Applications},
  volume={9},
  number={1},
  pages={141--142},
  year={1964},
  publisher={SIAM}
}

@misc{codigoTV,
  author       = {Almeida Gomez, Alvaro},
  title        = {Numerical Implementation of Total Variation CT Reconstruction in {MATLAB}},
  year         = {2026},
  howpublished = {\url{https://github.com/alvaroalmeidagomez/LTD/tree/main/Post_TV}},
 note         = {Accessed: April 1, 2026}}

@article{chambolle2004algorithm,
  title={An algorithm for total variation minimization and applications},
  author={Chambolle, Antonin},
  journal={Journal of Mathematical imaging and vision},
  volume={20},
  number={1},
  pages={89--97},
  year={2004},
  publisher={Springer}
}

@article{xuanhao2022effect,
  title={Effect of the number of projections in X-ray CT imaging on image quality and digital volume correlation measurement},
  author={Xuanhao, Zhang and Lijuan, Sun and Bo, Wang and Bing, Pan},
  journal={Measurement},
  volume={194},
  pages={111061},
  year={2022},
  publisher={Elsevier}
}

@article{candes2006stable,
  title={Stable signal recovery from incomplete and inaccurate measurements},
  author={Candes, Emmanuel J and Romberg, Justin K and Tao, Terence},
  journal={Communications on Pure and Applied Mathematics: A Journal Issued by the Courant Institute of Mathematical Sciences},
  volume={59},
  number={8},
  pages={1207--1223},
  year={2006},
  publisher={Wiley Online Library}
}

@ARTICLE{taocomp,
  author={Candes, E.J. and Romberg, J. and Tao, T.},
  journal={IEEE Transactions on Information Theory}, 
  title={Robust uncertainty principles: exact signal reconstruction from highly incomplete frequency information}, 
  year={2006},
  volume={52},
  number={2},
  pages={489-509},
  doi={10.1109/TIT.2005.862083}}

@ARTICLE{dohcomp,
  author={Donoho, D.L.},
  journal={IEEE Transactions on Information Theory}, 
  title={Compressed sensing}, 
  year={2006},
  volume={52},
  number={4},
  pages={1289-1306},
  doi={10.1109/TIT.2006.871582}}

@inproceedings{bao2019sparse,
  title={Sparse-view CT reconstruction via convolutional sparse coding},
  author={Bao, Peng and Xia, Wenjun and Yang, Kang and Zhou, Jiliu and Zhang, Yi},
  booktitle={2019 IEEE 16th International Symposium on Biomedical Imaging (ISBI 2019)},
  pages={1446--1449},
  year={2019},
  organization={IEEE}
}

@misc{imagesrad,
  author       = {NLM, National Library of Medicine},
  title        = {The Visible Human Project},
  year         = {2026},
  howpublished = {\url{https://www.nlm.nih.gov/research/visible/mri.html}},
 note         = {Accessed: April 1, 2026}

}

@misc{codigoCT,
  author       = {Almeida Gomez, Alvaro},
  title        = {Numerical implementation of sparse CT reconstruction in {MATLAB}},
  year         = {2026},
  howpublished = {\url{https://github.com/alvaroalmeidagomez/LTD/tree/main/Sparse_CT}},
 note         = {Accessed: April 1, 2026}

}

@article{zhang2018sparse,
  title={A sparse-view CT reconstruction method based on combination of DenseNet and deconvolution},
  author={Zhang, Zhicheng and Liang, Xiaokun and Dong, Xu and Xie, Yaoqin and Cao, Guohua},
  journal={IEEE transactions on medical imaging},
  volume={37},
  number={6},
  pages={1407--1417},
  year={2018},
  publisher={IEEE}
}

@article{consigli2024optimal,
  title={Optimal dynamic fixed-mix portfolios based on reinforcement learning with second order stochastic dominance},
  author={Consigli, Giorgio and Gomez, Alvaro A and Zubelli, Jorge P},
  journal={Engineering Applications of Artificial Intelligence},
  volume={133},
  pages={108599},
  year={2024},
  publisher={Elsevier}
}

@article{coache2024reinforcement,
  title={Reinforcement learning with dynamic convex risk measures},
  author={Coache, Anthony and Jaimungal, Sebastian},
  journal={Mathematical Finance},
  volume={34},
  number={2},
  pages={557--587},
  year={2024},
  publisher={Wiley Online Library}
}

@article{cartea2023reinforcement,
  title={Reinforcement learning for algorithmic trading},
  author={Cartea, {\'A}lvaro and Jaimungal, Sebastian and S{\'a}nchez-Betancourt, Leandro},
  journal={Machine Learning and Data Sciences for Financial Markets: A Guide to Contemporary Practices. Cambridge University Press},
  year={2023}
}

@book{fasshauer2007meshfree,
title={Meshfree Approximation Methods with MATLAB},
author={Fasshauer, Gregory},
year={2007},
publisher={World Scientific}
}

@article{kipf2017semi,
title={Semi-Supervised Classification with Graph Convolutional Networks},
author={Kipf, Thomas and Welling, Max},
journal={ICLR},
year={2017}
}

@article{hamilton2017inductive,
title={Inductive Representation Learning on Large Graphs},
author={Hamilton, William and Ying, Rex and Leskovec, Jure},
journal={NeurIPS},
year={2017}
}

@article{hornik1989multilayer,
  title={Multilayer feedforward networks are universal approximators},
  author={Hornik, Kurt and Stinchcombe, Maxwell and White, Halbert},
  journal={Neural networks},
  volume={2},
  number={5},
  pages={359--366},
  year={1989},
  publisher={Elsevier}
}

@inproceedings{rahaman2019spectral,
  title={On the spectral bias of neural networks},
  author={Rahaman, Nasim and Baratin, Aristide and Arpit, Devansh and Draxler, Felix and Lin, Min and Hamprecht, Fred and Bengio, Yoshua and Courville, Aaron},
  booktitle={International Conference on Machine Learning (ICML)},
  pages={5301--5310},
  year={2019},
  organization={PMLR}
}

@article{xu2019frequency,
  title={Training behavior of deep neural networks in frequency domain},
  author={Xu, Zhi-Qin John and Zhang, Yaoyu and Luo, Tao and Xiao, Yanyang and Ma, Zheng},
  journal={International Conference on Neural Information Processing (ICONIP)},
  pages={264--274},
  year={2019},
  publisher={Springer}
}

@article{wang2021understanding,
  title={Understanding and mitigating gradient flow pathologies in physics-informed neural networks},
  author={Wang, Sifan and Teng, Yujun and Perdikaris, Paris},
  journal={SIAM Journal on Scientific Computing},
  volume={43},
  number={5},
  pages={A3055--A3081},
  year={2021},
  publisher={SIAM}
}

@inproceedings{tancik2020fourier,
  title={Fourier features let networks learn high frequency functions in low dimensional domains},
  author={Tancik, Matthew and Srinivasan, Pratul and Mildenhall, Ben and Fridovich-Keil, Sara and Raghavan, Nithin and Singhal, Utkarsh and Ramamoorthi, Ravi and Barron, Jonathan and Ng, Ren},
  booktitle={Advances in Neural Information Processing Systems (NeurIPS)},
  volume={33},
  pages={7520--7531},
  year={2020}
}

@article{vanLieshout2024,
  author    = {van Lieshout, M. N. M.},
  title     = {Non-parametric adaptive bandwidth selection for kernel estimators of spatial intensity functions},
  journal   = {Annals of the Institute of Statistical Mathematics},
  year      = {2024},
  volume    = {76},
  number    = {2},
  pages     = {313--331},
  doi       = {10.1007/s10463-023-00890-6},
  url       = {https://doi.org/10.1007/s10463-023-00890-6},
  issn      = {1572-9052}
}

@article{abramson1982bandwidth,
  author  = {Abramson, Ian S.},
  title   = {On bandwidth variation in kernel estimates---a
             square root law},
  journal = {The Annals of Statistics},
  year    = {1982},
  volume  = {10},
  number  = {4},
  pages   = {1217--1223},
  doi     = {10.1214/aos/1176345986}
}

@article{terrell1992variable,
  author  = {Terrell, George R. and Scott, David W.},
  title   = {Variable kernel density estimation},
  journal = {The Annals of Statistics},
  year    = {1992},
  volume  = {20},
  number  = {3},
  pages   = {1236--1265},
  doi     = {10.1214/aos/1176348768}
}

@book{fan1996local,
  author    = {Fan, Jianqing and Gijbels, Ir\`{e}ne},
  title     = {Local Polynomial Modelling and Its Applications},
  series    = {Monographs on Statistics and Applied Probability},
  volume    = {66},
  publisher = {Chapman \& Hall},
  address   = {London},
  year      = {1996}
}

@book{silverman1986density,
  author    = {Silverman, Bernard W.},
  title     = {Density Estimation for Statistics and Data Analysis},
  publisher = {Chapman \& Hall},
  address   = {London},
  year      = {1986},
  doi       = {10.1007/978-1-4899-3324-9}
}

@inproceedings{wilson2013gaussian,
  title={Gaussian process kernels for pattern discovery and extrapolation},
  author={Wilson, Andrew and Adams, Ryan},
  booktitle={International conference on machine learning},
  pages={1067--1075},
  year={2013},
  organization={PMLR}
}

\appendix
\section*{Appendix}

\section{Proof of Theorem~\ref{thm:local_decay}}
\label{pruebateoprin}

\begin{proof}
Fix $i \in \{1,\dots,N\}$ and define
\[
r_i := \min_{j \neq i} \|x_i - x_j\| > 0 .
\]

For every $x$ in the open ball
\[
B(x_i,r_i/2) := \{ x \in \mathbb{R}^d : \|x-x_i\| < r_i/2 \},
\]
the function $\varepsilon(x)$ defined in~\eqref{parametroepsilon} satisfies
\[
\varepsilon(x) = \|x - x_i\|.
\]

Indeed, for any $j \neq i$, by the triangle inequality,
\[
\|x_j - x_i\| \le \|x_j - x\| + \|x - x_i\|,
\]
which implies
\[
r_i \le \|x_j - x\| + \|x - x_i\| < \|x_j - x\| + \frac{r_i}{2}.
\]
Therefore,
\begin{equation}
\label{lowebcompara}
\|x - x_i\| < \frac{r_i}{2} < \|x - x_j\| , \qquad j \neq i.
\end{equation}
Hence,
\[
\varepsilon(x) = \min_{1 \le j \le N} \|x - x_j\| = \|x - x_i\|.
\]

On the punctured ball $B(x_i,r_i/2)\cap \mathcal{M}\setminus\{x_i\}$, the operator $\bar{g}_{x_1,\dots,x_N}$ reads
\begin{equation}
\label{ecuacionaprox2}
\bar{g}_{x_1,\dots,x_N}(x)
=
\frac{1}{\mathbf{Nm}_{\varepsilon(x)}(x)}
\sum_{j=1}^{N}
\exp\!\left(-\frac{\|x - x_j\|^2}{\|x - x_i\|^2}\right)
g(x_j),
\end{equation}
with normalization factor
\begin{equation}
\label{normalizafactor}
\mathbf{Nm}_{\varepsilon(x)}(x)
=
\sum_{j=1}^{N}
\exp\!\left(-\frac{\|x - x_j\|^2}{\|x - x_i\|^2}\right).
\end{equation}

Since the Gaussian kernel is smooth and $\mathbf{Nm}_{\varepsilon(x)}(x)$ does not vanish on this set (see Section~\ref{sec:stabil}), the function $\bar{g}_{x_1,\dots,x_N}$ is smooth on $B(x_i,r_i/2)\cap \mathcal{M}\setminus\{x_i\}$.
Consider the function $F : \mathbb{R}^n \setminus \{x_i\} \to \mathbb{R}$ defined by
\begin{equation}
\label{funcionexponencial}
F(x)
=
\sum_{j=1}^{N}
\exp\!\left(-\frac{\|x - x_j\|^2}{\|x - x_i\|^2}\right)
g(x_j).
\end{equation}
A direct computation yields
\begin{equation}
\label{expansionderivada}
\nabla F(x)
=
2 \sum_{j=1}^{N}
\exp\!\left(-\frac{\|x - x_j\|^2}{\|x - x_i\|^2}\right)
\frac{(x-x_j)\|x-x_i\|^2 - (x-x_i)\|x-x_j\|^2}{\|x-x_i\|^4}
\, g(x_j).
\end{equation}

Observe that when $i=j$, the corresponding term in the sum in~\eqref{expansionderivada} vanishes. On the other hand, since $\mathcal{M}$ is compact, there exists a constant $C>0$ such that

\[
\|x-y\| \le C \quad \text{for all } x,y \in \mathcal{M}.
\]
Using~\eqref{lowebcompara}, we obtain
\begin{equation}
\label{desigualdadgradientefun}
\left\|
\nabla F(x)
\right\|
\le
\frac{4 C^3}{\|x - x_i\|^4}
\exp\!\left(-\frac{r_i^2}{4\|x - x_i\|^2}\right)
\|g\|_{\ell^1}.
\end{equation}

Since $\mathbf{Nm}_{\varepsilon(x)}(x)$ has the same structure as $F(x)$ with $g(x_j)=1$, the same bound applies to its gradient. By the quotient rule, we deduce that
\[
\| \bar{g}_{x_1,\dots,x_N} (x)\|
\le
\frac{
\|\nabla F(x)\|\,\mathbf{Nm}_{\varepsilon(x)}(x)
+
F(x)\,\|\nabla \mathbf{Nm}_{\varepsilon(x)}(x)\|
}{\mathbf{Nm}_{\varepsilon}(x)^2}.
\]
Using the uniform lower bound $\mathbf{Nm}_{\varepsilon(x)}(x) \ge M>0$ (Section~\ref{sec:stabil}) and~\eqref{desigualdadgradientefun}, we conclude
\[
\| \bar{g}_{x_1,\dots,x_N}(x) \|
\le
\frac{8 C^3 k}{M^2}
\frac{
\exp\!\left(-\frac{r_i^2}{4\|x - x_i\|^2}\right)
\|g\|_{\ell^1}
}{\|x - x_i\|^4}.
\]
This proves the first part of the theorem.

For the second part, fix $1 \le i \le k$ and let $\alpha : I \to \mathcal{M}$ be a smooth curve such that $0 \in I$ and $\alpha(0)=x_i$. For $t \in \mathbb{R}$ sufficiently small, we observe that
\begin{equation}
\label{ecuaciondomderivada}
\begin{aligned}
\left| \frac{\bar{g}_{x_1,\dots,x_N}(\alpha(t)) - \bar{g}_{x_1,\dots,x_N}(x_i)}{t} \right|
&=
\frac{1}{|t|}
\left|
\frac{\sum_{j=1}^{N}
\exp\!\left(-\frac{\|\alpha(t)-x_j\|^2}{\|\alpha(t)-x_i\|^2}\right)
\bigl(g(x_j)-g(x_i)\bigr)}
{\mathbf{Nm}_{\varepsilon(x)}(\alpha(t))}
\right| \\
&\le
\frac{1}{M}
\sum_{j=1}^{N}
\frac{1}{|t|}
\exp\!\left(-\frac{\|\alpha(t)-x_j\|^2}{\|\alpha(t)-x_i\|^2}\right)
\lvert g(x_j)-g(x_i)\rvert .
\end{aligned}
\end{equation}

Moreover, we have
\[
\lim_{h \to 0}
\frac{\|\alpha(t)-x_j\|^2}{\|\alpha(t)-x_i\|^2}
=
\begin{cases}
1, & j=i,\\[4pt]
+\infty, & j \neq i.
\end{cases}
\]
When $j=i$, we note that $g(x_j)-g(x_i)=0$. For $j \neq i$, the rapid decay of the exponential function implies that the corresponding terms in the sum in~\eqref{ecuaciondomderivada} converge to zero as $h \to 0$. Consequently, each summand in~\eqref{ecuaciondomderivada} vanishes in the limit.

Since this holds for every smooth curve $\alpha$ with $\alpha(0)=x_i$, we conclude 
\[
\nabla \bar L(x_i)=0.
\]
This completes the proof.

\end{proof}

\section{Proof of Theorem~\ref{thm:lp_decay}}
\label{pruebateocota}

\begin{proof}
Let $r := \min \{ r_1, r_2, \dots, r_N \} > 0$ and define the auxiliary function 
$s : (0,\infty) \to \mathbb{R}$ by
\begin{equation*}
    s(t) := t^4 \exp\!\left(-\frac{r^2 t^2}{4}\right).
\end{equation*}
Then $s$ is smooth on $(0,\infty)$ and its derivative is given by
\begin{equation*}
    s'(t) = \left(4t^3 - \tfrac{1}{2} r^2 t^5\right)
    \exp\!\left(-\frac{r^2 t^2}{4}\right).
\end{equation*}
In particular, $s'(t) < 0$ for all $t > \frac{2\sqrt{2}}{r}$, and hence $s$ is strictly decreasing on 
$\left(\frac{2\sqrt{2}}{r}, \infty\right)$.

Fix $\beta > 0$ such that
\begin{equation*}
    0 < \beta < \frac{r}{2\sqrt{2}}.
\end{equation*}
Let $x \in \mathcal{M} \cap \bigcup_{i=1}^N B(x_i,\beta)$. Then there exists 
$i \in \{1,\dots,N\}$ such that $\|x - x_i\| < \beta$, which implies
\[
\frac{1}{\|x - x_i\|} > \frac{1}{\beta} > \frac{2\sqrt{2}}{r}.
\]
Since $s$ is decreasing on this interval, it follows that
\begin{equation*}
s\!\left(\frac{1}{\|x - x_i\|}\right) 
\le 
s\!\left(\frac{1}{\beta}\right).
\end{equation*}

By Theorem~\ref{thm:local_decay}, we have
\begin{align*}
\|\nabla \bar{g}_{x_1,\dots,x_N}(x)\|
&\le 
C \frac{\|g\|_{\ell^1}}{\|x - x_i\|^4}
\exp\!\left(-\frac{r^2}{4\|x - x_i\|^2}\right) \\
&= 
C \|g\|_{\ell^1} \,
s\!\left(\frac{1}{\|x - x_i\|}\right) \\
&\le 
C \|g\|_{\ell^1} \,
s\!\left(\frac{1}{\beta}\right).
\end{align*}

A direct computation yields
\begin{equation*}
s\!\left(\frac{1}{\beta}\right)
=
\frac{1}{\beta^4}
\exp\!\left(-\frac{r^2}{4\beta^2}\right).
\end{equation*}
Therefore, for all $x \in \mathcal{M} \cap \bigcup_{i=1}^N B(x_i,\beta)$,
\begin{equation*}
\|\nabla \bar{g}_{x_1,\dots,x_N}(x)\|
\le 
C \|g\|_{\ell^1}
\, \beta^{-4}
\exp\!\left(-\frac{r^2}{4\beta^2}\right).
\end{equation*}

Taking the $L^p$ norm over $\mathcal{M} \cap \bigcup_{i=1}^N B(x_i,\beta)$, we obtain
\begin{align*}
\|\nabla \bar{g}_{x_1,\dots,x_N}\|_{L^p\!\left(\mathcal{M} \cap \bigcup_{i=1}^N B(x_i,\beta)\right)}
&\le 
C \|g\|_{\ell^1}
\, \beta^{-4}
\exp\!\left(-\frac{r^2}{4\beta^2}\right)
\left(
\mathrm{vol}\!\left(\mathcal{M}\right)
\right)^{1/p}.
\end{align*}
This proves the desired estimate.
\end{proof}

\section{Proof of Theorem~\ref{teoremafiltrofreq}}
\label{pruebacotareduc}

In order to prove Theorem~\ref{teoremafiltrofreq}, we first establish a collection of auxiliary lemmas that will be used throughout the proof. Before stating the lemmas, we recall some notation. Let $\mathcal{M}$ be a  compact and without boundary Riemannian manifold embedded in $\mathbb{R}^n$. For $t>0$ and $x\in \mathbb{R}^n$, we denote by
\begin{equation*}
B(x,t)=\{y \in \mathbb{R}^n \;|\; \|x-y\|<t \}
\end{equation*}
the Euclidean ball of radius $t$ centered at $x$.

\begin{lem}
\label{lemaauxreddim1}
Let $\mathcal{M}$ be a $d$-dimensional Riemannian manifold embedded in $\mathbb{R}^n$, and let $x\in \mathcal{M}$. Then there exist a constant $M_1^x>0$, radii $r(x)>0$ and $\varepsilon>0$, and a coordinate chart 
\[
\psi : U \subset \mathbb{R}^d \to B(x,\varepsilon) \cap \mathcal{M}
\]
such that the following holds.

For every $y \in B(x,\varepsilon/2)\cap \mathcal{M}$ and every $t<r(x)$,
\begin{equation}
\label{condilocal}
B\!\left(\psi^{-1}(y),\frac{t}{M_1^x}\right)
\subseteq
\psi^{-1}\!\bigl(B(y,t)\cap \mathcal{M}\bigr)
\subseteq
B(\psi^{-1}(y),2t),
\end{equation}
where $B(\psi^{-1}(y),\cdot)$ denotes the Euclidean ball in $\mathbb{R}^d$ and $B(y,t)$ denotes the Euclidean ball in $\mathbb{R}^n$.
\end{lem}

\begin{proof}

Let $x\in\mathcal{M}$. For sufficiently small $\varepsilon>0$, consider the map
\[
\psi=\exp_x\circ T : U\subset\mathbb{R}^d \to B(0,\varepsilon) \cap \mathcal{M},
\]
which defines a normal coordinate chart around $x$. Here $\exp_x$ denotes the exponential map at $x$, and $T:\mathbb{R}^d\to T_x\mathcal{M}$ is a linear isometry (rotation) from $\mathbb{R}^d$ onto the tangent space $T_x\mathcal{M}$, which we regard as a subspace of $\mathbb{R}^n$. Note that $\psi(0)=x$.

We recall some estimates in normal coordinates that are useful for approximating differential operators. The Taylor expansion of $\psi$ around $0$ is
\begin{equation}
\psi(v)
=
x+T(v)+\frac12 D^2\psi_0(v,v)+O(\|v\|^3),
\label{taylorexpo}
\end{equation}
where $D^2\psi_0$ denotes the second differential of $\psi$ at $0$.

Let $v\in U\subset\mathbb{R}^d$ and consider the geodesic $\gamma_{T(v)}$ with initial velocity $T(v)\in T_x\mathcal{M}$. Then expansion \eqref{taylorexpo} can be expressed in terms of this geodesic as
\begin{equation}
\gamma_{T(v)}(t)
=
x+T(v)t+\frac12 D^2\psi_0(v,v)t^2+O(\|v\|^3)t^3,
\label{taylorgeodesica}
\end{equation}
for $t\in\mathbb{R}$.

Since the covariant derivative of a geodesic vanishes, it follows that $\gamma''_{T(v)}(0)$ is orthogonal to $T_x\mathcal{M}$. Using \eqref{taylorexpo} and \eqref{taylorgeodesica}, we obtain that for $v_1,v_2\in U$
\begin{equation}
\|\psi(v_1)-\psi(v_2)\|^2
=
\|T(v_1-v_2)\|^2
+
O(\|v_1-v_2\|^4),
\label{estimativaorden2}
\end{equation}
and
\begin{equation}
\mathcal{P}_{T_x\mathcal{M}}(\psi(v_1)-\psi(v_2))
=
T(v_1-v_2)
+
O(\|v_1-v_2\|^3),
\label{estimativaorden3}
\end{equation}
where $\mathcal{P}_{T_x\mathcal{M}}$ denotes the orthogonal projection onto $T_x\mathcal{M}$. Here we use the fact that $\mathcal{M}$ is embedded in $\mathbb{R}^n$.

Using estimates \eqref{estimativaorden2} and \eqref{estimativaorden3}, we deduce that there exist positive constants $M_1^x$ and $M_2^x$ such that, for $\|v_1\|$ and $\|v_2\|$ sufficiently small,
\[
\|v_1-v_2\|-M_2^x\|v_1-v_2\|^3
\le
\|\mathcal{P}_{T_x\mathcal{M}}(\psi(v_1)-\psi(v_2))\|
\le
\|\psi(v_1)-\psi(v_2)\|
\le
M_1^x\|v_1-v_2\|.
\]

In particular, if $\|v_1-v_2\|^2\le \frac{1}{2M_2^x}$, then
\begin{equation}
\label{eqinclbola}
\frac12\|v_1-v_2\|
\le
\|\psi(v_1)-\psi(v_2)\|
\le
M_1^x\|v_1-v_2\|.
\end{equation}

Consequently, by \eqref{eqinclbola} there exists $r(x)>0$ such that for all $y\in B(x,\varepsilon/2)\cap\mathcal{M}$ and all $t\le r(x)$ we have
\[
B\!\left(\psi^{-1}(y),\frac{t}{M_1^x}\right)
\subseteq
\psi^{-1}\!\bigl(B(y,t)\cap\mathcal{M}\bigr)
\subseteq
B(\psi^{-1}(y),2t).
\]

This completes the proof.

\end{proof}

\begin{lem}
\label{lemaordebola}
Let $\mathcal{M}$ be a $d$-dimensional compact manifold without boundary embedded in $\mathbb{R}^n$, and let $\rho:\mathcal{M}\to \mathbb{R}$ be a smooth, positive probability density function defining a probability measure $\mathbb{P}$ on $\mathcal{M}$. Then there exist positive constants $L_1, L_2,$ and $r>0$ such that for all $y \in \mathcal{M}$ and all $0<t \le r$,
\[
L_1 t^d \le \mathbb{P}(B(y,t)\cap \mathcal{M}) \le L_2 t^d,
\]
where for any open set $A \subseteq \mathcal{M}$,
\[
\mathbb{P}(A)=\int_A \rho(x)\, d\mathrm{Vol}(x),
\]
and $d\mathrm{Vol}$ denotes the Riemannian volume measure on $\mathcal{M}$.
\end{lem}

\begin{proof}
Since $\mathcal{M}$ is compact, by Lemma~\ref{lemaauxreddim1} there exist finitely many points $z_1,\dots,z_s \in \mathcal{M}$ such that for each $i=1,\dots,s$ there exist constants $M_1^{z_i}>0$, $\varepsilon_i>0$, and $r(z_i)<\varepsilon_i/4$, together with a coordinate chart
\[
\psi_{z_i} : U_i \subset \mathbb{R}^d \to B(z_i,\varepsilon_i)\cap \mathcal{M},
\]
satisfying the following: for every $y \in B(z_i,\varepsilon_i/2)\cap \mathcal{M}$ and every $0<t<r(z_i)$,
\begin{equation}
\label{condilocal2_refined}
B\!\left(\psi^{-1}_{z_i}(y),\frac{t}{M_1^{z_i}}\right)
\subseteq
\psi^{-1}_{z_i}\!\bigl(B(y,t)\cap \mathcal{M}\bigr)
\subseteq
B\!\left(\psi^{-1}_{z_i}(y),2t\right).
\end{equation}

Moreover, the sets $B(z_i,\varepsilon_i/2)\cap \mathcal{M}$ form an open cover of $\mathcal{M}$:
\begin{equation}
\label{eqcover_refined}
\bigcup_{i=1}^s B(z_i,\varepsilon_i/2)\cap \mathcal{M} = \mathcal{M}.
\end{equation}

Fix $y \in \mathcal{M}$. By \eqref{eqcover_refined}, there exists $i$ such that $y \in B(z_i,\varepsilon_i/2)\cap \mathcal{M}$. For $t<r(z_i) <\varepsilon_i/4$, we have
\[
B(y,t)\cap \mathcal{M} \subset B(z_i,3\varepsilon_i/4)\cap \mathcal{M} \subset B(z_i,\varepsilon_i)\cap \mathcal{M},
\]
so the set $B(y,t)\cap \mathcal{M}$ is contained in the image of the chart $\psi_{z_i}$.

Thus, using local coordinates, we can write
\begin{equation}
\label{eqacotapro_refined}
\mathbb{P}(B(y,t)\cap \mathcal{M})
=
\int_{\psi^{-1}_{z_i}(B(y,t)\cap \mathcal{M})}
\rho(\psi_{z_i}(x)) \sqrt{|\det g(x)|}\, dx,
\end{equation}
where $g$ denotes the Riemannian metric in local coordinates.

Define
\[
C_1^i := \inf_{x \in K_i} \rho(\psi_{z_i}(x)) \sqrt{|\det g(x)|},
\quad
C_2^i := \sup_{x \in K_i} \rho(\psi_{z_i}(x)) \sqrt{|\det g(x)|},
\]
where
\[
K_i := \overline{\psi^{-1}_{z_i}(B(z_i,3\varepsilon_i/4)\cap \mathcal{M})}.
\]
Since $K_i$ is compact and the integrand is continuous and strictly positive, we have $0 < C_1^i \le C_2^i < \infty$.

Using \eqref{eqacotapro_refined} together with the inclusions in \eqref{condilocal2_refined}, we obtain
\begin{equation}
\label{eqfinalbo_refined}
C_1^i \, \omega_d \left(\frac{t}{M_1^{z_i}}\right)^d
\le
\mathbb{P}(B(y,t)\cap \mathcal{M})
\le
C_2^i \, \omega_d (2t)^d,
\end{equation}
where $\omega_d$ denotes the volume of the unit ball in $\mathbb{R}^d$.

Finally, define
\[
C_1 := \min_{1\le i\le s} C_1^i, \quad
C_2 := \max_{1\le i\le s} C_2^i, \quad
M_1 := \max_{1\le i\le s} M_1^{z_i}, \quad
r := \min_{1\le i\le s} r(z_i).
\]
Combining \eqref{eqcover_refined} and \eqref{eqfinalbo_refined}, we conclude that for all $y \in \mathcal{M}$ and all $0<t \le r$,
\[
\frac{C_1 \,\omega_d}{M_1^d} t^d
\le
\mathbb{P}(B(y,t)\cap \mathcal{M})
\le
C_2 \,\omega_d \, 2^d t^d.
\]

This completes the proof.
\end{proof}

\begin{lem}
\label{lem:covering_bound}
Let $\mathcal{M}$ be a $d$-dimensional compact manifold without boundary embedded in $\mathbb{R}^n$, and let $\rho:\mathcal{M}\to \mathbb{R}$ be a smooth, strictly positive probability density defining a probability measure $\mathbb{P}$ on $\mathcal{M}$. 

Let $\varepsilon>0$ be sufficiently small, and let $X_1,\dots,X_N$ be i.i.d.\ random variables with distribution $\rho$. Then there exist positive constants $s(\varepsilon)$ and $L_1$ such that
\[
\mathbb{P}\!\left( \max_{x \in \mathcal{M}} \min_{1 \le j \le N} \|x - X_j\| > \varepsilon \right)
\le s(\varepsilon)\,(1 - L_1 \varepsilon^d)^N.
\]
\end{lem}

\begin{proof}
Since $\mathcal{M}$ is compact, there exist points $z_1,\dots,z_{s(\varepsilon)} \in \mathcal{M}$ such that
\begin{equation}
\label{eq:cover}
\mathcal{M} \subset \bigcup_{i=1}^{s(\varepsilon)} B(z_i,\varepsilon/2).
\end{equation}

Fix $i \in \{1,\dots,s(\varepsilon)\}$. Since the $X_j$ are independent and identically distributed,
\[
\mathbb{P}\!\left( X_j \notin B(z_i,\varepsilon/2)\ \text{for all } 1 \le j \le N \right)
= \big(1 - \mathbb{P}(B(z_i,\varepsilon/2))\big)^N.
\]

By the positivity and smoothness of $\rho$, and by Lemma~\ref{lemaordebola}, there exists a constant $L_1>0$ such that for all sufficiently small $\varepsilon$,
\[
\mathbb{P}(B(z_i,\varepsilon/2)) \ge L_1 \varepsilon^d.
\]
Hence,
\[
\mathbb{P}\!\left( X_j \notin B(z_i,\varepsilon/2)\ \forall j \right)
\le (1 - L_1 \varepsilon^d)^N.
\]

Taking the union over $i=1,\dots,s(\varepsilon)$ and using the union bound,
\begin{equation}
\label{eq:union}
\mathbb{P}\!\left( \bigcup_{i=1}^{s(\varepsilon)} \bigcap_{j=1}^N \{X_j \notin B(z_i,\varepsilon/2)\} \right)
\le s(\varepsilon)\,(1 - L_1 \varepsilon^d)^N.
\end{equation}

We now prove the inclusion of events:
\begin{equation}
\label{eq:inclusion}
\left\{ \max_{x \in \mathcal{M}} \min_{1 \le j \le N} \|x - X_j\| > \varepsilon \right\}
\subset
\bigcup_{i=1}^{s(\varepsilon)} \bigcap_{j=1}^N \{X_j \notin B(z_i,\varepsilon/2)\}.
\end{equation}

Indeed, suppose
\[
\max_{x \in \mathcal{M}} \min_{1 \le j \le N} \|x - X_j\| > \varepsilon.
\]
Then there exists $x \in \mathcal{M}$ such that
\[
\|x - X_j\| > \varepsilon \quad \text{for all } j=1,\dots,N.
\]

By the covering \eqref{eq:cover}, there exists $i$ such that $x \in B(z_i,\varepsilon/2)$. Then for all $j$,
\[
\|x - X_j\|
\le \|x - z_i\| + \|z_i - X_j\|
\le \frac{\varepsilon}{2} + \|z_i - X_j\|.
\]
Thus,
\[
\varepsilon < \frac{\varepsilon}{2} + \|z_i - X_j\|
\quad \Rightarrow \quad
\|z_i - X_j\| > \frac{\varepsilon}{2},
\]
which implies $X_j \notin B(z_i,\varepsilon/2)$ for all $j$. This proves \eqref{eq:inclusion}.

Combining \eqref{eq:union} and \eqref{eq:inclusion}, we obtain
\[
\mathbb{P}\!\left( \max_{x \in \mathcal{M}} \min_{1 \le j \le N} \|x - X_j\| > \varepsilon \right)
\le s(\varepsilon)\,(1 - L_1 \varepsilon^d)^N,
\]
which completes the proof.
\end{proof}

\begin{lem}
\label{lem:voronoi_quadrature}
Let $\mathcal{M}$ be a compact Riemannian manifold without boundary, and let 
$g : \mathcal{M} \to \mathbb{R}$ be a smooth function. Let $x_1, \dots, x_N \in \mathcal{M}$ 
and consider the associated Voronoi partition $\{R_k\}_{k=1}^N$ defined by
\[
R_k = \left\{ y \in \mathcal{M} : \|y- x_k\| \le \|y- x_i\| \ \text{for all } 1 \le i \le N \right\},
\]
Define the radius of each Voronoi cell by
\[
\mathrm{Rad}(R_k) := \sup_{x \in R_k} \|x - x_k\|.
\]

Then there exist constants $C > 0$ and $\beta > 0$, depending only on $g$ and $\mathcal{M}$, such that if
\[
\max_{1 \le k \le N} \mathrm{Rad}(R_k) \le \beta,
\]
we have
\[
\left| \int_{\mathcal{M}} g(x)\, d\mathrm{Vol}(x) 
- \sum_{k=1}^N g(x_k)\, \mathrm{Vol}(R_k) \right|
\le C \, \mathrm{Vol}(\mathcal{M})\, \max_{1 \le k \le N} \mathrm{Rad}(R_k)\,.
\]
\end{lem}

\begin{proof}
Since $g$ is smooth and $\mathcal{M}$ is compact, its gradient is bounded. 
Hence $g$ is locally Lipschitz continuous: there exist constants $C > 0$ and $\beta > 0$ such that for all $x,y \in \mathcal{M}$ with $\|x -y\| \le \beta$,
\[
|g(x) - g(y)| \le C\, \|x -y\|,
\]
Using that the Voronoi cells form a partition of $\mathcal{M}$ up to sets of measure zero, we write
\begin{align*}
&\left| \int_{\mathcal{M}} g(x)\, d\mathrm{Vol}(x) 
- \sum_{k=1}^N g(x_k)\, \mathrm{Vol}(R_k) \right| \\
&= \left| \sum_{k=1}^N \int_{R_k} \big(g(x) - g(x_k)\big)\, d\mathrm{Vol}(x) \right| \\
&\le \sum_{k=1}^N \int_{R_k} |g(x) - g(x_k)| \, d\mathrm{Vol}(x).
\end{align*}

For $x \in R_k$, we have $\|x - x_k\| \le \mathrm{Rad}(R_k)$, hence
\[
|g(x) - g(x_k)| \le C\, \mathrm{Rad}(R_k).
\]
Therefore,
\begin{align*}
\left| \int_{\mathcal{M}} g(x)\, d\mathrm{Vol}(x) 
- \sum_{k=1}^N g(x_k)\, \mathrm{Vol}(R_k) \right|
&\le C \sum_{k=1}^N \mathrm{Rad}(R_k)\, \mathrm{Vol}(R_k) \\
&\le C \, \max_{1 \le k \le N} \mathrm{Rad}(R_k) \sum_{k=1}^N \mathrm{Vol}(R_k).
\end{align*}

Since the cells partition $\mathcal{M}$,
\[
\sum_{k=1}^N \mathrm{Vol}(R_k) = \mathrm{Vol}(\mathcal{M}),
\]
which concludes the proof.
\end{proof}

With the previous lemmas at hand, we are now ready to prove Theorem~\ref{teoremafiltrofreq}.

\medskip
\noindent\textbf{Proof of Theorem~\ref{teoremafiltrofreq}.}
Let $\varepsilon > 0$ be fixed and sufficiently small. For each $1 \le k \le N$, define the Voronoi cell associated with $X_k$ by
\begin{equation*}
R_k := \left\{ y \in \mathcal{M} \;:\; \|y - X_k\| \le \|y - X_i\| \quad \text{for all } 1 \le i \le N \right\}.
\end{equation*}
We define the radius of $R_k$ as
\[
\mathrm{Rad}(R_k) := \sup_{y \in R_k} \|y - X_k\|.
\]

By Lemma~\eqref{lem:covering_bound}, there exist positive constants $s(\varepsilon)$ and $L_1$ such that
\begin{equation*}
\mathbb{P}\!\left( \sup_{x \in \mathcal{M}} \min_{1 \le j \le N} \|x - X_j\| > \varepsilon \right)
\le s(\varepsilon)\bigl(1 - L_1 \varepsilon^d\bigr)^N.
\end{equation*}
In particular, this implies
\begin{equation*}
\mathbb{P}\!\left( \max_{1 \le k \le N} \mathrm{Rad}(R_k) \le \varepsilon \right)
\ge 1 - s(\varepsilon)\bigl(1 - L_1 \varepsilon^d\bigr)^N.
\end{equation*}

Assume now that
\[
\max_{1 \le k \le N} \mathrm{Rad}(R_k) \le \varepsilon.
\]
Applying Lemma~\eqref{lem:voronoi_quadrature} to the function
\[
x \mapsto \|\nabla \overline{g}_{x_1,\dots,x_N}(x)\|^p,
\]
we obtain the existence of a constant $M > 0$ such that
\begin{align*}
\left| 
\int_{\mathcal{M}} \|\nabla \overline{g}_{x_1,\dots,x_N}(x)\|^p \, d\mathrm{Vol}(x)
- \sum_{k=1}^N \|\nabla \overline{g}_{x_1,\dots,x_N}(x_k)\|^p \, \mathrm{Vol}(R_k)
\right|
\le M \, \mathrm{Vol}(\mathcal{M}) \, \varepsilon.
\end{align*}

On the other hand, by Theorem~\ref{thm:local_decay}, we have for all $1 \le k \le N$ that
\[
\|\nabla \overline{g}_{x_1,\dots,x_N}(x_k)\|^p = 0.
\]
Therefore,
\[
\sum_{k=1}^N \|\nabla \overline{g}_{x_1,\dots,x_N}(x_k)\|^p \, \mathrm{Vol}(R_k) = 0.
\]

Combining the previous estimates, we deduce
\begin{align*}
\|\nabla \overline{g}_{x_1,\dots,x_N}\|_{L^p(\mathcal{M})}^p
= \int_{\mathcal{M}} \|\nabla \overline{g}_{x_1,\dots,x_N}(x)\|^p \, d\mathrm{Vol}(x)
\le M \, \mathrm{Vol}(\mathcal{M}) \, \varepsilon.
\end{align*}

Finally, recalling the probability estimate above, we conclude that
\begin{equation*}
\mathbb{P}\!\left(
\|\nabla \overline{g}_{x_1,\dots,x_N}\|_{L^p(\mathcal{M})}^p
\le M \, \mathrm{Vol}(\mathcal{M}) \, \varepsilon
\right)
\ge 1 - s(\varepsilon)\bigl(1 - L_1 \varepsilon^d\bigr)^N.
\end{equation*}

\hfill $\blacksquare$

\end{document}